\documentclass[runningheads]{llncs}

\PassOptionsToPackage{table}{xcolor}
\newcommand{\best}[1]{\cellcolor{red!25}\textbf{#1}}   %
\newcommand{\snd}[1]{\cellcolor{orange!25}#1}          %
\newcommand{\trd}[1]{\cellcolor{yellow!25}#1}          %

\usepackage[mobile]{eccv}

\newlength\paramargin
\newlength\figmargin
\newlength\subfigmargin
\newlength\presecmargin
\newlength\secmargin
\newlength\subsecmargin
\newlength\subsubsecmargin
\newlength\tabmargin
\newlength\eqmargin
\newlength\paraskip

\usepackage{eccvabbrv}

\usepackage{graphicx}
\usepackage{booktabs}
\usepackage[table]{xcolor}

\usepackage{enumitem}
\usepackage{subcaption}
\usepackage{gensymb}
\usepackage{arydshln}
\usepackage{multirow}
\usepackage{booktabs}      %
\usepackage{threeparttable}
\usepackage{xcolor}        %
\usepackage{pifont}        %
\usepackage{xspace}
\usepackage{algorithm}      %
\usepackage{algorithmicx}
\usepackage{algpseudocode}  %

\definecolor{ForestGreen}{RGB}{34,139,34}

\algrenewcommand\algorithmicrequire{\textbf{Input:}}
\algrenewcommand\algorithmicensure{\textbf{Output:}}

\usepackage[accsupp]{axessibility}  %

\usepackage{hyperref}

\usepackage{orcidlink}

\begin{document}

\title{Skyfall-GS: Synthesizing Immersive 3D Urban Scenes from Satellite Imagery} 

\titlerunning{Skyfall-GS}

\author{Jie-Ying Lee\inst{1} \and Yi-Ruei Liu\inst{2} \and Shr-Ruei Tsai\inst{1} \and Wei-Cheng Chang\inst{1} \and Chung-Ho Wu\inst{1} \and Jiewen Chan\inst{1} \and Zhenjun Zhao\inst{3} \and Chieh Hubert Lin\inst{4} \and Yu-Lun Liu\inst{1}}

\authorrunning{J.-Y.~Lee et al.}

\institute{\textsuperscript{\rm 1} National Yang Ming Chiao Tung University, \textsuperscript{\rm 2} UIUC\\\textsuperscript{\rm 3} University of Zaragoza, \textsuperscript{\rm 4} UC Merced\\
\email{jayinnn.cs14@nycu.edu.tw, yulunliu@cs.nycu.edu.tw}}

\maketitle

\begin{figure}[htbp]
    \centering
    \includegraphics[width=\textwidth]{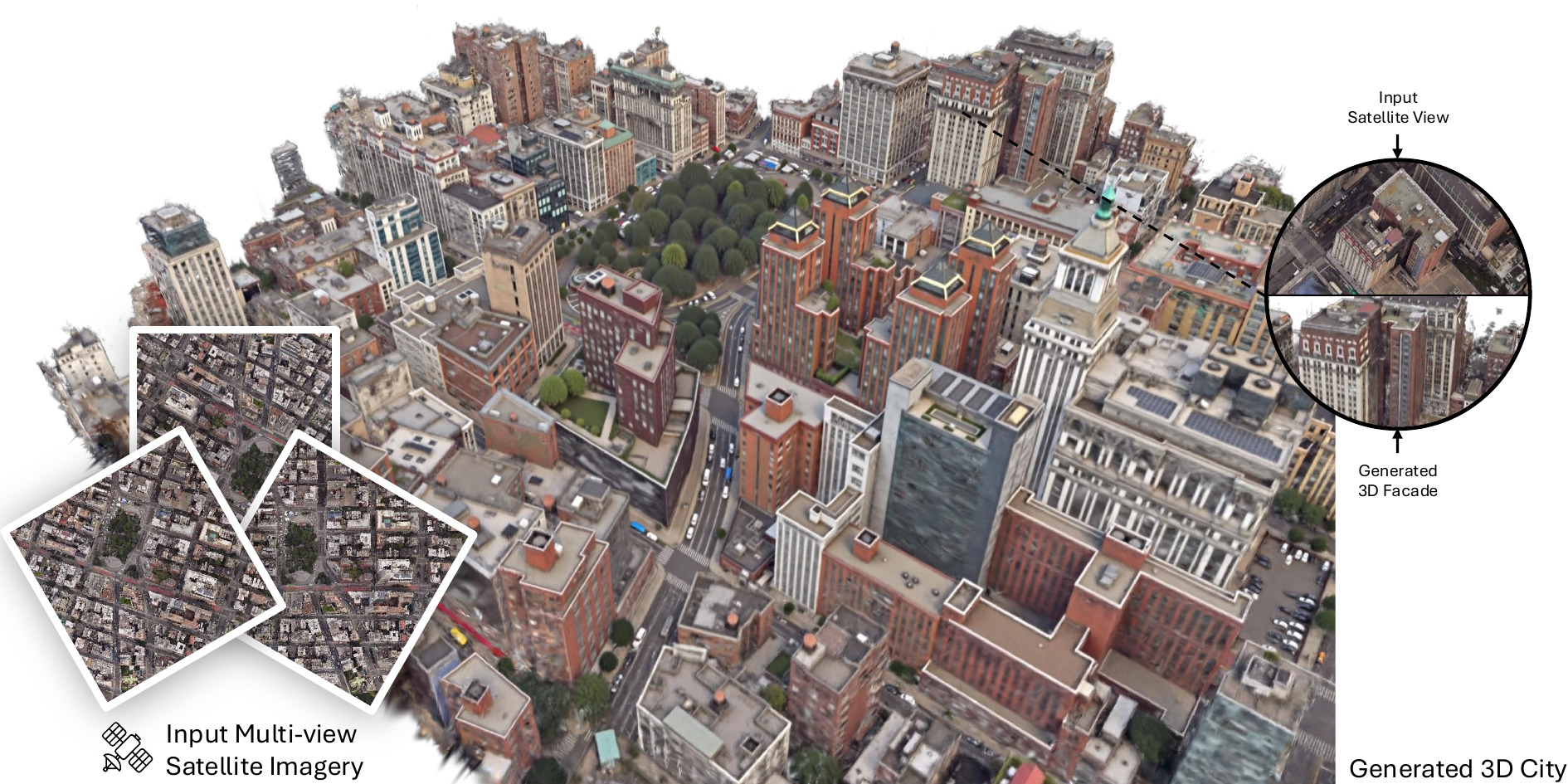}
    \caption{\textbf{Our method synthesizes high-quality, immersive 3D urban scenes solely from multi-view satellite imagery, enabling realistic drone-view navigation without relying on additional 3D or street-level training data.} Given multiple satellite images from diverse viewpoints and dates (\emph{left}), our method leverages 3D Gaussian Splatting combined with pre-trained text-to-image diffusion models in an iterative refinement framework to generate realistic 3D block-scale city from limited satellite-view input (\emph{right}). Our method significantly enhances visual fidelity, geometric sharpness, and semantic consistency, enabling real-time immersive exploration.}
    \label{teaser}
\end{figure}

\begin{abstract}
    Synthesizing large-scale, explorable, and geometrically accurate 3D urban scenes is a challenging yet valuable task for immersive and embodied applications. The challenge lies in the lack of large-scale and high-quality real-world 3D scans for training generalizable generative models. In this paper, we take an alternative route to create large-scale 3D scenes by leveraging readily available satellite imagery for realistic coarse geometry and open-domain diffusion models for high-quality close-up appearance synthesis. We propose \textbf{Skyfall-GS}, a novel hybrid framework that synthesizes immersive city-block scale 3D urban scenes by combining satellite reconstruction with diffusion refinement, eliminating the need for costly 3D annotations, and also featuring real-time, immersive 3D exploration. We tailor a curriculum-driven iterative refinement strategy to progressively enhance geometric completeness and photorealistic texture. Extensive experiments demonstrate that Skyfall-GS provides improved cross-view consistent geometry and more realistic textures compared to state-of-the-art approaches.
    Project page: \url{https://skyfall-gs.jayinnn.dev/}
\end{abstract}

\vspace{\secmargin}
\section{Introduction}
\label{sec:intro}
\vspace{\subsecmargin}

High-quality, immersive, and semantically plausible 3D urban scenes are essential for a wide range of applications, including gaming, filmmaking, navigation planning, and robotics.
The ability to create a large-scale and 3D-grounded environment supports realistic rendering and immersive experience for storytelling, demonstration, and embodied physics simulation.
However, due to limited 3D-informed data, building a generative model for realistic and navigable 3D cities remains challenging.
It is expensive and labor-intensive to acquire large-scale 3D and textured reconstructions of cities with detailed geometry, while using Internet image collections faces challenges in camera pose registration and excessive data noise (\eg, transient objects and different times of the day).
These constraints prevent existing 3D city generation frameworks from creating realistic and diverse appearances.
With this observation, we propose an alternative route for virtual city creation with a two-stage pipeline: partial and coarse geometry reconstruction from multi-view satellite imagery, then close-up appearance completion and synthesis using an open-domain diffusion model.

\begin{figure}[t]
    \centering
    \includegraphics[width=\linewidth]{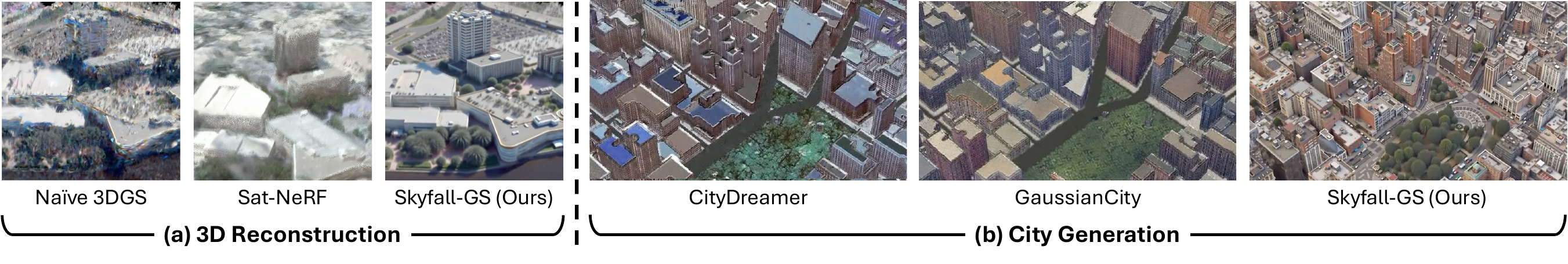}
    \caption{
    \textbf{Limitations of existing novel-view synthesis methods from satellite imagery.} (a) Sat-NeRF~\cite{mari2022sat} and naive 3DGS~\cite{kerbl20233d} yield blurred or distorted building facades due to insufficient geometric detail and limited parallax from satellite viewpoints. (b) City generation methods~\cite{xie2024citydreamer,xie2025gaussiancity} produce oversimplified building geometries and unrealistic appearances, primarily due to strong assumptions about the input data, and overfitting to small-scale, domain-specific datasets. In comparison, our method synthesizes more realistic appearances and geometries from aerial views.
    }
    \label{fig:why}
\end{figure}

Satellite imagery offers a compelling alternative due to its extensive geographic coverage, automated collection, and high-resolution capabilities. For instance, Maxar's WorldView-3 satellite captures approximately 680,000 $\text{km}^\text{2}$ of imagery daily at resolutions up to 31 cm per pixel. Such data encodes a large volume of real-world environment semantics, enabling scalable 3D urban scene creation. 
However, in \Cref{fig:why}(a), we show that directly applying 3D reconstruction methods to satellite imagery is insufficient for creating \textit{navigable and immersive} 3D cities.
The substantial invisible regions (\eg, building facades) and limited satellite-view parallax create incorrect geometry and artifacts.

Completing and enhancing the geometry and texture in the ground view requires a significant influx of extra information. 
In \Cref{fig:why}(b), we study a few state-of-the-art methods in city generation~\cite{xie2024citydreamer, xie2025gaussiancity}.
These methods produce oversimplified building geometries and unrealistic appearances due to strong assumptions, particularly the reliance on semantic maps and height fields as the sole inputs, and overfitting to small-scale, domain-specific datasets.
Such an observation motivates us to leverage an open-domain foundation image generation model as an external information source, providing better zero-shot generalization and diversity.
Ground-level novel-view renderings from the GS-reconstructed scene suffer from severe degradation, including floater artifacts and texture smearing, caused by insufficiently constrained Gaussians in regions with limited satellite-view parallax. To address this, we employ an open-domain foundation image generation model to directly refine these degenerate renderings into photorealistic and geometrically consistent outputs, exploiting the model's rich visual priors to recover plausible appearances.
The refined outputs serve as pseudo ground-truth to supervise iterative GS scene optimizations, progressively improving the visual fidelity and geometric consistency across the scene.
To stabilize the convergence, we carefully design a curriculum-based view selection and iterative refinement process, where the sampled view angles gradually \textit{fall} from the \textit{sky} to the ground over time.
Accordingly, we name our framework \textbf{Skyfall-GS}.
In \Cref{teaser} and \Cref{fig:why}, we show that Skyfall-GS substantially enhances texture with 3D-justified geometry compared to the relevant baselines.

Skyfall-GS is a novel hybrid framework that synthesizes immersive 3D urban scenes by combining satellite reconstruction with diffusion refinement, eliminating the need for fixed-domain training on 3D data.
Skyfall-GS operates on readily available satellite imagery as the only input, then synthesizes realistic aerial-view appearances and maintains a strong satellite-to-ground 3D consistency.
Moreover, Skyfall-GS supports real-time and interactive rendering.
Through experiments on diverse environments, we show that Skyfall-GS has better generalization and robustness compared to state-of-the-art methods.
Our ablation study shows that each component improves perceptual plausibility and semantic consistency.
Skyfall-GS paves the way for scalable 3D urban virtual scene creation, enabling applications in virtual entertainment, simulation, and robotics.

In summary, our contributions include:
\begin{itemize}[leftmargin=*,itemsep=0pt,topsep=0pt]
\item We introduce Skyfall-GS, the first method to synthesize immersive, real-time free-flight navigable 3D urban scenes solely from multi-view satellite imagery, without domain-specific 3D training. Its novelty lies in three integrations tailored to the satellite-to-ground setting:
\item A satellite-specific 3DGS reconstruction that pairs entropy-based opacity regularization with pseudo-camera depth supervision; we are the first to demonstrate that these designs tame the limited-parallax, multi-date satellite-imagery setting.
\item A curriculum-based iterative dataset update that couples a high-to-low elevation schedule with multi-sample consensus, distilling open-domain diffusion priors into photorealistic, cross-view-consistent facades in occluded regions.
\end{itemize}

\section{Related Work}
\label{sec:related}

\subsubsection{Gaussian Splatting.}
3D Gaussian Splatting (3DGS)~\cite{kerbl20233d} offers real-time view synthesis rivaling NeRFs~\cite{mildenhall2021nerf,barron2021mipnerf,barron2022mipnerf360, mueller2022instant,barron2023zipnerf,martin2021nerf}, with Mip-Splatting~\cite{yu2024mip} addressing scale-change aliasing. In-the-wild variants handle appearance variation and transient objects~\cite{xu2024wildgsrealtimenovelview,sabourgoli2024spotlesssplats, wang2024wegsinthewildefficient3d,dahmani2024swagsplattingwildimages, zhang2024gaussian,kulhanek2024wildgaussians,hou20253d,fan2025spectromotion}, while scaling to large scenes is addressed via partitioning and LOD~\cite{kerbl2024hierarchical,lin2024vastgaussianvast3dgaussians,turki2022mega,tancik2022block,yuchen2024dogaussian,gao2025citygsx}. For sparse-view settings, depth and co-regularization priors guide reconstruction~\cite{li2024dngaussian,zhang2024cor,zhu2023FSGS,niemeyer2022regnerf, lin2025frugalnerf,zhang2023sparsesat}.

\vspace{\subsubsecmargin}
\subsubsection{Satellite and aerial 3D reconstruction.}
Classical SfM-MVS pipelines extract DSMs from satellite imagery~\cite{schoenberger2016sfm,VisSat-2019,GAO2023446,bosch2017mvs3d}, with neural variants improving geometric fidelity~\cite{Derksen_2021_CVPR,Mari_2023_CVPR,app14072729,Leotta_2019_CVPR_Workshops,liu2025satdnimplicitsurfacereconstruction,rs15174297,gao2024enhanced3durbanscene}. Sat-NeRF~\cite{mari2022sat} and SatMVS~\cite{gao2021satmvs,gao2023satmvs} apply NeRF and RPC-based warping to satellite imagery respectively, yet neither recovers occluded facades, while FusionRF~\cite{sprintson2024fusionrf} improves depth via multispectral acquisitions. EOGS~\cite{savantaira2024eogs} adapts 3DGS with affine cameras and shadow mapping for sparse multi-date satellite imagery, and SkySplat~\cite{huang2025skysplat} outputs Gaussians from satellite imagery in a feed-forward manner. For aerial imagery, AGS~\cite{wu2024ags} introduces Ray-Gaussian Intersection for large-scale surface reconstruction, CityGaussian~\cite{liu2025citygaussian,liu2024citygaussianv2} achieves city-scale rendering via scene partitioning and LOD, and Horizon-GS~\cite{jiang2025horizon} unifies aerial-to-ground reconstruction. However, no existing method jointly handles multi-date appearance variations in satellite imagery while synthesizing ground-level perspectives.

\vspace{\subsubsecmargin}
\subsubsection{Urban scene synthesis.}
Cross-view synthesis methods~\cite{regmi2018crossview, shi2022geometry, xu2024geospecific, xu2025satellitetogroundscape, ze2025controllablesatellitetostreetviewsynthesisprecise, deng2024streetscapes, toker2021coming} generate ground-level images or videos from satellite imagery without explicit 3D representations, while others employ intermediate 3D structures such as point clouds~\cite{li2021sat2vidstreetviewpanoramicvideo}, density fields~\cite{qian2023sat2density, qian2026sat2densitypp}, or voxels~\cite{li2024crossviewdiffcrossviewdiffusionmodel}, and recent methods~\cite{li2024sat2scene, yao2025magiccity, qian2026sat3dgen, hua2025sat2city, kang2025sat2realcity} advance toward explicit 3D generation via diffusion- and lifting-based approaches. Layout-conditioned methods synthesize cities from BEV maps: InfiniCity~\cite{lin2023infinicity} lifts infinite-pixel BEV maps to 3D via octree voxel completion and neural rendering, CityDreamer~\cite{xie2024citydreamer} introduces compositional generative models separating buildings from backgrounds, and GaussianCity~\cite{xie2025gaussiancity} extends this paradigm to 3D Gaussian Splatting~\cite{xie2025citydreamer4d}; LLM-driven procedural approaches~\cite{zhang2024cityxcontrollableproceduralcontent, zhou2025scenex, shang2024urbanworld} further generate layouts from text, OSM, or semantic maps. Although CityX~\cite{zhang2024cityxcontrollableproceduralcontent} can accept satellite images as layout guidance, none of these methods faithfully synthesize photorealistic ground-level views directly conditioned on the observed satellite texture.

\vspace{\subsubsecmargin}
\subsubsection{Diffusion models for 3D reconstruction and editing.} Diffusion models~\cite{Rombach_2022_CVPR,flux2024} have emerged as powerful generative priors for image synthesis. Score distillation pipelines~\cite{poole2022dreamfusion,lin2023magic3d,wang2023prolificdreamer} lift 2D priors into 3D; DreamGaussian~\cite{tang2024dreamgaussian} and GaussianDreamer~\cite{yi2024gaussiandreamer} extend this to Gaussian Splatting, while MVDream~\cite{shi2023mvdream} and sparse-view methods~\cite{wu2023reconfusion,liu2023deceptive,chen2024vi3drmtowardsmeticulous3dreconstruction,gao2024cat3d,wu2024cat4dcreate4dmultiview,melaskyriazi2024im3d,chung2023luciddreamer,liu2023zero} address multi-view consistency. Inversion-based methods~\cite{meng2022sdedit,mokady2022null,miyake2024negativepromptinversionfastimage,kulikov2024flowedit} and occlusion-aware inpainting works~\cite{mirzaei2023spinnerf,liu2025corrfill,wu2025aurafusion360augmentedunseenregion,gaussian_grouping,GaussianEditor,mirzaei2023watchyoursteps,signerf,weber2023nerfiller,gaussctrl2024,wang2025viewconsistent3deditinggaussian,shen2022cfnerf} enable scene editing and 3D Gaussian inpainting. Instruct-NeRF2NeRF~\cite{instructnerf2023} introduced the Iterative Dataset Update (IDU) paradigm, iteratively refining a NeRF via InstructPix2Pix~\cite{brooks2023instructpix2pix} edits; LucidDreamer~\cite{chung2023luciddreamer} and WonderWorld~\cite{yu2025wonderworld} adopt IDU for progressive scene expansion via extrapolation, and RealmDreamer~\cite{shriram2024realmdreamer} generates forward-facing scenes via iterative inpainting and depth diffusion. However, none of these can be naively applied to our setting: inpainting-based approaches treat facade regions as unknown and synthesize them freely, whereas satellite imagery does capture building facades at an oblique angle, providing appearance constraints that must be respected. Ignoring this leads to synthesized facades inconsistent with the satellite observations.

\vspace{\secmargin}
\section{Method}
\label{sec:method}
\vspace{\subsecmargin}

\begin{figure}[t]
\centering
\includegraphics[width=\linewidth]{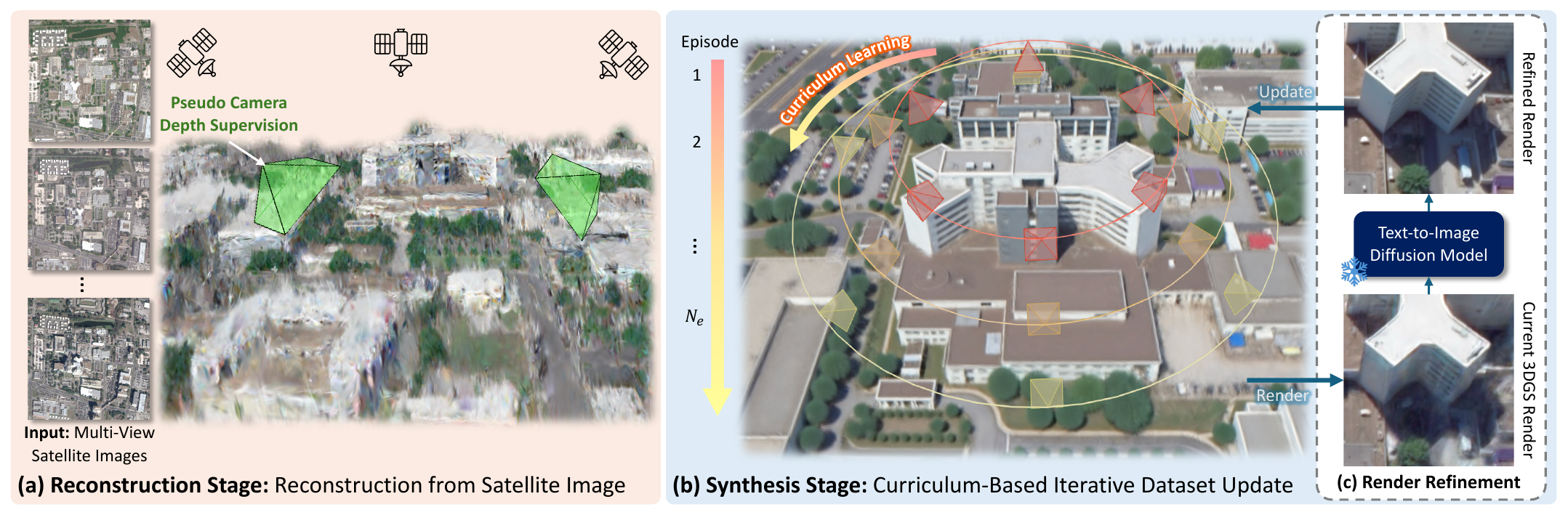}
\caption{\textbf{Overview of the proposed Skyfall-GS pipeline.}
Our method synthesizes immersive and free-flight navigable city-block scale 3D scenes solely from multi-view satellite imagery in two stages. (a) In the Reconstruction Stage, we first reconstruct the initial 3D scene using 3DGS, enhanced by pseudo-camera depth supervision to address limited parallax in satellite images. We integrate an appearance modeling to handle varying illumination conditions across multi-date satellite images. (b) In the Synthesis Stage, we introduce a curriculum-based Iterative Dataset Update (IDU) refinement technique leveraging (c) a pre-trained T2I diffusion model~\cite{flux2024} with prompt-to-prompt editing~\cite{kulikov2024flowedit}. By iteratively updating training datasets with progressively refined renders, our approach significantly reduces visual artifacts, improving geometric accuracy and texture realism, particularly in previously occluded areas such as building facades.
}
\label{fig:framework}
\end{figure}

Our two-stage pipeline (\Cref{fig:framework}) transforms satellite images into immersive 3D cities.
In the Reconstruction Stage (\Cref{sec:sat-view}), we fit a 3D Gaussian Splatting model, adding illumination-adaptive appearance modeling and regularizers for sparse, multi-date views.
In the Synthesis Stage (\Cref{sec:idu}), we recover occluded regions, e.g., facades, through curriculum-based Iterative Dataset Update, repeatedly refining renders with text-guided diffusion edits.
The loop keeps textures faithful to the satellite input while preserving geometry, yielding complete, navigable urban scenes from satellite data alone.

\vspace{\subsubsecmargin}
\subsubsection{Preliminary.} 3D Gaussian Splatting (3DGS) \cite{kerbl20233d} encodes a scene as Gaussians with center $\mu_i$, covariance $\Sigma_i$, opacity $\alpha_i$, and view-dependent color. Each Gaussian projects to the image plane with covariance: $\Sigma_i' = JW\Sigma_iW^TJ^T$, 
where $W$ is the viewing transformation and $J$ is the affine-projection Jacobian. Pixels are alpha-composited front-to-back. Parameters are trained with: %
\begin{equation} \small
    \mathcal{L}_{\text{color}} = \lambda_{\text{D-SSIM}} \,\text{DSSIM}(\hat{C}, C) + (1-\lambda_{\text{D-SSIM}})\lVert \hat{C} - C\rVert_1 \,\, .
\end{equation}

\subsection{Initial 3DGS Reconstruction from Satellite Imagery}
\label{sec:sat-view}
\vspace{\subsecmargin}

The initial 3DGS reconstruction must faithfully preserve the texture and geometry of satellite imagery to provide a robust foundation for synthesis. We employ appearance modeling to handle variations in multi-date imagery. Since limited satellite parallax creates floating artifacts, we apply regularization techniques to constrain both texture and geometry.

\vspace{\subsubsecmargin}
\subsubsection{Approximated camera parameters.}
Satellite imagery typically uses the rational polynomial camera (RPC) model, directly mapping image coordinates to geographic coordinates. To integrate with the 3DGS pipeline, we employ SatelliteSfM~\cite{VisSat-2019} to approximate perspective camera parameters (extrinsic and intrinsic) from RPC and generate sparse SfM points as initial 3DGS points.

\vspace{\subsubsecmargin}
\subsubsection{Appearance modeling.}
Multi-date satellite imagery exhibits significant appearance variations due to global illumination changes, seasonal factors, and transient objects (\Cref{fig:framework}(a)). Following WildGaussians~\cite{kulhanek2024wildgaussians}, we use trainable per-image embeddings $\{e_j\}_{j=1}^N$ (with $N$ input images) to handle varying illumination. We also employ trainable per-Gaussian embeddings $g_i$ to capture localized appearance changes, \eg shadow variations. A lightweight MLP $f$ computes affine color transformation parameters $(\beta,\gamma)$ as $(\beta,\gamma) = f(e_j,g_i,\bar{c}_i),$ where $e_j$ is the per-image embedding, $g_i$ is the per-Gaussian embedding, and $\bar{c}_i$ denotes the zeroth-order spherical harmonics (SH). Let $\hat{c}_i(\mathbf{r})$ be the i-th Gaussian’s view-dependent color conditioned on the ray direction $\mathbf{r}$. The transformed color $\tilde{c}_i$ is computed as $\tilde{c}_i(\mathbf{r}) = \gamma \cdot \hat{c}_i(\mathbf{r}) + \beta$. To avoid modeling the appearance changes as view-dependent effects, we limit SH coefficients to zero- and first-order terms. At inference, we fuse the learned appearance into a standard 3DGS by selecting a fixed image embedding $e^*$ and evaluating $f(e^*, g_i, \bar{c}_i)$ for every Gaussian to compute a static color. The embeddings and MLP are then discarded, yielding a portable representation compatible with standard 3DGS renderers and enabling real-time rendering at 60~FPS (1920$\times$1080) on a MacBook Pro M4 Pro.

\vspace{\subsubsecmargin}
\subsubsection{Opacity regularization.}
We observe that numerous floaters in reconstructed scenes exhibit low opacity. To encourage geometry to adhere closely to actual surfaces, we propose entropy-based opacity regularization: 
\begin{equation} 
\small
\mathcal{L}_{\text{op}} = -\sum_i \left[ \alpha_i \log(\alpha_i) + (1-\alpha_i)\log(1-\alpha_i) \right] \,\, . 
\end{equation}
This regularization promotes binary opacity distributions, allowing low-opacity Gaussians to be more aggressively pruned during densification. Incorporating this term significantly sharpens geometric reconstruction, providing a better foundation for subsequent synthesis.

\vspace{\subsubsecmargin}
\subsubsection{Pseudo camera depth supervision.}
To further reduce floating artifacts, we sample pseudo-cameras positioned closer to the ground during optimization. From these pseudo-cameras, we render RGB images $I_{\text{RGB}}$ and corresponding alpha-blended depth maps $\hat{D}_{\text{GS}}$. We then use an off-the-shelf monocular depth estimator, MoGe~\cite{wang2024moge}, to predict scale-invariant depths $\hat{D}_{\text{est}}$ from these renders. We use the absolute value of Pearson correlation (PCorr) to supervise the depth:
\begin{equation} \small
\mathcal{L}_{\text{depth}} = 1 - \lVert \text{PCorr}(\hat{D}_{\text{GS}}, \hat{D}_{\text{est}})\rVert_1 \,\, ; \quad
\text{PCorr}(\hat{D}_{\text{GS}}, \hat{D}_{\text{est}}) = \frac{\text{Cov}(\hat{D}_{\text{GS}}, \hat{D}_{\text{est}})}{\sqrt{\text{Var}(\hat{D}_{\text{GS}})\text{Var}(\hat{D}_{\text{est}})}} \,\, .
\end{equation}

\vspace{\subsubsecmargin}
\subsubsection{Optimization.}
Combining all components, the overall loss for the reconstruction stage is defined as:
\begin{equation} \label{eq:sat-loss}
    \small
    \mathcal{L}_{\text{sat}}(G, C) = \mathcal{L}_{\text{color}} + \lambda_{\text{op}} \mathcal{L}_{\text{op}} + \lambda_{\text{depth}} \mathcal{L}_{\text{depth}} \,\, ,
\end{equation}
where $G$ is the 3DGS representation, $C$ is the set of ground-truth satellite images, $\lambda_{\text{op}}$ and $\lambda_{\text{depth}}$ 
weight opacity regularization and depth supervision relative to the color reconstruction loss.

\begin{figure}[t]
    \centering
    \begin{minipage}[t]{0.38\textwidth}
        \centering
        \vspace{0mm} %
        \includegraphics[width=\textwidth]{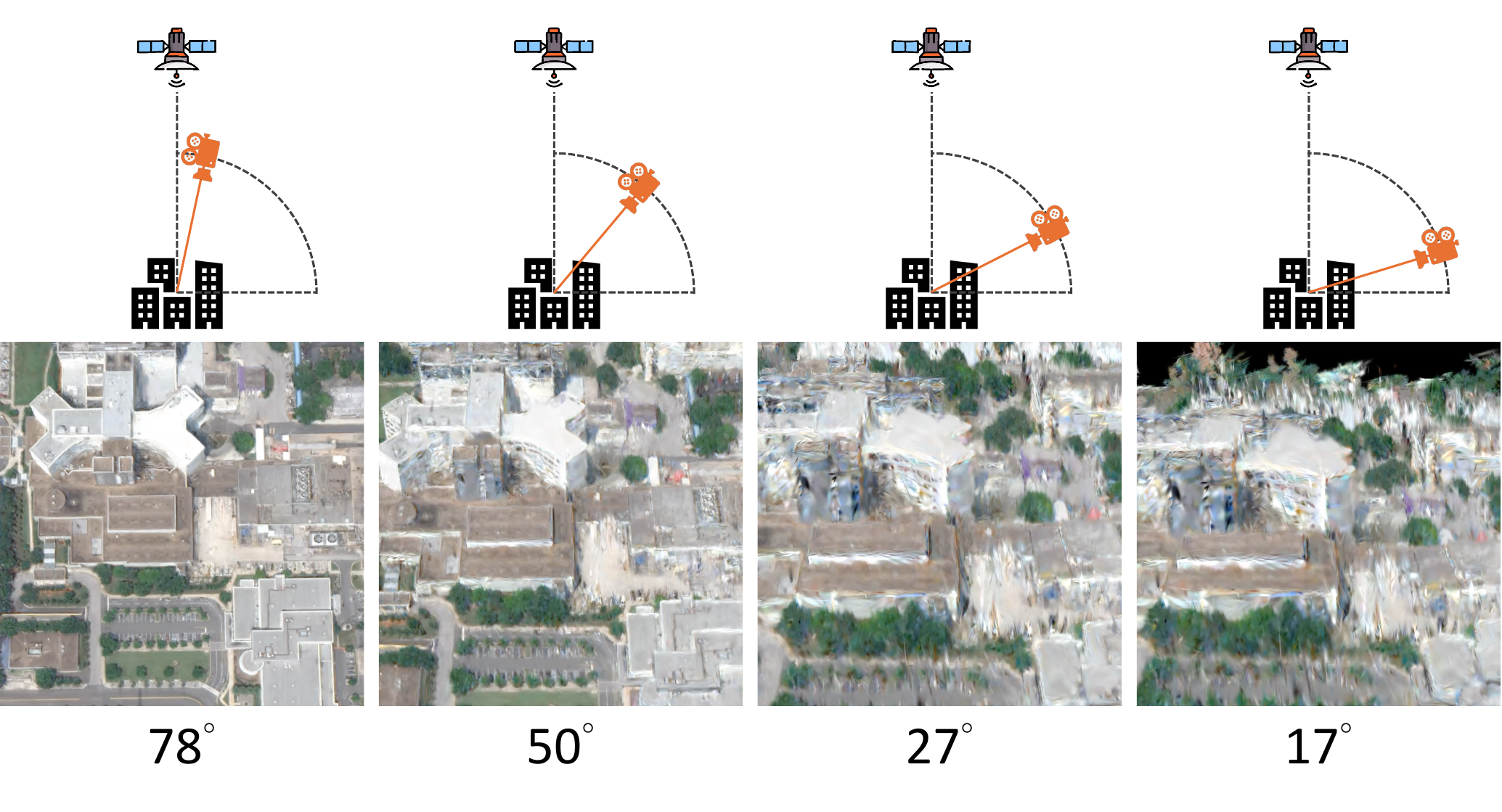}
        \caption{
        \textbf{The motivation of curriculum strategy.} Renderings of the initial 3D reconstruction from varied elevation angles reveal progressive degradation as the viewing angle decreases. 
        }
        \label{fig:curri_motivation}
    \end{minipage}
    \hfill
    \begin{minipage}[t]{0.58\textwidth}
    \vspace{0mm} %
        \centering
        \includegraphics[width=\textwidth]{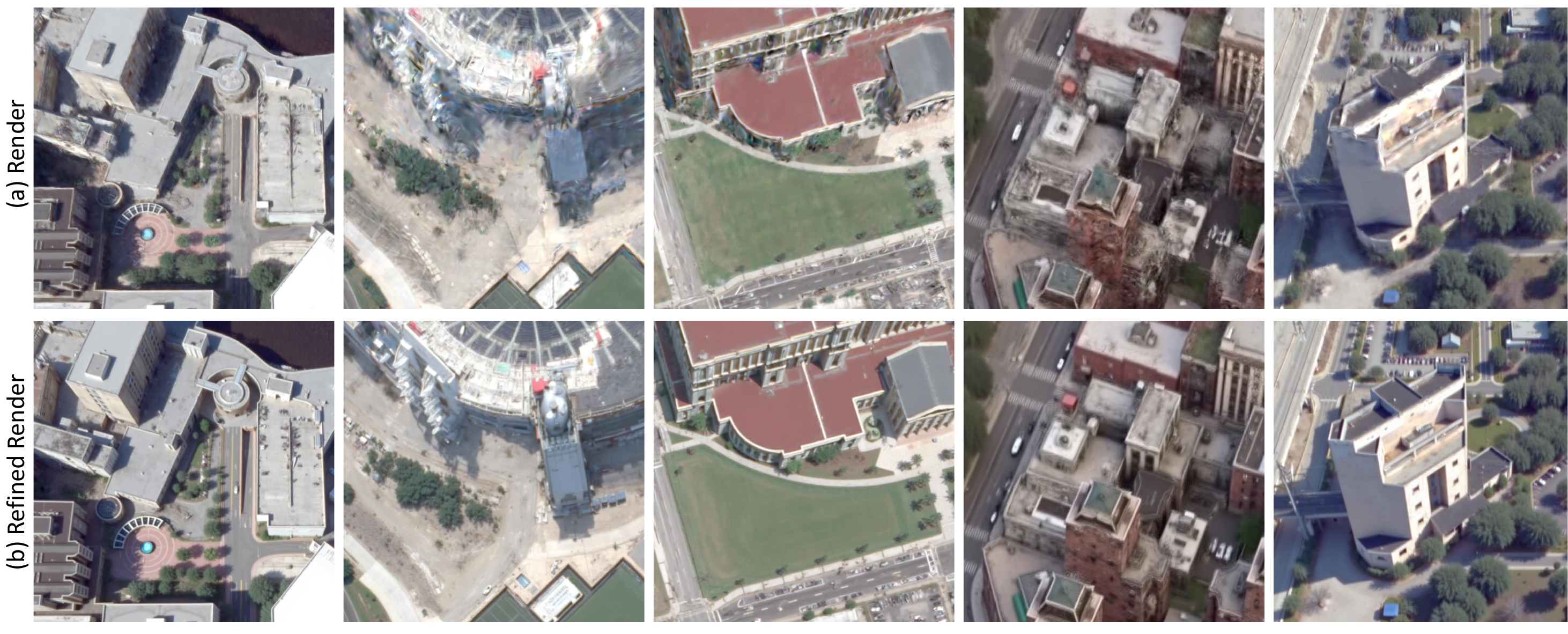}
        \caption{
        \textbf{Render refinement.} (a) Original 3DGS render with artifacts and blurry textures; (b)~Refined result showing enhanced geometry and texture quality.
        }
        \label{fig:flowedit}
    \end{minipage}
\end{figure}

\vspace{\subsecmargin}
\subsection{Synthesize via Curriculum-Based Iterative Datasets Update}
\label{sec:idu}

The iterative dataset update (IDU) technique~\cite{instructnerf2023,melaskyriazi2024im3d} repeatedly executes render-edit-update cycles across multiple episodes to progressively synthesize 3D scenes. Unlike previous methods that sample camera poses from original training views~\cite{instructnerf2023} or simple orbits~\cite{melaskyriazi2024im3d}, we introduce a curriculum-based refinement schedule over $N_e$ episodes that specifically addresses the geometric and visual limitations inherent to satellite imagery, producing structurally accurate and photorealistic reconstructions of occluded areas.

\vspace{\subsubsecmargin}
\subsubsection{Curriculum learning strategy.}
As illustrated in \Cref{fig:curri_motivation}, we observe that 3DGS trained from satellite imagery produces higher-quality renders at higher elevation angles but degenerates at lower elevation angles.
Leveraging this insight, we introduce a curriculum-based synthesis strategy, which progressively lowers viewpoints across optimization episodes. Specifically, we define $N_p$ look-at points $\{P_j\}_{j=1}^{N_p}$ uniformly placed throughout the scene and uniformly sample $N_v$ camera positions along orbital trajectories with controlled elevation angles and radii. Our iterative dataset update (IDU) process starts at higher elevations and progressively descends toward lower perspectives. This approach gradually reveals previously occluded regions, improving geometric detail and texture realism, as validated in our ablation studies (\Cref{sec:ablation}).

\vspace{\subsubsecmargin}
\subsubsection{Render refinement by text-to-image diffusion model.}
As illustrated in \Cref{fig:flowedit}(a), renderings from initial 3DGS contain blurry texture and artifacts. To address this, we leverage prompt-to-prompt editing with a pre-trained text-to-image diffusion model to synthesize disocclusion areas, remove artifacts, and enhance geometry. 
Prompt-to-prompt editing~\cite{hertz2022prompt} modifies input images, which are described by the source prompt, to align with the target prompt while preserving structural content. Although originally designed for real or synthetically generated images, we demonstrate its effectiveness for refining degraded 3DGS renders produced from satellite-view training, enabling high-quality appearance enhancement without disrupting the scene's satellite-consistent geometry.
Specifically, we use FlowEdit~\cite{kulikov2024flowedit} with the \texttt{FLUX.1 [dev]} diffusion model~\cite{blackforestlabs2024fluxweights}.
Our source prompts describe the original degraded features. The target prompts specify the desired high-quality attributes (details in Supplementary).
This approach significantly improves rendering quality. It yields sharper geometry, richer textures, and physically coherent shadows (see \Cref{fig:flowedit}). This strengthens the 3DGS training dataset for more accurate reconstructions.

\vspace{\subsubsecmargin}
\subsubsection{Multiple diffusion samples.}
While diffusion models effectively refine individual 3DGS renders, independently applying the diffusion model across viewpoints introduces inconsistencies. Furthermore, 3DGS is well known to suffer from overfitting on single views, as pointed out by CoR-GS~\cite{zhang2024cor}, causing artifacts when rendering from novel viewpoints. 

Ideally, the optimal denoising diffusion process should produce a distribution where all views maintain consistent 3D appearance. However, independent 2D denoising on each view does not preserve 3D consistency, resulting in a denoising trajectory distribution that is a superset of the optimal trajectories. Selecting a single denoising trajectory from this expanded distribution is unlikely to yield the optimal 3D-consistent result, leading to the artifacts observed in \Cref{fig:ablation_combined}(c).

To mitigate this, we synthesize $N_s$ independently refined samples per view, effectively sampling multiple trajectories from the denoising distribution. During optimization, the photometric loss $\mathcal{L}_\text{color}$ implicitly averages over these $N_s$ samples. Rather than committing to a single potentially suboptimal denoising path, this approach allows the 3DGS optimization to find a consensus representation that balances fidelity to individual samples while promoting geometric coherence across views. Ablation studies (\Cref{sec:ablation}) and \Cref{fig:ablation_combined}(c) confirm that this strategy successfully balances detail preservation with structural coherence.

\vspace{\subsubsecmargin}
\subsubsection{Iterative dataset update.}
Our curriculum-based Iterative Dataset Update (IDU), detailed in Algorithm~\ref{alg:3dgs_refinement}, optimizes the 3DGS over $N_e$ episodes. In each episode, we render curriculum-guided views and refine them using FlowEdit~\cite{kulikov2024flowedit} with specified prompts and strengths to generate a new training set. As the curriculum descends to lower altitudes, rendering quality steadily improves, particularly in previously occluded regions, as illustrated in \Cref{fig:progressive_refinement}. We provide detailed parameters in the Supplementary.

\begin{algorithm}[h]
\caption{3DGS Refinement via Iterative Dataset Updates}
\label{alg:3dgs_refinement}

\scriptsize 

\renewcommand{\alglinenumber}[1]{\scriptsize #1:} 

\begin{algorithmic}[1]
\Require $N_e$, $N_v$, $N_s$, $N_p$: Number of episodes, views per point, samples per view, and look-at points.
\Require $\{R_i\}_{i=1}^{N_e}$,$\{E_i\}_{i=1}^{N_e}$: Radius and elevation sequences; $\{P_j\}_{j=1}^{N_p}$: Target look-at points.
\Require $\Phi = (T_\text{src}, T_\text{tgt}, n_{\min}, n_{\max})$: FlowEdit parameters.
\Require $G$: Initial 3DGS from satellite-view training.
\Ensure $G'$: Refined 3DGS.
\State $G' \gets G$
\For{i = 1 to $N_e$}
    \State $\texttt{cam\_views} \gets \textproc{OrbitViews}(\{P\}, R_i, E_i, N_v)$ \Comment{Generate $N_p \times N_v$ views}
    \State $\texttt{render\_views} \gets \textproc{Render}(G', \texttt{cam\_views})$ \Comment{Render RGB images}
    \State $\texttt{refine\_views} \gets \textproc{Refine}(\texttt{render\_views}, \Phi, N_s)$ \Comment{Refine renders using FlowEdit}
    \State $G' \gets \textproc{Train}(G', \texttt{refine\_views})$ \Comment{Update 3DGS using refined views}
\EndFor
\State \textbf{return} $G'$
\end{algorithmic}
\end{algorithm}

\begin{figure*}[t]
\centering
\small
\setlength{\tabcolsep}{1pt}
\resizebox{\columnwidth}{!}{
\begin{tabular}{ccc}
  \includegraphics[width=0.36\textwidth, keepaspectratio]{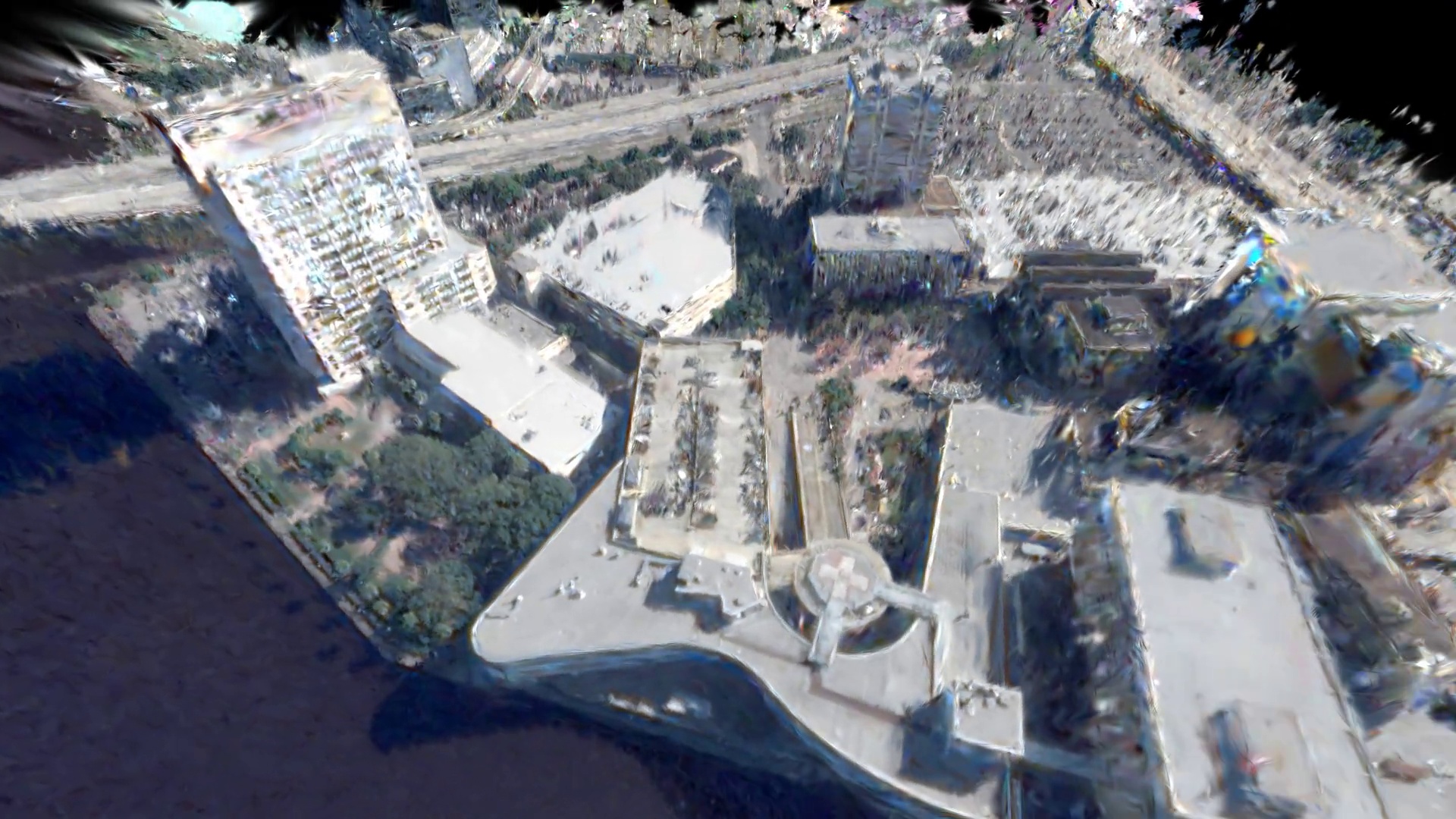} &
  \includegraphics[width=0.36\textwidth, keepaspectratio]{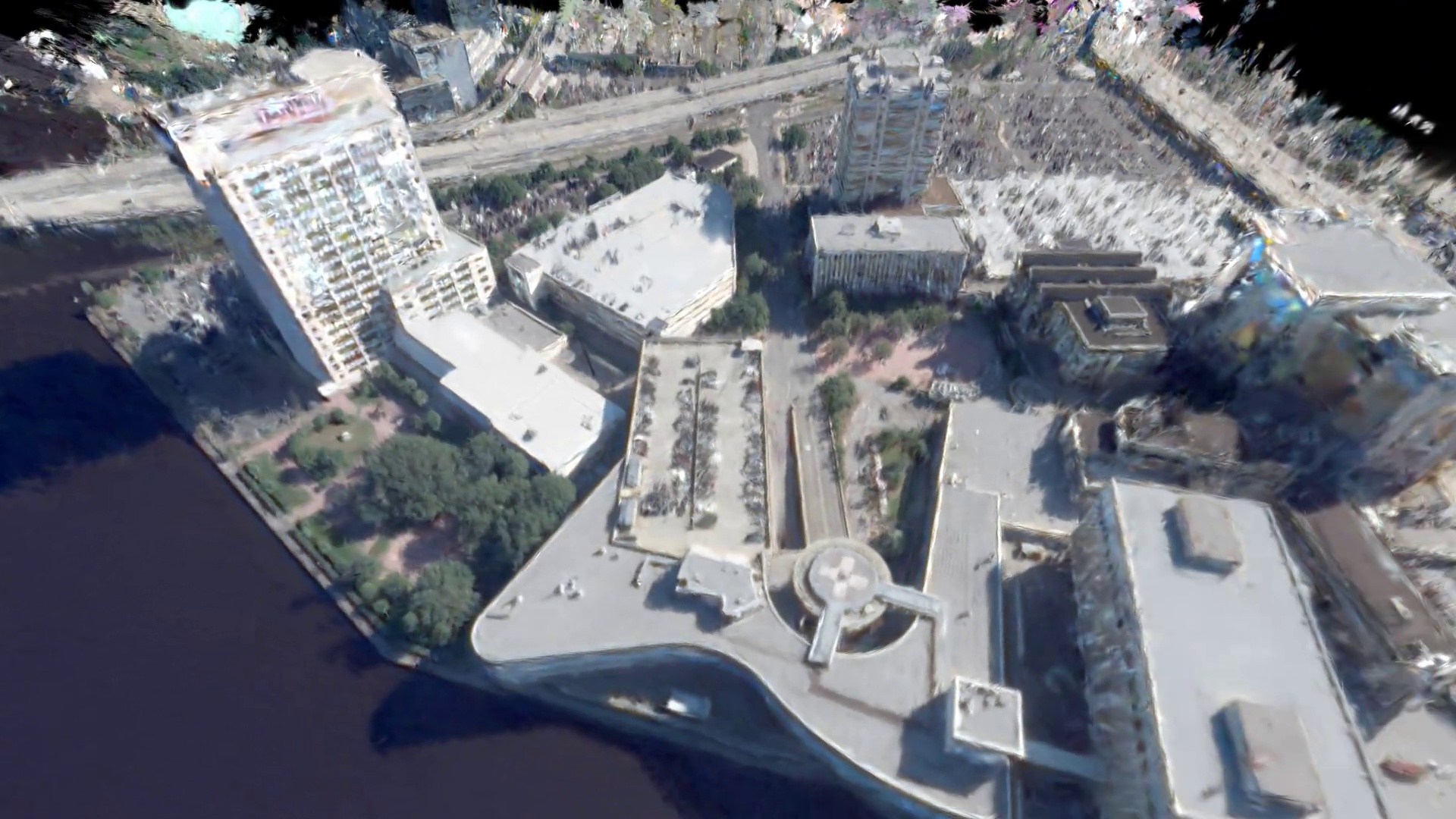} &
  \includegraphics[width=0.36\textwidth, keepaspectratio]{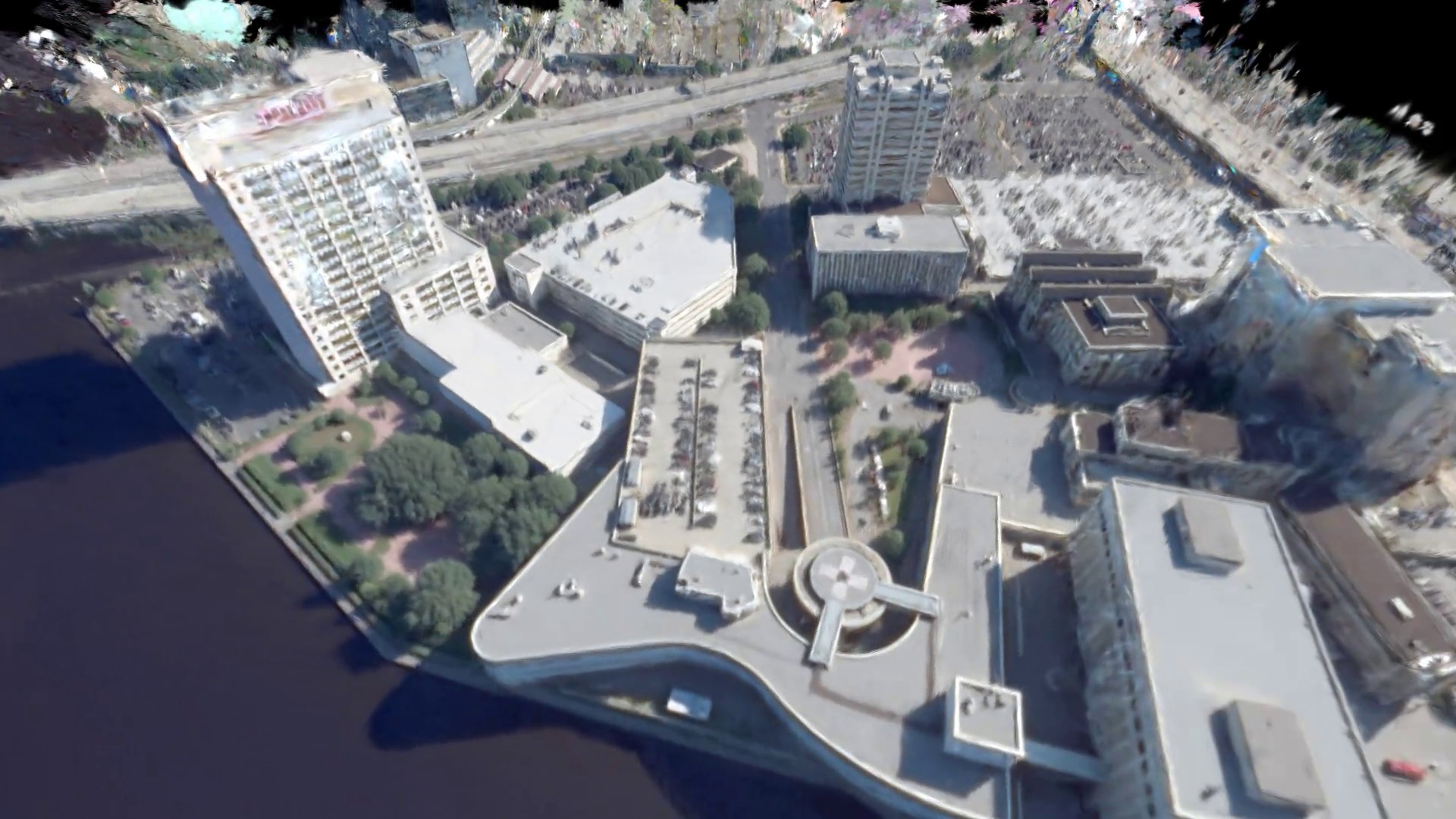} \\ [-0.3em]
  \scriptsize (a) After Reconstruction Stage & \scriptsize (b) After Episode 1 & \scriptsize (c) After Episode 2 \\[3pt] %

  \includegraphics[width=0.36\textwidth, keepaspectratio]{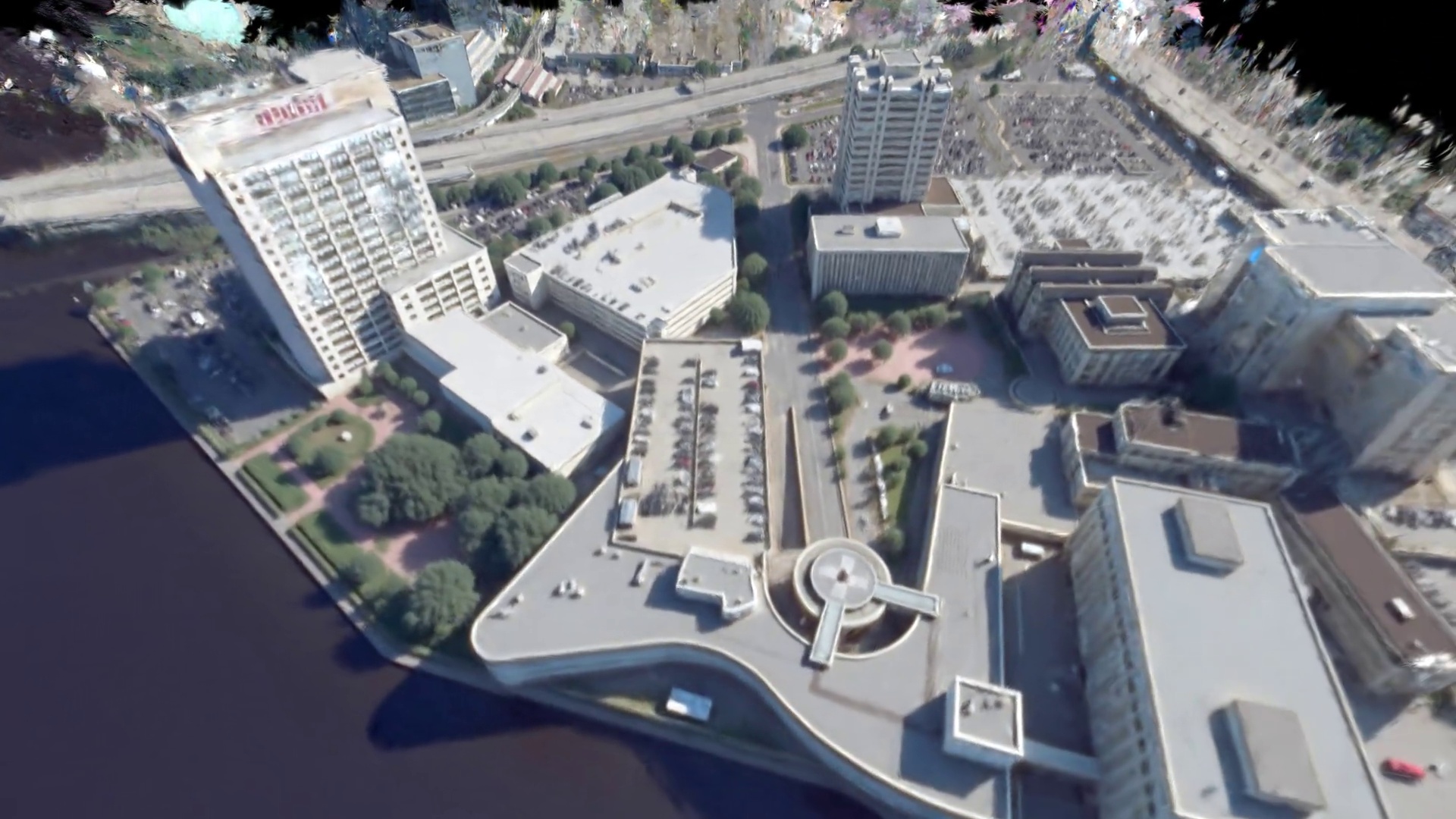} &
  \includegraphics[width=0.36\textwidth, keepaspectratio]{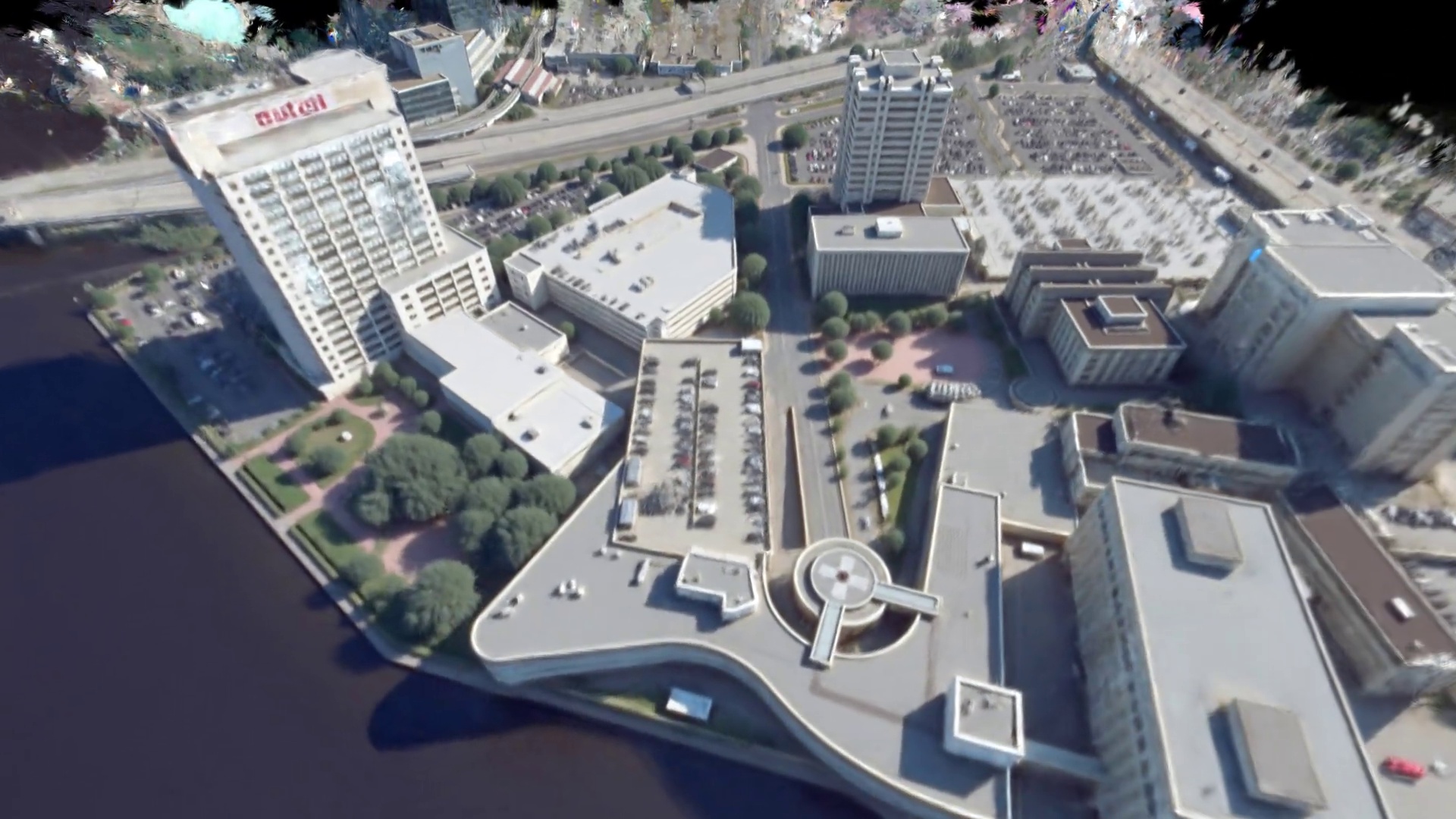} &
  \includegraphics[width=0.36\textwidth, keepaspectratio]{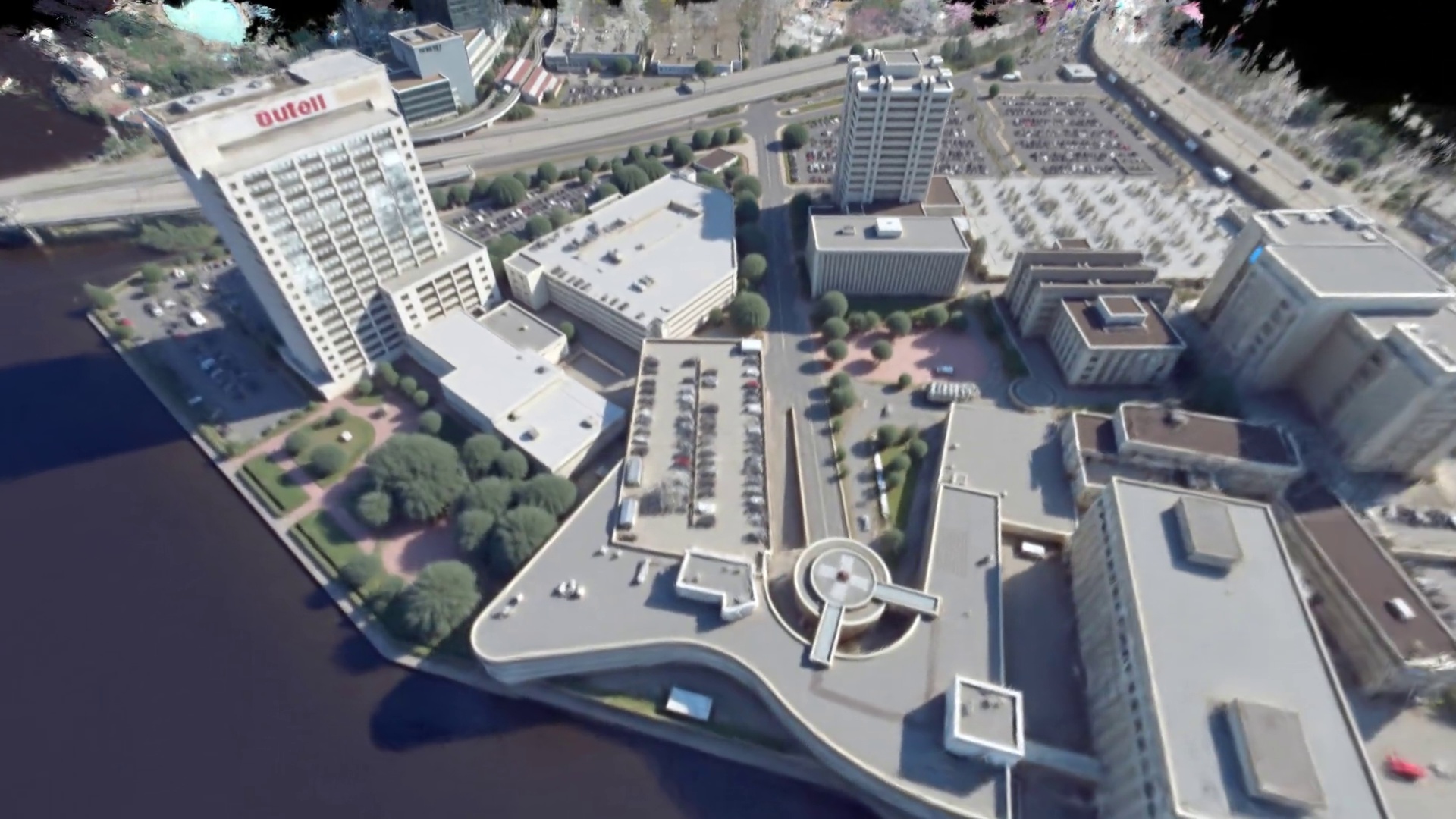} \\ [-0.3em]
  \scriptsize (d) After Episode 3 & \scriptsize (e) After Episode 4 & \scriptsize (f) After Episode 5 (Final) \\
\end{tabular}
}
\caption{\textbf{Visualization of progressive refinement.} This figure illustrates the step-by-step evolution of the synthesized 3D scene. Starting from the initial reconstruction state (a), the geometry and texture are progressively refined through successive stages of the iterative process (b-e), culminating in the final high-fidelity result (f).}
\label{fig:progressive_refinement}
\end{figure*}

\vspace{\subsubsecmargin}
\subsubsection{Optimization.}
For each episode $i$, we optimize the 3DGS using:
\begin{equation}
\small
    \mathcal{L}_{\text{IDU}}(G_{i-1}, \tilde{C}_i)= \mathcal{L}_{\text{color}} + \lambda_{\text{depth}} \mathcal{L}_{\text{depth}} \,\, ,
\end{equation}
where $G_{i-1}$ denotes the previous episode’s 3DGS model, and $\tilde{C}_i$ are the current refined images. We provide more implementation details in Supplementary.

\vspace{\secmargin}
\section{Experiments}
\label{sec:exp}
\vspace{\subsecmargin}

\subsubsection{Implementation details.}
The Reconstruction Stage runs for 30,000 iterations with $\lambda_{\text{D-SSIM}}=0.2$, $\lambda_{\text{op}}=10$, and $\lambda_{\text{depth}}=0.5$. The Synthesis Stage comprises $N_e=5$ episodes of 10,000 iterations each, with $N_v=6$ cameras and $N_s=2$ samples per look-at point. Training images are sampled 75\% from IDU-refined views and 25\% from original satellite views, preserving consistency with the input satellite imagery. All experiments run on a single RTX A6000 (48GB) GPU. Full details are in Supplementary.

\vspace{\subsubsecmargin}
\subsubsection{Datasets.} 
We evaluate our method on high-resolution RGB satellite imagery from two sources. First, the 2019 IEEE GRSS Data Fusion Contest (DFC2019)~\cite{c6tm-vw12-19} featuring WorldView-3 captures of Jacksonville, Florida (2048$\times$2048 pixels, 35 cm/pixel resolution). Camera parameters and sparse points are generated using SatelliteSfM~\cite{VisSat-2019}. We evaluate on four standard AOIs: JAX\_004, JAX\_068, JAX\_214, and JAX\_260, following Sat-NeRF~\cite{mari2022sat} and EOGS~\cite{savantaira2024eogs}. Second, for geographic diversity, we use the GoogleEarth dataset~\cite{xie2024citydreamer} (training data for CityDreamer~\cite{xie2024citydreamer} and GaussianCity~\cite{xie2025gaussiancity}) containing NYC scenes. We use four scenes (004, 010, 219, 336) with training views rendered at 80° elevation to approximate satellite conditions. Google Earth Studio (GES)~\cite{google_earth_studio} renders serve as ground truth for both datasets. See Supplementary for more detail.

\vspace{\subsubsecmargin}
\subsubsection{Baselines.}
As our method bridges satellite-based 3D reconstruction and city generation, we select baselines from both fields. For \emph{satellite reconstruction}, we compare with Sat-NeRF~\cite{mari2022sat} and EOGS~\cite{savantaira2024eogs} on DFC2019 (they require RPC input unavailable in GoogleEarth), plus Mip-Splatting~\cite{yu2024mip} (with our appearance modeling enabled on DFC2019) and CoR-GS~\cite{zhang2024cor} on both datasets. For \emph{city generation}, we compare with CityDreamer~\cite{xie2024citydreamer} and GaussianCity~\cite{xie2025gaussiancity} on GoogleEarth dataset (their training dataset).\footnote{\scriptsize Many methods lack available code or models: Sat2Scene~\cite{li2024sat2scene} (inference only), Sat2Vid~\cite{li2021sat2vidstreetviewpanoramicvideo}, EO-NeRF~\cite{Mari_2023_CVPR}, Sat-DN~\cite{liu2025satdnimplicitsurfacereconstruction}, SatelliteRF~\cite{app14072729}, Sat-Mesh~\cite{rs15174297}, CrossViewDiff~\cite{li2024crossviewdiffcrossviewdiffusionmodel}, SkySplat~\cite{huang2025skysplat}, MagicCity~\cite{yao2025magiccity}, Sat3DGen~\cite{qian2026sat3dgen}, Sat2City~\cite{hua2025sat2city}, and Sat2RealCity~\cite{kang2025sat2realcity}.} We use official implementations with default settings.

\vspace{\subsubsecmargin}
\subsubsection{Evaluation metrics.} 
Skyfall-GS primarily aims to synthesize and enhance the invisible or low-coverage regions from the satellite view; therefore, our evaluation focuses on quality and diversity assessment metrics.
We report $\text{FID}_\text{CLIP}$~\cite{Kynkaanniemi2022} and CMMD~\cite{jayasumana2024rethinking} which use the CLIP~\cite{radford2021learning} backbone, as InceptionV3~\cite{szegedy2016rethinking} used in classic FID~\cite{heusel2017gans} and KID~\cite{binkowski2018demystifying} is unsuitable for modern generative tasks.
We complement these with user studies for perceptual quality assessment.
We also report pixel-aligned metrics (PSNR~\cite{huynh2008scope}, SSIM~\cite{wang2004image}, LPIPS~\cite{zhang2018unreasonable}) as secondary references. However, these metrics are unsuitable for generative tasks, as the synthetic elements in the obscured regions cannot match the invisible ground-truth.
Moreover, on DFC2019, systematic illumination and color gaps between WorldView-3 imagery and GES references make pixel-level scores unreliable; we therefore report DFC2019 pixel metrics per scene in the supplementary material.

\vspace{\subsecmargin}
\subsection{Comparisons with Baselines}
\label{sec:compare}

\begin{table*}[t]
\begin{minipage}[t]{.45\textwidth}
\centering
\caption{\textbf{Quantitative comparison on DFC2019~\cite{c6tm-vw12-19}.} Distribution metrics computed against GES references; per-scene pixel-level metrics are in the supplementary material. \colorbox{red!25}{Red}, \colorbox{orange!25}{orange}, and \colorbox{yellow!25}{yellow} indicate the best, second, and third best.}
\label{tab:quantitative_results}
\resizebox{\textwidth}{!}{%
\begin{threeparttable}
\begin{tabular}{lcc}
\toprule
Methods & $\text{FID}_{\text{CLIP}}\,\downarrow$ & CMMD$\,\downarrow$ \\
\midrule
\multicolumn{3}{l}{\textit{3D Reconstruction}} \\
\enskip Sat-NeRF~\cite{mari2022sat} & \trd{86.52} & \snd{4.788} \\
\enskip EOGS~\cite{savantaira2024eogs} & 87.67 & \trd{5.291} \\
\enskip CoR-GS~\cite{zhang2024cor} & \snd{84.95} & 5.692 \\
\enskip Mip-Splatting$^\dagger$~\cite{yu2024mip} & 86.72 & 5.404 \\
\midrule
\multicolumn{3}{l}{\textit{Our Approach}} \\
\enskip Ours & \best{27.03} & \best{2.110} \\
\bottomrule
\end{tabular}
\begin{tablenotes}
\footnotesize
\item[$\dagger$]Enhanced with our appearance modeling.
\end{tablenotes}
\end{threeparttable}%
}
\end{minipage}
\hfill
\begin{minipage}[t]{.52\textwidth}
\centering
\caption{\textbf{Quantitative comparison on the GoogleEarth dataset~\cite{xie2024citydreamer}.} Metrics are computed against GES reference frames. \colorbox{red!25}{Red}, \colorbox{orange!25}{orange}, and \colorbox{yellow!25}{yellow} indicate the best, second best, and third best results, respectively.}
\label{tab:quantitative_results_nyc}
\resizebox{\textwidth}{!}{%
\begin{tabular}{lccccc}
\toprule
& \multicolumn{2}{c}{Distribution Metrics} & \multicolumn{3}{c}{Pixel-level Metrics} \\
\cmidrule(lr){2-3} \cmidrule(lr){4-6}
Methods & $\text{FID}_{\text{CLIP}}\,\downarrow$ & CMMD$\,\downarrow$ & PSNR$\,\uparrow$ & SSIM$\,\uparrow$ & LPIPS$\,\downarrow$ \\
\midrule
\multicolumn{6}{l}{\textit{City Generation}} \\
\enskip CityDreamer~\cite{xie2024citydreamer} & 36.66 & 4.200 & 12.58 & 0.267 & 0.558 \\
\enskip GaussianCity~\cite{xie2025gaussiancity} & 28.76 & \trd{2.915} & \trd{13.41} & \trd{0.291} & 0.540 \\
\midrule
\multicolumn{6}{l}{\textit{3D Reconstruction}} \\
\enskip CoR-GS~\cite{zhang2024cor} & \trd{26.35} & 3.758 & 13.35 & \snd{0.299} & \trd{0.412} \\
\enskip Mip-Splatting~\cite{yu2024mip} & \snd{16.09} & \snd{2.086} & \snd{14.13} & \best{0.302} & \best{0.379} \\
\midrule
\multicolumn{6}{l}{\textit{Our Approach}} \\
\enskip Ours & \best{10.29} & \best{1.959} & \best{14.42} & \best{0.302} & \snd{0.393} \\
\bottomrule
\end{tabular}%
}
\end{minipage}
\end{table*}

\subsubsection{Quantitative comparison.}
We evaluate against satellite reconstruction and city generation baselines by dividing rendered frames into 144 patches ($512 \times 512$ pixels). Reference frames are extracted from GES at $17^\circ$ elevation for DFC2019 (30 frames/AOI, 4,320 total images) and $45^\circ$ for GoogleEarth (24 frames/scene, 3,456 total images); all methods are then evaluated on matching videos generated with identical camera parameters. As shown in \Cref{tab:quantitative_results,tab:quantitative_results_nyc}, our method substantially outperforms all baselines on the distribution metrics ($\text{FID}_\text{CLIP}$ and CMMD) that best reflect generative quality, and remains competitive on pixel-level metrics (Mip-Splatting attains a slightly better average LPIPS on GoogleEarth), as further corroborated by the qualitative and user studies below. On DFC2019 pixel metrics (supplementary material), CoR-GS occasionally achieves a higher SSIM, which we attribute to its overly smooth reconstructions inflating SSIM given its known insensitivity to blurring.

\begin{figure}[t]
    \centering
    \includegraphics[width=1.0\linewidth]{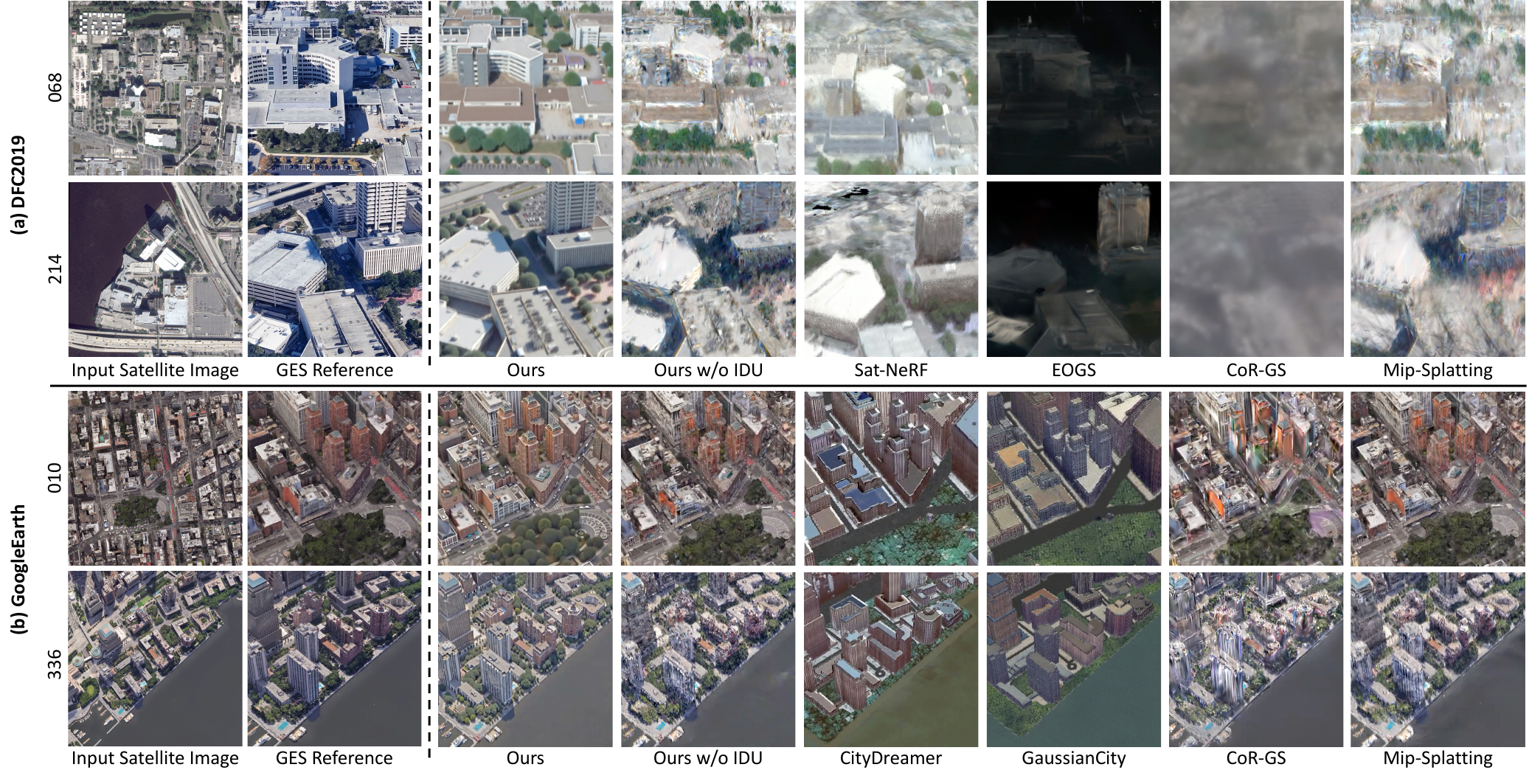}
    \caption{\textbf{Qualitative comparison on (a)~DFC2019 and (b)~GoogleEarth datasets.} The leftmost column shows a representative input satellite image. Our method consistently surpasses baselines in geometric accuracy and texture quality at low-altitude novel views. On~(a), our synthesis exhibit sharp building contours and detailed rooftops, whereas Sat-NeRF~\cite{mari2022sat} produces severe fog, EOGS~\cite{savantaira2024eogs} yields excessive darkness, CoR-GS~\cite{zhang2024cor} suffers from heavy blurring, and Mip-Splatting~\cite{yu2024mip} exhibits prominent floaters and artifacts. On~(b), CityDreamer~\cite{xie2024citydreamer} and GaussianCity~\cite{xie2025gaussiancity} over-simplify building geometry and miss scene-specific details, CoR-GS~\cite{zhang2024cor} and Mip-Splatting~\cite{yu2024mip} produce blurry facades, while our method delivers sharper geometry, richer facade details, and correctly recovers distinctive features such as the red pavement in scene~010.}
    \label{fig:qualitative}
\end{figure}

\subsubsection{Qualitative comparison.}
\Cref{fig:qualitative}(a) compares our method on DFC2019 against Sat-NeRF~\cite{mari2022sat}, EOGS~\cite{savantaira2024eogs}, CoR-GS~\cite{zhang2024cor}, and Mip-Splatting~\cite{yu2024mip}, all of which exhibit significant distortions and blurry textures at lower viewpoints. \Cref{fig:qualitative}(b) further includes CityDreamer~\cite{xie2024citydreamer} and GaussianCity~\cite{xie2025gaussiancity}, which over-simplify geometry and miss scene-specific details (\eg, the red pavement in scene 010). Furthermore, our approach yields sharper building contours, higher texture fidelity, and fewer artifacts across both datasets. Notice that Skyfall-GS also recovers the challenging facade details in occluded regions and complex structures (\eg, vegetation and bridges) that require high precision. Additional qualitative results are provided in Supplementary.

\begin{figure}[t]
\small
    \centering
    \begin{subfigure}[t]{0.49\linewidth}
        \centering
        \includegraphics[width=\linewidth]{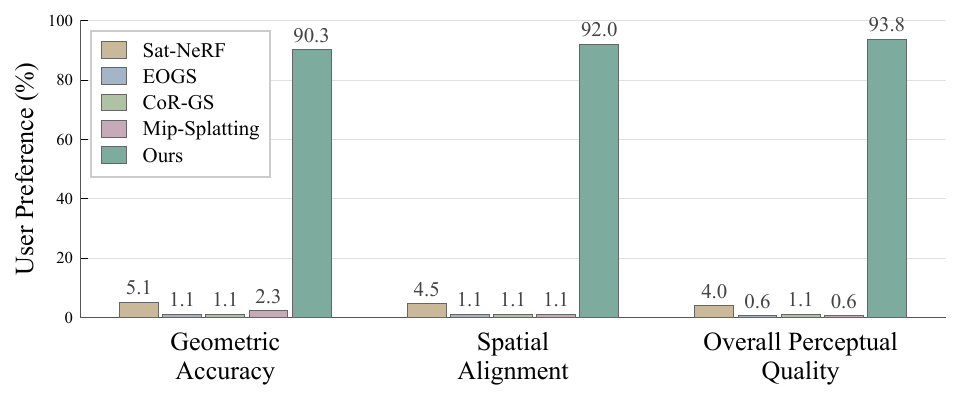}
        \caption{Compare on DFC2019 dataset.}
        \label{fig:user-study-1}
    \end{subfigure}
    \hfill
    \begin{subfigure}[t]{0.49\linewidth}
        \centering
        \includegraphics[width=\linewidth]{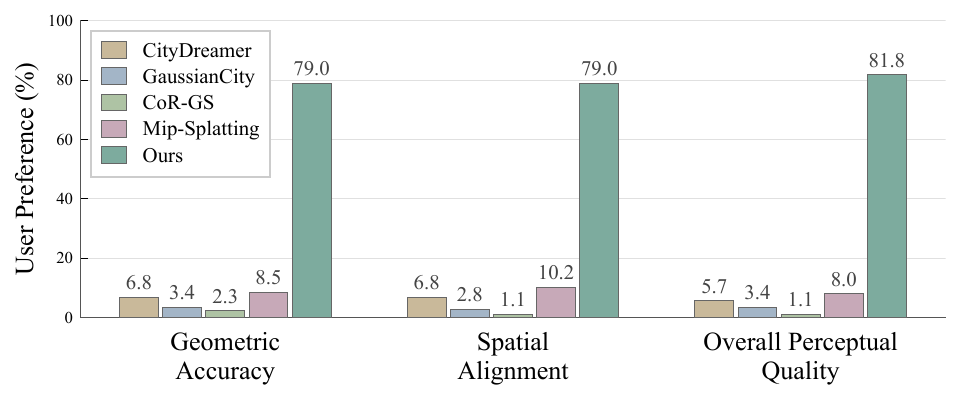}
        \caption{Compare on GoogleEarth dataset.}
        \label{fig:user-study-2}
    \end{subfigure}
    \caption{\textbf{User study results.} Our method consistently outperforms Sat-NeRF~\cite{mari2022sat}, EOGS~\cite{savantaira2024eogs}, CoR-GS~\cite{zhang2024cor}, Mip-Splatting~\cite{yu2024mip}, CityDreamer~\cite{xie2024citydreamer} and GaussianCity~\cite{xie2025gaussiancity}, achieving particularly high scores in geometric accuracy and overall perceptual quality. (a) details the comparison on the DFC2019 dataset~\cite{c6tm-vw12-19}, while (b) details the comparison on the GoogleEarth dataset~\cite{xie2024citydreamer}.}
    \label{fig:user_study}
\end{figure}

\vspace{\subsubsecmargin}
\subsubsection{User studies.}
We conducted two user studies with 44 participants each, evaluating geometric accuracy, spatial alignment, and overall perceptual quality across 4 scenes per study. In each study, participants were presented with side-by-side renderings and asked to select the result that best matched each criterion. In the first study, participants compared Sat-NeRF~\cite{mari2022sat}, EOGS~\cite{savantaira2024eogs}, CoR-GS~\cite{zhang2024cor}, Mip-Splatting~\cite{yu2024mip}, and our approach; in the second, CityDreamer~\cite{xie2024citydreamer} and GaussianCity~\cite{xie2025gaussiancity} replaced Sat-NeRF and EOGS (see Supplementary for full survey details). As shown in \Cref{fig:user_study}, our method achieves dominant win rates of $\approx$90--94\% on DFC2019 and $\approx$79--82\% on GoogleEarth across all three criteria, confirming strong and consistent human preference for our approach over all baselines in both \emph{satellite reconstruction} and \emph{city generation} settings.

\begin{figure}[t]
    \centering
    \includegraphics[width=\linewidth]{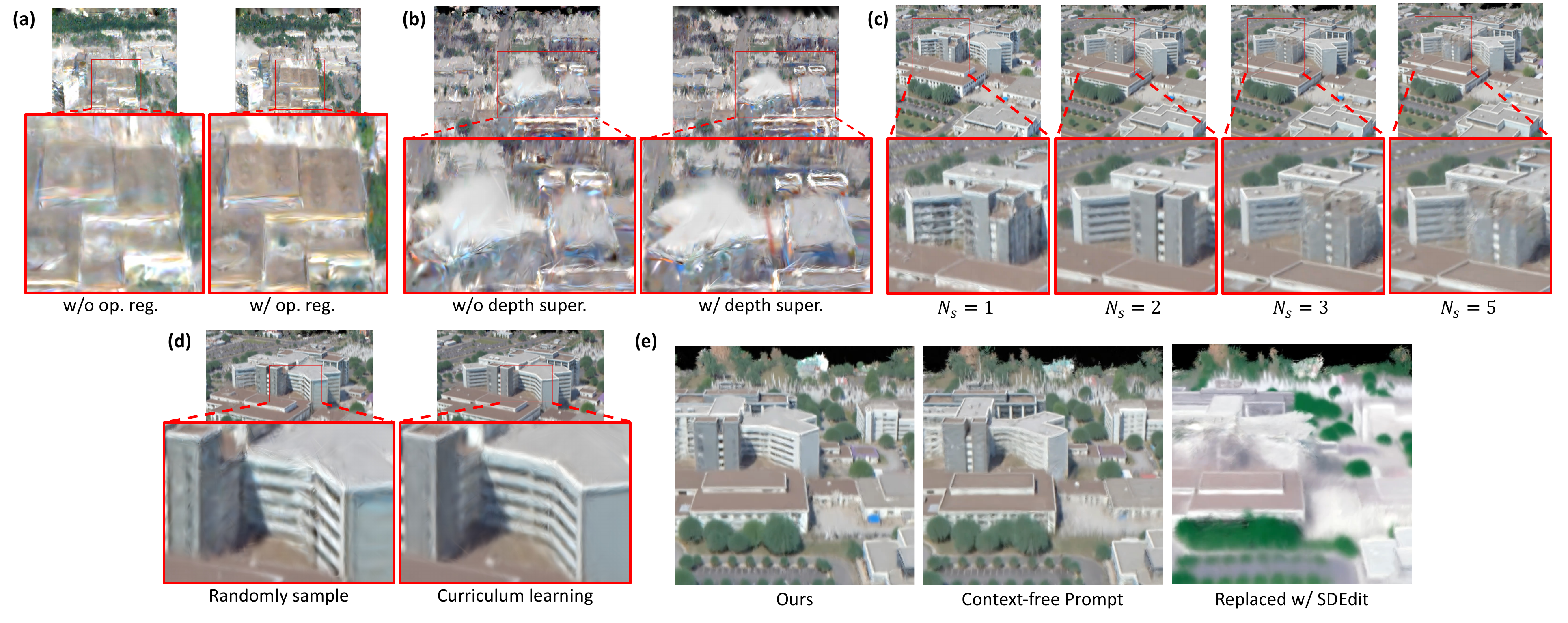}
    \caption{\textbf{Satellite-view training and IDU refinement ablation.}
    (a) Opacity regularization reduces floating artifacts and yields denser reconstructions. (b) Pseudo-camera depth supervision improves geometry in planar, texture-less regions such as rooftops and roads. (c) Multiple diffusion samples per view reduce high-frequency geometric noise and enhance texture consistency, $N_s=2$ achieves the optimal results. (d) Curriculum learning progressively introduces challenging views, significantly improving geometric coherence in occluded regions over random sampling. (e) A context-free prompt causes only minor degradation in facade details, demonstrating robustness to prompts. Replacing our refinement with SDEdit leads to severe quality degradation, as noising-denoising fails to synthesize details while preserving satellite-defined geometry.
    }
    \label{fig:ablation_combined}
\end{figure}

\vspace{\subsecmargin}
\subsection{Ablation Studies} 
\label{sec:ablation}

We conduct ablation studies on the JAX\_068 AOI.

\vspace{\subsubsecmargin}
\subsubsection{Ablation on the reconstruction stage.}
We ablate appearance modeling, opacity regularization, and pseudo-camera depth supervision (see \Cref{tab:sat-view-ablation} and Figure~\ref{fig:ablation_combined}). For this ablation, we evaluate at higher elevation angles to assess render quality during the IDU process, rather than at the final low-altitude viewpoints. Appearance modeling is crucial for multi-date convergence, opacity regularization removes floating artifacts (\Cref{fig:ablation_combined}(a)), and depth supervision sharpens geometry in planar regions (\Cref{fig:ablation_combined}(b)). Together, they yield the lowest $\text{FID}_\text{CLIP}$/CMMD scores. Additionally, we validate geometric accuracy using DFC2019 LiDAR data~\cite{c6tm-vw12-19} by unprojecting 3DGS depth renders into DSMs for comparison. Both opacity regularization and pseudo-depth supervision improve geometry, with their combination achieving the lowest MAE/RMSE.

\begin{table*}[t]
    \centering
    \begin{minipage}[b]{.58\textwidth}
        \centering
        \caption{
            \textbf{Ablation on the reconstruction stage.} Appearance modeling secures convergence. Opacity regularization and depth supervision enhance visual fidelity and geometric accuracy.
        }
        \label{tab:sat-view-ablation}
        \resizebox{1.0\textwidth}{!}{
            \setlength{\tabcolsep}{4pt} 
            \begin{tabular}{ccccccc}
                \toprule
                \multicolumn{3}{c}{Components} & \multicolumn{2}{c}{Perceptual Metrics} & \multicolumn{2}{c}{Geometric Metrics} \\
                \cmidrule(lr){1-3} \cmidrule(lr){4-5} \cmidrule(lr){6-7}
                \begin{tabular}[c]{@{}c@{}}App.\\Mod.\end{tabular} & 
                \begin{tabular}[c]{@{}c@{}}Op.\\Reg.\end{tabular} & 
                \begin{tabular}[c]{@{}c@{}}Depth\\Sup.\end{tabular} & 
                $\text{FID}_{\text{CLIP}}\downarrow$ & 
                CMMD $\downarrow$ &
                MAE (m)$\downarrow$ &
                RMSE (m)$\downarrow$ \\
                \midrule
                \textcolor{red}{\ding{55}} & \textcolor{red}{\ding{55}} & \textcolor{red}{\ding{55}} & \textit{Failed} & \textit{Failed} & \textit{Failed} & \textit{Failed} \\
                \textcolor{ForestGreen}{\checkmark} & \textcolor{red}{\ding{55}} & \textcolor{red}{\ding{55}} & 41.90 & 2.450 & 3.542 & 5.218 \\
                \textcolor{ForestGreen}{\checkmark} & \textcolor{ForestGreen}{\checkmark} & \textcolor{red}{\ding{55}} & 39.95 & 2.395 & 2.980 & 4.527 \\
                \textcolor{ForestGreen}{\checkmark} & \textcolor{ForestGreen}{\checkmark} & \textcolor{ForestGreen}{\checkmark} & \textbf{38.01} & \textbf{2.307} & \textbf{2.250} & \textbf{3.483} \\
                \bottomrule
            \end{tabular}
        }
    \end{minipage}%
    \hfill
    \begin{minipage}[b]{.39\textwidth}
        \centering
        \caption{
            \textbf{Ablation on the synthesis stage.}
            We evaluate sample counts ($N_s$), core components, and compare against baselines.
        }
        \label{tab:ablation-idu}
        \resizebox{1.0\textwidth}{!}{
            \setlength{\tabcolsep}{6pt}
            \begin{tabular}{lccc}
                \toprule
                Method Variation & $\text{FID}_{\text{CLIP}}\downarrow$ & CMMD $\downarrow$ & Time (h) \\
                \midrule
                \multicolumn{4}{l}{\textit{Multiple Samples ($N_s$)}} \\
                \enskip $N_s=1$ & 34.11 & 3.189 & 3.44 \\
                \enskip \textbf{Ours ($N_s=2$)} & \textbf{28.35} & 2.875 & 6.37 \\
                \enskip $N_s=3$ & 28.64 & 2.769 & 7.19 \\
                \enskip $N_s=5$ & 29.17 & \textbf{2.677} & 9.80 \\
                \midrule
                \multicolumn{4}{l}{\textit{Component Ablation}} \\
                \enskip w/o Curriculum (random) & 33.79 & 3.361 & - \\
                \enskip Reverse Curriculum & 53.03 & 4.170 & - \\
                \enskip w/ Context-free Pmt. & 30.78 & 2.981 & - \\
                \enskip Replaced w/ SDEdit & 64.74 & 4.138 & - \\
                \bottomrule
            \end{tabular}
        }
    \end{minipage}
\end{table*}

\vspace{\subsubsecmargin}
\subsubsection{Ablation on the synthesis stage.}
We isolate multi-sample diffusion and curriculum view progression. $N_s=2$ achieves optimal visual results (\Cref{fig:ablation_combined}(c)), while $N_s=5$ yields the lowest CMMD but requires $1.5\times$ longer training with marginal gains, so we adopt $N_s=2$. The ``w/o Curriculum'' variant in \Cref{tab:ablation-idu} replaces our high-to-low schedule with \emph{random} elevation sampling and is markedly worse at restoring occluded geometry (\Cref{fig:ablation_combined}(d)); reversing the schedule (low-to-high) is worse still, as it feeds the most degraded low-elevation renders to the diffusion model first, confirming that \emph{both} the curriculum and its high-to-low direction drive the gain. Replacing our refinement with SDEdit~\cite{meng2022sdedit} under the same schedule causes significant quality degradation, as noising-denoising cannot hallucinate plausible details while preserving satellite-defined geometry (\Cref{fig:ablation_combined}(e), \Cref{tab:ablation-idu}). A context-free prompt yields negligible difference, confirming robustness to prompts, with full prompts provided in Supplementary. The supplement further shows that multi-sample consensus improves cross-view consistency and that the iterative update is essential.

\vspace{\subsecmargin}
\subsection{Performance and Scalability}

\subsubsection{Training efficiency.}
We evaluate runtime on the JAX\_214 AOI using a single NVIDIA RTX A6000 (48GB). The full pipeline completes in $\sim$6h 45min, split across 1h 35min for reconstruction and 5h 10min for synthesis (per-episode breakdown in Supplementary). This per-scene cost is a one-time, offline expense on par with reconstruction baselines (Sat-NeRF/EO-NeRF report 10--20h; Mip-Splatting 1h 38min on our hardware) and amortizes across unlimited real-time renders, a reasonable trade-off that entirely bypasses time- and labor-intensive physical data collection.

\begin{figure}[t!]
    \centering
    \includegraphics[width=\linewidth]{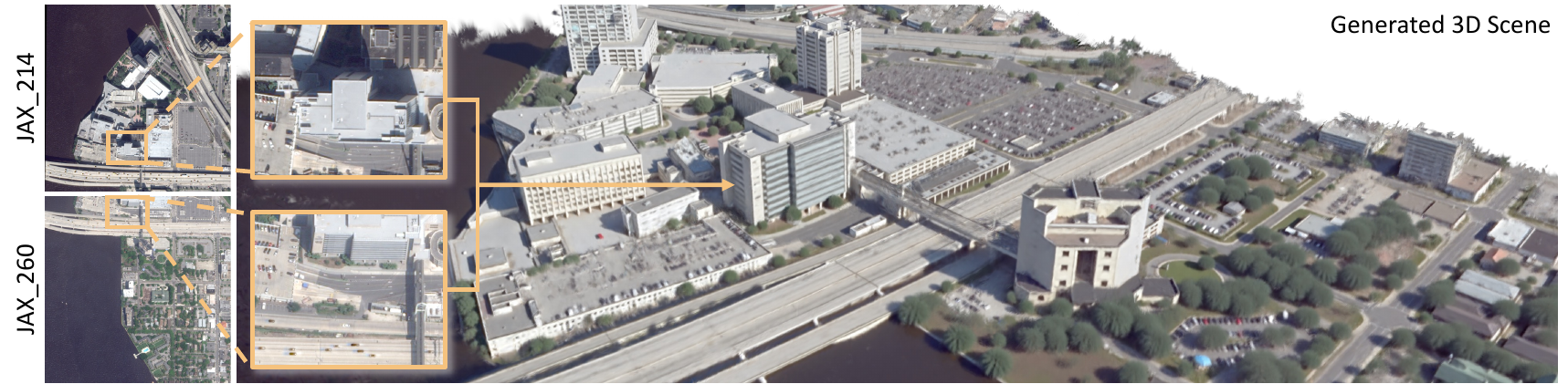}
    \caption{\textbf{Multi-block scalability on combined JAX\_214 and JAX\_260 AOIs.} Skyfall-GS jointly optimizes a seamless $\sim$1km $\times$ 512m scene from two adjacent satellite tiles; the inset shows a shared building with no stitching artifacts.}
    \label{fig:combined}
\end{figure}

\vspace{\subsubsecmargin}
\subsubsection{Multi-block scalability via combined imagery.}
We further combine two adjacent AOIs (JAX\_214, JAX\_260) overlapping along a highway into a single $\sim$1km $\times$ 512m dataset ($\sim$9h on one RTX A6000). As shown in \Cref{fig:combined}, structures shared across the boundary remain geometrically consistent with no stitching artifacts; on a boundary-spanning orbit, the merged model also outperforms either single-AOI model ($\sim$36--38\% lower $\text{FID}_\text{CLIP}$; Supplementary), covering boundary regions neither can.

\vspace{\secmargin}
\section{Conclusion}
\label{sec:conclusion}
\vspace{\secmargin}

We present Skyfall-GS, the first method to synthesize real-time, immersive, and freely navigable 3D urban scenes solely from multi-view satellite imagery, requiring no human intervention or domain-specific 3D training data. By combining 3D Gaussian Splatting with open-domain diffusion priors in a curriculum-based iterative refinement strategy, our method effectively addresses long-standing challenges, including limited parallax, illumination variations, and large-scale occlusions, and consistently outperforms state-of-the-art reconstruction and generation baselines. We hope this work paves the way for scalable, automated 3D urban scene creation, with promising future directions including city-wide scaling and dynamic scene modeling.

\vspace{\subsubsecmargin}
\subsubsection{Limitations.} Our fixed heuristic camera trajectory works well in general but may leave blind spots in complex geometries, occasionally causing minor artifacts at extreme street-level views. Strict pedestrian-level synthesis ($<\!10^\circ$, facade-close) is beyond our scope, as transient objects across the multi-date captures and GES reference distortion at low elevations preclude faithful close-up reconstruction and reliable evaluation. Finally, synthesizing facades requires off-nadir views; while most satellite imagery is off-nadir (\eg, WorldView spans $\sim\!\pm40^\circ$), purely nadir input remains an open challenge.

\section*{Acknowledgements}
This research was funded by the National Science and Technology Council, Taiwan, under Grants NSTC 112-2222-E-A49-004-MY2 and 113-2628-EA49-023-. The authors are grateful to Google, NVIDIA, and MediaTek Inc. for their generous donations. Yu-Lun Liu acknowledges the Yushan Young Fellow Program by the MOE in Taiwan.

\clearpage
\bibliographystyle{splncs04}
\bibliography{main}

@String(CVPR  = {IEEE Conf. Comput. Vis. Pattern Recog.})

@String(ICCV  = {Int. Conf. Comput. Vis.})

@String(ECCV  = {Eur. Conf. Comput. Vis.})

@String(NeurIPS = {Adv. Neural Inform. Process. Syst.})

@String(ICLR  = {Int. Conf. Learn. Represent.})

@String(CVPRW = {IEEE Conf. Comput. Vis. Pattern Recog. Worksh.})

@String(AAAI  = {AAAI})

@String(ICASSP=	{ICASSP})

@String(TMLR  = {Trans. Mach. Learn Res.})

@String(TOG   = {ACM Trans. Graph.})

@String(CVPR  = {CVPR})

@String(ICCV  = {ICCV})

@String(ECCV  = {ECCV})

@String(NeurIPS = {NeurIPS})

@String(ICLR  = {ICLR})

@String(CVPRW = {CVPRW})

@String(TMLR  = {TMLR})

@String(TOG   = {ACM TOG})

@article{kerbl20233d,
  title={{3D} Gaussian Splatting for Real-Time Radiance Field Rendering},
  author={Kerbl, Bernhard and Kopanas, Georgios and Leimk{\"u}hler, Thomas and Drettakis, George},
  journal={ACM TOG},
  year={2023}
}

@article{mildenhall2021nerf,
  title={{NeRF}: Representing Scenes as Neural Radiance Fields for View Synthesis},
  author={Mildenhall, Ben and Srinivasan, Pratul P and Tancik, Matthew and Barron, Jonathan T and Ramamoorthi, Ravi and Ng, Ren},
  journal={Communications of the ACM},
  year={2021}
}

@inproceedings{yu2024mip,
  title={{Mip-Splatting}: Alias-Free {3D} Gaussian Splatting},
  author={Yu, Zehao and Chen, Anpei and Huang, Binbin and Sattler, Torsten and Geiger, Andreas},
  booktitle={CVPR},
  year={2024}
}

@article{kerbl2024hierarchical,
  title={A Hierarchical {3D} Gaussian Representation for Real-Time Rendering of Very Large Datasets},
  author={Kerbl, Bernhard and Meuleman, Andreas and Kopanas, Georgios and Wimmer, Michael and Lanvin, Alexandre and Drettakis, George},
  journal={ACM TOG},
  year={2024}
}

@misc{xu2024wildgsrealtimenovelview,
      title={{Wild-GS}: Real-Time Novel View Synthesis from Unconstrained Photo Collections}, 
      author={Jiacong Xu and Yiqun Mei and Vishal M. Patel},
      year={2024},
      eprint={2406.10373},
      archivePrefix={arXiv},
      primaryClass={cs.CV},
      url={https://arxiv.org/abs/2406.10373}, 
}

@article{sabourgoli2024spotlesssplats,
    title={{SpotLessSplats}: Ignoring Distractors in {3D} Gaussian Splatting},
    author={Sabour, Sara and Goli, Lily and Kopanas, George and Matthews, Mark and Lagun, Dmitry and Guibas, Leonidas and Jacobson, Alec and Fleet, David J. and Tagliasacchi, Andrea},
    journal={arXiv preprint arXiv:2406.20055},
    year={2024}
}

@misc{wang2024wegsinthewildefficient3d,
      title={{WE-GS}: An In-the-Wild Efficient {3D} Gaussian Representation for Unconstrained Photo Collections}, 
      author={Yuze Wang and Junyi Wang and Yue Qi},
      year={2024},
      eprint={2406.02407},
      archivePrefix={arXiv},
      primaryClass={cs.CV},
      url={https://arxiv.org/abs/2406.02407}, 
}

@misc{dahmani2024swagsplattingwildimages,
      title={{SWAG}: Splatting in the Wild Images with Appearance-Conditioned Gaussians}, 
      author={Hiba Dahmani and Moussab Bennehar and Nathan Piasco and Luis Roldao and Dzmitry Tsishkou},
      year={2024},
      eprint={2403.10427},
      archivePrefix={arXiv},
      primaryClass={cs.CV},
      url={https://arxiv.org/abs/2403.10427}, 
}

@article{zhang2024gaussian,
  title={Gaussian in the Wild: {3D} Gaussian Splatting for Unconstrained Image Collections},
  author={Zhang, Dongbin and Wang, Chuming and Wang, Weitao and Li, Peihao and Qin, Minghan and Wang, Haoqian},
  journal={arXiv preprint arXiv:2403.15704},
  year={2024}
}

@article{kulhanek2024wildgaussians,
  title={{WildGaussians}: {3D} Gaussian Splatting in the Wild},
  author={Kulhanek, Jonas and Peng, Songyou and Kukelova, Zuzana and Pollefeys, Marc and Sattler, Torsten},
  journal={NeurIPS},
  year={2024}
}

@misc{lin2024vastgaussianvast3dgaussians,
      title={{VastGaussian}: Vast {3D} Gaussians for Large Scene Reconstruction}, 
      author={Jiaqi Lin and Zhihao Li and Xiao Tang and Jianzhuang Liu and Shiyong Liu and Jiayue Liu and Yangdi Lu and Xiaofei Wu and Songcen Xu and Youliang Yan and Wenming Yang},
      year={2024},
      eprint={2402.17427},
      archivePrefix={arXiv},
      primaryClass={cs.CV},
      url={https://arxiv.org/abs/2402.17427}, 
}

@inproceedings{liu2025citygaussian,
  title={{CityGaussian}: Real-Time High-Quality Large-Scale Scene Rendering with Gaussians},
  author={Liu, Yang and Luo, Chuanchen and Fan, Lue and Wang, Naiyan and Peng, Junran and Zhang, Zhaoxiang},
  booktitle={European Conference on Computer Vision},
  pages={265--282},
  year={2025},
  organization={Springer}
}

@misc{liu2024citygaussianv2,
      title={{CityGaussianV2}: Efficient and Geometrically Accurate Reconstruction for Large-Scale Scenes}, 
      author={Yang Liu and Chuanchen Luo and Zhongkai Mao and Junran Peng and Zhaoxiang Zhang},
      year={2024},
      eprint={2411.00771},
      archivePrefix={arXiv},
      primaryClass={cs.CV},
      url={https://arxiv.org/abs/2411.00771}, 
}

@inproceedings{lin2023infinicity,
   title={{InfiniCity}: Infinite-Scale City Synthesis},
   author={Lin, Chieh Hubert and Lee, Hsin-Ying and Menapace, Willi and Chai, Menglei and Siarohin, Aliaksandr and Yang, Ming-Hsuan and Tulyakov, Sergey},
   booktitle={Proceedings of the IEEE/CVF International Conference on Computer Vision},
   year={2023},
}

@misc{li2021sat2vidstreetviewpanoramicvideo,
      title={{Sat2Vid}: Street-View Panoramic Video Synthesis from a Single Satellite Image}, 
      author={Zuoyue Li and Zhenqiang Li and Zhaopeng Cui and Rongjun Qin and Marc Pollefeys and Martin R. Oswald},
      year={2021},
      eprint={2012.06628},
      archivePrefix={arXiv},
      primaryClass={cs.CV},
      url={https://arxiv.org/abs/2012.06628}, 
}

@InProceedings{li2024sat2scene,
    author    = {Li, Zuoyue and Li, Zhenqiang and Cui, Zhaopeng and Pollefeys, Marc and Oswald, Martin R.},
    title     = {{Sat2Scene}: {3D} Urban Scene Generation from Satellite Images with Diffusion},
    booktitle = {Proceedings of the IEEE/CVF Conference on Computer Vision and Pattern Recognition (CVPR)},
    month     = {June},
    year      = {2024},
    pages     = {7141-7150}
}

@inproceedings{deng2024streetscapes,
  title     = {Streetscapes: Large-Scale Consistent Street View Generation Using Autoregressive Video Diffusion},
  author    = {Deng, Boyang and Tucker, Richard and Li, Zhengqi
               and Guibas, Leonidas and Snavely, Noah and Wetzstein, Gordon},
  booktitle = {SIGGRAPH 2024 Conference Papers},
  year      = {2024}
}

@inproceedings{qian2023sat2density,
  author    = {Qian, Ming and Xiong, Jincheng and Xia, Gui-Song and Xue, Nan},
  title     = {Sat2Density: Faithful Density Learning from Satellite-Ground Image Pairs},
  booktitle   = {IEEE/CVF International Conference on Computer Vision (ICCV)},
  year      = {2023},
}

@misc{ze2025controllablesatellitetostreetviewsynthesisprecise,
      title={Controllable Satellite-to-Street-View Synthesis with Precise Pose Alignment and Zero-Shot Environmental Control}, 
      author={Xianghui Ze and Zhenbo Song and Qiwei Wang and Jianfeng Lu and Yujiao Shi},
      year={2025},
      eprint={2502.03498},
      archivePrefix={arXiv},
      primaryClass={eess.IV},
      url={https://arxiv.org/abs/2502.03498}, 
}

@misc{li2024crossviewdiffcrossviewdiffusionmodel,
      title={{CrossViewDiff}: A Cross-View Diffusion Model for Satellite-to-Street View Synthesis}, 
      author={Weijia Li and Jun He and Junyan Ye and Huaping Zhong and Zhimeng Zheng and Zilong Huang and Dahua Lin and Conghui He},
      year={2024},
      eprint={2408.14765},
      archivePrefix={arXiv},
      primaryClass={cs.CV},
      url={https://arxiv.org/abs/2408.14765}, 
}

@inproceedings{
      meng2022sdedit,
      title={{SDEdit}: Guided Image Synthesis and Editing with Stochastic Differential Equations},
      author={Chenlin Meng and Yutong He and Yang Song and Jiaming Song and Jiajun Wu and Jun-Yan Zhu and Stefano Ermon},
      booktitle={International Conference on Learning Representations},
      year={2022},
}

@misc{miyake2024negativepromptinversionfastimage,
      title={Negative-Prompt Inversion: Fast Image Inversion for Editing with Text-Guided Diffusion Models}, 
      author={Daiki Miyake and Akihiro Iohara and Yu Saito and Toshiyuki Tanaka},
      year={2024},
      eprint={2305.16807},
      archivePrefix={arXiv},
      primaryClass={cs.CV},
      url={https://arxiv.org/abs/2305.16807}, 
}

@article{mokady2022null,
  title={Null-Text Inversion for Editing Real Images Using Guided Diffusion Models},
  author={Mokady, Ron and Hertz, Amir and Aberman, Kfir and Pritch, Yael and Cohen-Or, Daniel},
  journal={arXiv preprint arXiv:2211.09794},
  year={2022}
}

@article{kulikov2024flowedit,
	title = {{FlowEdit}: Inversion-Free Text-Based Editing Using Pre-Trained Flow Models},
	author = {Kulikov, Vladimir and Kleiner, Matan and Huberman-Spiegelglas, Inbar and Michaeli, Tomer},
	journal = {arXiv preprint arXiv:2412.08629},
	year = {2024}
}

@inproceedings{instructnerf2023,
         author = {Haque, Ayaan and Tancik, Matthew and Efros, Alexei and Holynski, Aleksander and Kanazawa, Angjoo},
         title = {{Instruct-NeRF2NeRF}: Editing {3D} Scenes with Instructions},
         booktitle = {Proceedings of the IEEE/CVF International Conference on Computer Vision},
         year = {2023},
}

@misc{wu2025aurafusion360augmentedunseenregion,
      title={{AuraFusion360}: Augmented Unseen Region Alignment for Reference-Based 360${}^\circ$ Unbounded Scene Inpainting}, 
      author={Chung-Ho Wu and Yang-Jung Chen and Ying-Huan Chen and Jie-Ying Lee and Bo-Hsu Ke and Chun-Wei Tuan Mu and Yi-Chuan Huang and Chin-Yang Lin and Min-Hung Chen and Yen-Yu Lin and Yu-Lun Liu},
      year={2025},
      eprint={2502.05176},
      archivePrefix={arXiv},
      primaryClass={cs.CV},
      url={https://arxiv.org/abs/2502.05176}, 
}

@article{liu2023deceptive,
    title={{Deceptive-NeRF/3DGS}: Diffusion-Generated Pseudo-Observations for High-Quality Sparse-View Reconstruction},
    author={Liu, Xinhang and Chen, Jiaben and Kao, Shiu-hong and Tai, Yu-Wing and Tang, Chi-Keung},
    journal={arXiv preprint arXiv:2305.15171},
    year={2023}
}

@article{wu2023reconfusion,
    title={{ReconFusion}: {3D} Reconstruction with Diffusion Priors},
    author={Rundi Wu and Ben Mildenhall and Philipp Henzler and 
			Keunhong Park and Ruiqi Gao and Daniel Watson and 
			Pratul P. Srinivasan and Dor Verbin and Jonathan T. Barron 
			and Ben Poole and Aleksander Holynski},
    journal={arXiv preprint arXiv:2312.02981},
    year={2023}
}

@misc{chen2024vi3drmtowardsmeticulous3dreconstruction,
      title={{VI3DRM}: Towards Meticulous {3D} Reconstruction from Sparse Views via Photo-Realistic Novel View Synthesis}, 
      author={Hao Chen and Jiafu Wu and Ying Jin and Jinlong Peng and Xiaofeng Mao and Mingmin Chi and Mufeng Yao and Bo Peng and Jian Li and Yun Cao},
      year={2024},
      eprint={2409.08207},
      archivePrefix={arXiv},
      primaryClass={cs.CV},
      url={https://arxiv.org/abs/2409.08207}, 
}

@article{melaskyriazi2024im3d,
    title={{IM-3D}: Iterative Multiview Diffusion and Reconstruction for High-Quality {3D} Generation},
    author={Luke Melas-Kyriazi and Iro Laina and Christian Rupprecht and Natalia Neverova and Andrea Vedaldi and Oran Gafni and Filippos Kokkinos},
    journal={International Conference on Machine Learning},
    year={2024}
}

@article{chung2023luciddreamer,
  title={{LucidDreamer}: Domain-Free Generation of {3D} Gaussian Splatting Scenes},
  author={Chung, Jaeyoung and Lee, Suyoung and Nam, Hyeongjin and Lee, Jaerin and Lee, Kyoung Mu},
  journal={arXiv preprint arXiv:2311.13384},
  year={2023}
}

@inproceedings{VisSat-2019,
  title={Leveraging Vision Reconstruction Pipelines for Satellite Imagery},
  author={Zhang, Kai and Sun, Jin and Snavely, Noah},
  booktitle={IEEE International Conference on Computer Vision Workshops},
  year={2019}
}

@inproceedings{schoenberger2016sfm,
  author={Sch\"{o}nberger, Johannes Lutz and Frahm, Jan-Michael},
  title={Structure-from-Motion Revisited},
  booktitle={Conference on Computer Vision and Pattern Recognition (CVPR)},
  year={2016},
}

@InProceedings{Derksen_2021_CVPR,
    author    = {Derksen, Dawa and Izzo, Dario},
    title     = {Shadow Neural Radiance Fields for Multi-View Satellite Photogrammetry},
    booktitle = {Proceedings of the IEEE/CVF Conference on Computer Vision and Pattern Recognition (CVPR) Workshops},
    month     = {June},
    year      = {2021},
    pages     = {1152-1161}
}

@inproceedings{mari2022sat,
  title={{Sat-NeRF}: Learning Multi-View Satellite Photogrammetry with Transient Objects and Shadow Modeling Using {RPC} Cameras},
  author={Mar{\'\i}, Roger and Facciolo, Gabriele and Ehret, Thibaud},
  booktitle={2022 IEEE/CVF Conference on Computer Vision and Pattern Recognition Workshops (CVPRW)},
  pages={1310-1320},
  year={2022}
}

@InProceedings{Mari_2023_CVPR,
    author    = {Mar{\'\i}, Roger and Facciolo, Gabriele and Ehret, Thibaud},
    title     = {Multi-Date Earth Observation {NeRF}: The Detail Is in the Shadows},
    booktitle = {Proceedings of the IEEE/CVF Conference on Computer Vision and Pattern Recognition (CVPR) Workshops},
    month     = {June},
    year      = {2023},
    pages     = {2034-2044}
}

@article{GAO2023446,
title = {A General Deep Learning Based Framework for {3D} Reconstruction from Multi-View Stereo Satellite Images},
journal = {ISPRS Journal of Photogrammetry and Remote Sensing},
volume = {195},
pages = {446-461},
year = {2023},
issn = {0924-2716},
doi = {https://doi.org/10.1016/j.isprsjprs.2022.12.012},
url = {https://www.sciencedirect.com/science/article/pii/S0924271622003276},
author = {Jian Gao and Jin Liu and Shunping Ji},
}

@InProceedings{Leotta_2019_CVPR_Workshops,
author = {Leotta, Matthew J. and Long, Chengjiang and Jacquet, Bastien and Zins, Matthieu and Lipsa, Dan and Shan, Jie and Xu, Bo and Li, Zhixin and Zhang, Xu and Chang, Shih-Fu and Purri, Matthew and Xue, Jia and Dana, Kristin},
title = {Urban Semantic {3D} Reconstruction from Multiview Satellite Imagery},
booktitle = {Proceedings of the IEEE/CVF Conference on Computer Vision and Pattern Recognition (CVPR) Workshops},
month = {June},
year = {2019}
}

@Article{app14072729,
AUTHOR = {Zhou, Xin and Wang, Yang and Lin, Daoyu and Cao, Zehao and Li, Biqing and Liu, Junyi},
TITLE = {{SatelliteRF}: Accelerating {3D} Reconstruction in Multi-View Satellite Images with Efficient Neural Radiance Fields},
JOURNAL = {Applied Sciences},
VOLUME = {14},
YEAR = {2024},
NUMBER = {7},
ARTICLE-NUMBER = {2729},
URL = {https://www.mdpi.com/2076-3417/14/7/2729},
ISSN = {2076-3417},
DOI = {10.3390/app14072729}
}

@misc{liu2025satdnimplicitsurfacereconstruction,
      title={{Sat-DN}: Implicit Surface Reconstruction from Multi-View Satellite Images with Depth and Normal Supervision}, 
      author={Tianle Liu and Shuangming Zhao and Wanshou Jiang and Bingxuan Guo},
      year={2025},
      eprint={2502.08352},
      archivePrefix={arXiv},
      primaryClass={cs.CV},
      url={https://arxiv.org/abs/2502.08352}, 
}

@Article{rs15174297,
AUTHOR = {Qu, Yingjie and Deng, Fei},
TITLE = {{Sat-Mesh}: Learning Neural Implicit Surfaces for Multi-View Satellite Reconstruction},
JOURNAL = {Remote Sensing},
VOLUME = {15},
YEAR = {2023},
NUMBER = {17},
ARTICLE-NUMBER = {4297},
URL = {https://www.mdpi.com/2072-4292/15/17/4297},
ISSN = {2072-4292},
DOI = {10.3390/rs15174297}
}

@misc{c6tm-vw12-19,
doi = {10.21227/c6tm-vw12},
url = {https://dx.doi.org/10.21227/c6tm-vw12},
author = {Le Saux, Bertrand and Yokoya, Naoto and Hänsch, Ronny and Brown, Myron},
publisher = {IEEE Dataport},
title = {Data Fusion Contest 2019 ({DFC2019})},
year = {2019} }

@article{huynh2008scope,
  title={Scope of Validity of {PSNR} in Image/Video Quality Assessment},
  author={Huynh-Thu, Quan and Ghanbari, Mohammed},
  journal={Electronics Letters},
  volume={44},
  number={13},
  pages={800--801},
  year={2008},
  publisher={IET}
}

@article{wang2004image,
  title={Image Quality Assessment: From Error Visibility to Structural Similarity},
  author={Wang, Zhou and Bovik, Alan C and Sheikh, Hamid R and Simoncelli, Eero P},
  journal={IEEE Transactions on Image Processing},
  volume={13},
  number={4},
  pages={600--612},
  year={2004},
  publisher={IEEE}
}

@inproceedings{zhang2018unreasonable,
  title={The Unreasonable Effectiveness of Deep Features as a Perceptual Metric},
  author={Zhang, Richard and Isola, Phillip and Efros, Alexei A and Shechtman, Eli and Wang, Oliver},
  booktitle={Proceedings of the IEEE Conference on Computer Vision and Pattern Recognition},
  pages={586--595},
  year={2018}
}

@inproceedings{heusel2017gans,
  title={{GANs} Trained by a Two Time-Scale Update Rule Converge to a Local Nash Equilibrium},
  author={Heusel, Martin and Ramsauer, Hubert and Unterthiner, Thomas and Nessler, Bernhard and Hochreiter, Sepp},
  booktitle={Advances in Neural Information Processing Systems},
  pages={6626--6637},
  year={2017}
}

@inproceedings{binkowski2018demystifying,
  title={Demystifying {MMD GANs}},
  author={Binkowski, Mikołaj and Sutherland, Dougal J and Arbel, Michael and Gretton, Arthur},
  booktitle={International Conference on Learning Representations},
  year={2018}
}

@inproceedings{radford2021learning,
  title={Learning Transferable Visual Models from Natural Language Supervision},
  author={Radford, Alec and Kim, Jong Wook and Hallacy, Chris and Ramesh, Aditya and Goh, Gabriel and Agarwal, Sandhini and Sastry, Girish and Askell, Amanda and Mishkin, Pamela and Clark, Jack and others},
  booktitle={Proceedings of the 38th International Conference on Machine Learning},
  pages={8748--8763},
  year={2021},
  organization={PMLR}
}

@inproceedings{Kynkaanniemi2022,
  author    = {Tuomas Kynkäänniemi and
               Tero Karras and
               Miika Aittala and
               Timo Aila and
               Jaakko Lehtinen},
  title     = {The Role of {ImageNet} Classes in Fréchet Inception Distance},
  booktitle = {Proc. ICLR},
  year      = {2023},
}

@article{gao2024cat3d,
    title={{CAT3D}: Create Anything in {3D} with Multi-View Diffusion Models},
    author={Ruiqi Gao and Aleksander Holynski and Philipp Henzler and Arthur Brussee and Ricardo Martin-Brualla and Pratul P. Srinivasan and Jonathan T. Barron and Ben Poole},
    journal={Advances in Neural Information Processing Systems},
    year={2024}
}

@misc{wu2024cat4dcreate4dmultiview,
      title={{CAT4D}: Create Anything in {4D} with Multi-View Video Diffusion Models}, 
      author={Rundi Wu and Ruiqi Gao and Ben Poole and Alex Trevithick and Changxi Zheng and Jonathan T. Barron and Aleksander Holynski},
      year={2024},
      eprint={2411.18613},
      archivePrefix={arXiv},
      primaryClass={cs.CV},
      url={https://arxiv.org/abs/2411.18613}, 
}

@inproceedings{brooks2023instructpix2pix,
  title={{InstructPix2Pix}: Learning to Follow Image Editing Instructions},
  author={Brooks, Tim and Holynski, Aleksander and Efros, Alexei A},
  booktitle={Proceedings of the IEEE/CVF Conference on Computer Vision and Pattern Recognition},
  pages={18392--18402},
  year={2023}
}

@misc{wang2024moge,
    title={{MoGe}: Unlocking Accurate Monocular Geometry Estimation for Open-Domain Images with Optimal Training Supervision},
    author={Wang, Ruicheng and Xu, Sicheng and Dai, Cassie and Xiang, Jianfeng and Deng, Yu and Tong, Xin and Yang, Jiaolong},
    year={2024},
    eprint={2410.19115},
    archivePrefix={arXiv},
    primaryClass={cs.CV},
    url={https://arxiv.org/abs/2410.19115}, 
}

@misc{zhu2023FSGS, 
    title={{FSGS}: Real-Time Few-Shot View Synthesis Using Gaussian Splatting}, 
    author={Zehao Zhu and Zhiwen Fan and Yifan Jiang and Zhangyang Wang}, 
    year={2023},
    eprint={2312.00451},
    archivePrefix={arXiv},
    primaryClass={cs.CV} 
}

@inproceedings{zhang2024cor,
      title={{CoR-GS}: Sparse-View 3D Gaussian Splatting via Co-Regularization},
      author={Zhang, Jiawei and Li, Jiahe and Yu, Xiaohan and Huang, Lei and Gu, Lin and Zheng, Jin and Bai, Xiao},
      year={2024},
      booktitle={ECCV},
}

@article{li2024dngaussian,
    title={{DNGaussian}: Optimizing Sparse-View {3D} Gaussian Radiance Fields with Global-Local Depth Normalization}, 
    author={Jiahe Li and Jiawei Zhang and Xiao Bai and Jin Zheng and Xin Ning and Jun Zhou and Lin Gu},
    journal={arXiv preprint arXiv:2403.06912},
    year={2024}
}

@inproceedings{xie2024citydreamer,
  title     = {{CityDreamer}: Compositional Generative Model of Unbounded {3D} Cities},
  author    = {Xie, Haozhe and 
               Chen, Zhaoxi and 
               Hong, Fangzhou and 
               Liu, Ziwei},
  booktitle = {CVPR},
  year      = {2024}
}

@inproceedings{xie2025gaussiancity,
  author       = {Haozhe Xie and
                  Zhaoxi Chen and
                  Fangzhou Hong and
                  Ziwei Liu},
  title        = {Generative Gaussian Splatting for Unbounded {3D} City Generation},
  booktitle    = {CVPR},
  year         = {2025}
}

@article{xie2025citydreamer4d,
  title     = {{CityDreamer4D}: Compositional Generative Model of Unbounded {4D} Cities},
  author    = {Xie, Haozhe and 
               Chen, Zhaoxi and 
               Hong, Fangzhou and 
               Liu, Ziwei},
  journal   = {arXiv preprint arXiv:2501.08983},
  year      = {2025}
}

@misc{barron2021mipnerf,
      title={Mip-{NeRF}: A Multiscale Representation for Anti-Aliasing Neural Radiance Fields},
      author={Jonathan T. Barron and Ben Mildenhall and Matthew Tancik and Peter Hedman and Ricardo Martin-Brualla and Pratul P. Srinivasan},
      year={2021},
      eprint={2103.13415},
      archivePrefix={arXiv},
      primaryClass={cs.CV}
}

@article{mueller2022instant,
    author = {Thomas M\"uller and Alex Evans and Christoph Schied and Alexander Keller},
    title = {Instant Neural Graphics Primitives with a Multiresolution Hash Encoding},
    journal = {ACM Trans. Graph.},
    issue_date = {July 2022},
    volume = {41},
    number = {4},
    month = jul,
    year = {2022},
    pages = {102:1--102:15},
    articleno = {102},
    numpages = {15},
    url = {https://doi.org/10.1145/3528223.3530127},
    doi = {10.1145/3528223.3530127},
    publisher = {ACM},
    address = {New York, NY, USA},
}

@article{barron2022mipnerf360,
    title={Mip-{NeRF} 360: Unbounded Anti-Aliased Neural Radiance Fields},
    author={Jonathan T. Barron and Ben Mildenhall and 
            Dor Verbin and Pratul P. Srinivasan and Peter Hedman},
    journal={CVPR},
    year={2022}
}

@inproceedings{barron2023zipnerf,
    title={Zip-{NeRF}: Anti-Aliased Grid-Based Neural Radiance Fields},
    author={Jonathan T. Barron and Ben Mildenhall and 
            Dor Verbin and Pratul P. Srinivasan and Peter Hedman},
    booktitle={ICCV},
    year={2023}
}

@InProceedings{Rombach_2022_CVPR,
    author    = {Rombach, Robin and Blattmann, Andreas and Lorenz, Dominik and Esser, Patrick and Ommer, Bj\"orn},
    title     = {High-Resolution Image Synthesis with Latent Diffusion Models},
    booktitle = {Proceedings of the IEEE/CVF Conference on Computer Vision and Pattern Recognition (CVPR)},
    month     = {June},
    year      = {2022},
    pages     = {10684-10695}
}

@misc{flux2024,
    author={Black Forest Labs},
    title={{FLUX}},
    year={2024},
    howpublished={\url{https://github.com/black-forest-labs/flux}},
    note = {Accessed: 2025-02-28}
}

@misc{blackforestlabs2024fluxweights,
  author = {Black Forest Labs},
  title = {Official Weights of {FLUX.1} Dev},
  howpublished = {\url{https://huggingface.co/black-forest-labs/FLUX.1-dev}},
  year = {2024},
  note = {Accessed: 2025-02-28}
}

@inproceedings{mirzaei2023watchyoursteps,
  title={Watch Your Steps: Local Image and Scene Editing by Text Instructions}, 
  author={Ashkan Mirzaei and Tristan Aumentado-Armstrong and Marcus A. Brubaker and Jonathan Kelly and Alex Levinshtein and Konstantinos G. Derpanis and Igor Gilitschenski},
  year={2024},
  booktitle={ECCV},
}

@inproceedings{GaussianEditor,
  author = {Fang, Jiemin and Wang, Junjie and Zhang, Xiaopeng and Xie, Lingxi and Tian, Qi},
  title = {{GaussianEditor}: Editing {3D} Gaussians Delicately with Text Instructions},
  year = {2024},
  booktitle = {CVPR}
}

@inproceedings{signerf,
  author ={Dihlmann, Jan-Niklas and Engelhardt, Andreas and Lensch, Hendrik P.A.},
  title ={{SIGNeRF}: Scene Integrated Generation for Neural Radiance Fields},
  booktitle ={Proceedings of the IEEE/CVF Conference on Computer Vision and Pattern Recognition (CVPR)},
  year ={2024}
}

@inproceedings{weber2023nerfiller,
    title = {{NeRFiller}: Completing Scenes via Generative {3D} Inpainting},
    author = {Ethan Weber and Aleksander Holynski and Varun Jampani and Saurabh Saxena and
        Noah Snavely and Abhishek Kar and Angjoo Kanazawa},
    booktitle = {CVPR},
    year = {2024},
}

@inproceedings{gaussian_grouping,
    title={Gaussian Grouping: Segment and Edit Anything in {3D} Scenes},
    author={Ye, Mingqiao and Danelljan, Martin and Yu, Fisher and Ke, Lei},
    booktitle={ECCV},
    year={2024}
}

@article{hertz2022prompt,
  title={Prompt-to-Prompt Image Editing with Cross Attention Control},
  author={Hertz, Amir and Mokady, Ron and Tenenbaum, Jay and Aberman, Kfir and Pritch, Yael and Cohen-Or, Daniel},
  journal={arXiv preprint arXiv:2208.01626},
  year={2022}
}

@inproceedings{savantaira2024eogs,
      title={Gaussian Splatting for Efficient Satellite Image Photogrammetry},
      author={Savant Aira, Luca and Facciolo, Gabriele and Ehret, Thibaud},
      booktitle={Proceedings of the IEEE/CVF Conference on Computer Vision and Pattern Recognition},
      year={2025}
}

@misc{gao2024enhanced3durbanscene,
      title={Enhanced {3D} Urban Scene Reconstruction and Point Cloud Densification Using Gaussian Splatting and Google Earth Imagery}, 
      author={Kyle Gao and Dening Lu and Hongjie He and Linlin Xu and Jonathan Li},
      year={2024},
      eprint={2405.11021},
      archivePrefix={arXiv},
      primaryClass={cs.CV},
      url={https://arxiv.org/abs/2405.11021}, 
}

@article{gaussctrl2024,
author = {Wu, Jing and Bian, Jia-Wang and Li, Xinghui and Wang, Guangrun and Reid, Ian and Torr, Philip and Prisacariu, Victor},
title = {{GaussCtrl}: Multi-View Consistent Text-Driven {3D} Gaussian Splatting Editing},
journal = {ECCV},
year = {2024},
}

@misc{wang2025viewconsistent3deditinggaussian,
      title={View-Consistent {3D} Editing with Gaussian Splatting}, 
      author={Yuxuan Wang and Xuanyu Yi and Zike Wu and Na Zhao and Long Chen and Hanwang Zhang},
      year={2025},
      eprint={2403.11868},
      archivePrefix={arXiv},
      primaryClass={cs.GR},
      url={https://arxiv.org/abs/2403.11868}, 
}

@inproceedings{martin2021nerf,
  title={{NeRF} in the Wild: Neural Radiance Fields for Unconstrained Photo Collections},
  author={Martin-Brualla, Ricardo and Radwan, Noha and Sajjadi, Mehdi SM and Barron, Jonathan T and Dosovitskiy, Alexey and Duckworth, Daniel},
  booktitle={CVPR},
  year={2021}
}

@inproceedings{turki2022mega,
  title={Mega-{NeRF}: Scalable Construction of Large-Scale {NeRF}s for Virtual Fly-Throughs},
  author={Turki, Haithem and Ramanan, Deva and Satyanarayanan, Mahadev},
  booktitle={CVPR},
  year={2022}
}

@inproceedings{tancik2022block,
  title={Block-{NeRF}: Scalable Large Scene Neural View Synthesis},
  author={Tancik, Matthew and Casser, Vincent and Yan, Xinchen and Pradhan, Sabeek and Mildenhall, Ben and Srinivasan, Pratul P and Barron, Jonathan T and Kretzschmar, Henrik},
  booktitle={CVPR},
  year={2022}
}

@inproceedings{niemeyer2022regnerf,
  title={{RegNeRF}: Regularizing Neural Radiance Fields for View Synthesis from Sparse Inputs},
  author={Niemeyer, Michael and Barron, Jonathan T and Mildenhall, Ben and Sajjadi, Mehdi SM and Geiger, Andreas and Radwan, Noha},
  booktitle={CVPR},
  year={2022}
}

@article{poole2022dreamfusion,
  title={{DreamFusion}: Text-to-{3D} Using {2D} Diffusion},
  author={Poole, Ben and Jain, Ajay and Barron, Jonathan T and Mildenhall, Ben},
  journal={arXiv preprint arXiv:2209.14988},
  year={2022}
}

@inproceedings{lin2023magic3d,
  title={{Magic3D}: High-Resolution Text-to-{3D} Content Creation},
  author={Lin, Chen-Hsuan and Gao, Jun and Tang, Luming and Takikawa, Towaki and Zeng, Xiaohui and Huang, Xun and Kreis, Karsten and Fidler, Sanja and Liu, Ming-Yu and Lin, Tsung-Yi},
  booktitle={CVPR},
  year={2023}
}

@inproceedings{liu2023zero,
  title={Zero-1-to-3: Zero-Shot One Image to {3D} Object},
  author={Liu, Ruoshi and Wu, Rundi and Van Hoorick, Basile and Tokmakov, Pavel and Zakharov, Sergey and Vondrick, Carl},
  booktitle={CVPR},
  year={2023}
}

@misc{google_earth_studio,
  author       = {{Google}},
  title        = {Google Earth Studio},
  year         = {2024},
  howpublished = {\url{https://earth.google.com/studio}},
  note         = {Accessed: 2025-05-14}
}

@article{sprintson2024fusionrf,
  title={{FusionRF}: High-Fidelity Satellite Neural Radiance Fields from Multispectral and Panchromatic Acquisitions},
  author={Sprintson, Michael and Chellappa, Rama and Peng, Cheng},
  journal={arXiv preprint arXiv:2409.15132},
  year={2024}
}

@article{zhang2023sparsesat,
  title={{Sparsesat-NeRF}: Dense depth supervised neural radiance fields for sparse satellite images},
  author={Zhang, Lulin and Rupnik, Ewelina},
  journal={arXiv preprint arXiv:2309.00277},
  year={2023}
}

@article{wang2023prolificdreamer,
  title={{ProlificDreamer}: High-fidelity and diverse text-to-3d generation with variational score distillation},
  author={Wang, Zhengyi and Lu, Cheng and Wang, Yikai and Bao, Fan and Li, Chongxuan and Su, Hang and Zhu, Jun},
  journal={Advances in neural information processing systems},
  volume={36},
  pages={8406--8441},
  year={2023}
}

@article{tang2024dreamgaussian,
  title={{DreamGaussian}: Generative gaussian splatting for efficient 3d content creation},
  author={Tang, Jiaxiang and Ren, Jiawei and Zhou, Hang and Liu, Ziwei and Zeng, Gang},
  journal={arXiv preprint arXiv:2309.16653},
  year={2023}
}

@inproceedings{yi2024gaussiandreamer,
  title={{GaussianDreamer}: Fast generation from text to 3d gaussians by bridging 2d and 3d diffusion models},
  author={Yi, Taoran and Fang, Jiemin and Wang, Junjie and Wu, Guanjun and Xie, Lingxi and Zhang, Xiaopeng and Liu, Wenyu and Tian, Qi and Wang, Xinggang},
  booktitle={Proceedings of the IEEE/CVF Conference on Computer Vision and Pattern Recognition},
  pages={6796--6807},
  year={2024}
}

@article{shi2023mvdream,
  title={{MVDream}: Multi-view diffusion for 3d generation},
  author={Shi, Yichun and Wang, Peng and Ye, Jianglong and Long, Mai and Li, Kejie and Yang, Xiao},
  journal={arXiv preprint arXiv:2308.16512},
  year={2023}
}

@inproceedings{mirzaei2023spinnerf,
  title={{Spin-NeRF}: Multiview segmentation and perceptual inpainting with neural radiance fields},
  author={Mirzaei, Ashkan and Aumentado-Armstrong, Tristan and Derpanis, Konstantinos G and Kelly, Jonathan and Brubaker, Marcus A and Gilitschenski, Igor and Levinshtein, Alex},
  booktitle={Proceedings of the IEEE/CVF Conference on Computer Vision and Pattern Recognition},
  pages={20669--20679},
  year={2023}
}

@inproceedings{shen2022cfnerf,
  title={{Conditional-flow NeRF}: Accurate 3d modelling with reliable uncertainty quantification},
  author={Shen, Jianxiong and Agudo, Antonio and Moreno-Noguer, Francesc and Ruiz, Adria},
  booktitle={European Conference on Computer Vision},
  pages={540--557},
  year={2022},
  organization={Springer}
}

@inproceedings{bosch2017mvs3d,
  title={A multiple view stereo benchmark for satellite imagery},
  author={Bosch, Marc and Kurtz, Zachary and Hagstrom, Shea and Brown, Myron},
  booktitle={2016 IEEE Applied Imagery Pattern Recognition Workshop (AIPR)},
  pages={1--9},
  year={2016},
  organization={IEEE}
}

@inproceedings{gao2021satmvs,
  title={Rational polynomial camera model warping for deep learning based satellite multi-view stereo matching},
  author={Gao, Jian and Liu, Jin and Ji, Shunping},
  booktitle={Proceedings of the IEEE/CVF international conference on computer vision},
  pages={6148--6157},
  year={2021}
}

@article{gao2023satmvs,
  title={A general deep learning based framework for 3D reconstruction from multi-view stereo satellite images},
  author={Gao, Jian and Liu, Jin and Ji, Shunping},
  journal={ISPRS Journal of Photogrammetry and Remote Sensing},
  volume={195},
  pages={446--461},
  year={2023},
  publisher={Elsevier}
}

@inproceedings{xu2024geospecific,
  title={Geospecific view generation geometry-context aware high-resolution ground view inference from satellite views},
  author={Xu, Ningli and Qin, Rongjun},
  booktitle={European Conference on Computer Vision},
  pages={349--366},
  year={2024},
  organization={Springer}
}

@inproceedings{jayasumana2024rethinking,
  title={Rethinking {FID}: Towards a better evaluation metric for image generation},
  author={Jayasumana, Sadeep and Ramalingam, Srikumar and Veit, Andreas and Glasner, Daniel and Chakrabarti, Ayan and Kumar, Sanjiv},
  booktitle={CVPR},
  year={2024}
}

@inproceedings{szegedy2016rethinking,
  title={Rethinking the inception architecture for computer vision},
  author={Szegedy, Christian and Vanhoucke, Vincent and Ioffe, Sergey and Shlens, Jon and Wojna, Zbigniew},
  booktitle={CVPR},
  year={2016}
}

@article{huang2025skysplat,
  title={{SkySplat}: Generalizable 3D Gaussian Splatting from Multi-Temporal Sparse Satellite Images},
  author={Huang, Xuejun and Liu, Xinyi and Wan, Yi and Zheng, Zhi and Zhang, Bin and Xiong, Mingtao and Pei, Yingying and Zhang, Yongjun},
  journal={arXiv preprint arXiv:2508.09479},
  year={2025}
}

@inproceedings{fan2025spectromotion,
  title={{SpectroMotion}: Dynamic 3d reconstruction of specular scenes},
  author={Fan, Cheng-De and Chang, Chen-Wei and Liu, Yi-Ruei and Lee, Jie-Ying and Huang, Jiun-Long and Tseng, Yu-Chee and Liu, Yu-Lun},
  booktitle={Proceedings of the Computer Vision and Pattern Recognition Conference},
  pages={21328--21338},
  year={2025}
}

@inproceedings{hou20253d,
  title={3D Gaussian Splatting with Grouped Uncertainty for Unconstrained Images},
  author={Hou, Hao-Yu and Hsu, Chia-Chi and Huang, Yu-Chen and Shen, Mu-Yi and Sun, Wei-Fang and Sun, Cheng and Chang, Chia-Che and Liu, Yu-Lun and Lee, Chun-Yi},
  booktitle={ICASSP 2025-2025 IEEE International Conference on Acoustics, Speech and Signal Processing (ICASSP)},
  pages={1--5},
  year={2025},
  organization={IEEE}
}

@inproceedings{lin2025frugalnerf,
  title={{FrugalNeRF}: Fast Convergence for Extreme Few-shot Novel View Synthesis without Learned Priors},
  author={Lin, Chin-Yang and Wu, Chung-Ho and Yeh, Chang-Han and Yen, Shih-Han and Sun, Cheng and Liu, Yu-Lun},
  booktitle={Proceedings of the Computer Vision and Pattern Recognition Conference},
  pages={11227--11238},
  year={2025}
}

@inproceedings{liu2025corrfill,
  title={Corrfill: Enhancing faithfulness in reference-based inpainting with correspondence guidance in diffusion models},
  author={Liu, Kuan-Hung and Yang, Cheng-Kun and Chen, Min-Hung and Liu, Yu-Lun and Lin, Yen-Yu},
  booktitle={2025 IEEE/CVF Winter Conference on Applications of Computer Vision (WACV)},
  pages={1618--1627},
  year={2025},
  organization={IEEE}
}

@article{yuchen2024dogaussian,
  title={{DOGS}: Distributed-oriented gaussian splatting for large-scale {3D} reconstruction via gaussian consensus},
  author={Chen, Yu and Lee, Gim Hee},
  journal={Advances in Neural Information Processing Systems},
  volume={37},
  pages={34487--34512},
  year={2024}
}

@misc{gao2025citygsx,
      title={{CityGS-X}: A Scalable Architecture for Efficient and Geometrically Accurate Large-Scale Scene Reconstruction}, 
      author={Yuanyuan Gao and Hao Li and Jiaqi Chen and Zhengyu Zou and Zhihang Zhong and Dingwen Zhang and Xiao Sun and Junwei Han},
      year={2025},
      eprint={2503.23044},
      archivePrefix={arXiv},
      primaryClass={cs.CV},
      url={https://arxiv.org/abs/2503.23044}, 
}

@inproceedings{jiang2025horizon,
  title={{Horizon-GS}: Unified {3D} Gaussian Splatting for Large-Scale Aerial-to-Ground Scenes},
  author={Jiang, Lihan and Ren, Kerui and Yu, Mulin and Xu, Linning and Dong, Junting and Lu, Tao and Zhao, Feng and Lin, Dahua and Dai, Bo},
  booktitle={Proceedings of the Computer Vision and Pattern Recognition Conference},
  pages={26789--26799},
  year={2025}
}

@article{qian2026sat2densitypp,
  author={Qian, Ming and Tan, Bin and Wang, Qiuyu and Zheng, Xianwei and Xiong, Hanjiang and Xia, Gui-Song and Shen, Yujun and Xue, Nan},
  journal={IEEE Transactions on Pattern Analysis and Machine Intelligence}, 
  title={Seeing through Satellite Images at Street Views}, 
  year={2026},
  volume={},
  number={},
  pages={1-18},
  doi={10.1109/TPAMI.2026.3652860}}

@inproceedings{qian2026sat3dgen,
  title={{Sat3DGen}: Comprehensive Street-Level {3D} Scene Generation from Single Satellite Image},
  author={Qian, Ming and Xia, Zimin and Liu, Changkun and Ma, Shuailei and Wang, Wen and Ke, Zeran and Tan, Bin and Zhang, Hang and Xia, Gui-Song},
  booktitle={International Conference on Learning Representations (ICLR)},
  year={2026}
}

@inproceedings{yao2025magiccity,
  title={{MagicCity}: Geometry-Aware 3D City Generation from Satellite Imagery with Multi-View Consistency},
  author={Yao, Xingbo and Wang, Xuanmin and Wu, Hao and Ping, Chengliang and Zhang, Doudou and Xiong, Hui},
  booktitle={Proceedings of the IEEE/CVF International Conference on Computer Vision},
  pages={25325--25334},
  year={2025}
}

@inproceedings{hua2025sat2city,
  title={{Sat2City}: {3D} city generation from a single satellite image with cascaded latent diffusion},
  author={Hua, Tongyan and Jiang, Lutao and Chen, Ying-Cong and Zhao, Wufan},
  booktitle={Proceedings of the IEEE/CVF International Conference on Computer Vision},
  pages={27978--27988},
  year={2025}
}

@misc{kang2025sat2realcity,
  title={{Sat2RealCity}: Geometry-Aware and Appearance-Controllable {3D} Urban Generation from Satellite Imagery}, 
  author={Yijie Kang and Xinliang Wang and Zhenyu Wu and Yifeng Shi and Hailong Zhu},
  year={2025},
  eprint={2511.11470},
  archivePrefix={arXiv},
  primaryClass={cs.CV},
  url={https://arxiv.org/abs/2511.11470}, 
}

@misc{wu2024ags,
      title={{3D} Gaussian Splatting for Large-scale Surface Reconstruction from Aerial Images}, 
      author={YuanZheng Wu and Jin Liu and Shunping Ji},
      year={2024},
      eprint={2409.00381},
      archivePrefix={arXiv},
      primaryClass={cs.CV},
      url={https://arxiv.org/abs/2409.00381}, 
}

@misc{regmi2018crossview,
      title={Cross-View Image Synthesis using Conditional GANs}, 
      author={Krishna Regmi and Ali Borji},
      year={2018},
      eprint={1803.03396},
      archivePrefix={arXiv},
      primaryClass={cs.CV},
      url={https://arxiv.org/abs/1803.03396}, 
}

@article{shi2022geometry,
  title={Geometry-guided street-view panorama synthesis from satellite imagery},
  author={Shi, Yujiao and Campbell, Dylan and Yu, Xin and Li, Hongdong},
  journal={IEEE Transactions on Pattern Analysis and Machine Intelligence},
  volume={44},
  number={12},
  pages={10009--10022},
  year={2022},
  publisher={IEEE}
}

@inproceedings{xu2025satellitetogroundscape,
  title={Satellite to groundscape-large-scale consistent ground view generation from satellite views},
  author={Xu, Ningli and Qin, Rongjun},
  booktitle={Proceedings of the Computer Vision and Pattern Recognition Conference},
  pages={6068--6077},
  year={2025}
}

@misc{zhang2024cityxcontrollableproceduralcontent,
      title={CityX: Controllable Procedural Content Generation for Unbounded 3D Cities}, 
      author={Shougao Zhang and Mengqi Zhou and Yuxi Wang and Chuanchen Luo and Rongyu Wang and Yiwei Li and Zhaoxiang Zhang and Junran Peng},
      year={2024},
      eprint={2407.17572},
      archivePrefix={arXiv},
      primaryClass={cs.CV},
      url={https://arxiv.org/abs/2407.17572}, 
}

@inproceedings{zhou2025scenex,
  title={{SceneX}: Procedural controllable large-scale scene generation},
  author={Zhou, Mengqi and Wang, Yuxi and Hou, Jun and Zhang, Shougao and Li, Yiwei and Luo, Chuanchen and Peng, Junran and Zhang, Zhaoxiang},
  booktitle={Proceedings of the AAAI Conference on Artificial Intelligence},
  volume={39},
  pages={10806--10814},
  year={2025}
}

@inproceedings{toker2021coming,
  title={Coming down to earth: Satellite-to-street view synthesis for geo-localization},
  author={Toker, Aysim and Zhou, Qunjie and Maximov, Maxim and Leal-Taix{\'e}, Laura},
  booktitle={Proceedings of the IEEE/CVF Conference on Computer Vision and Pattern Recognition},
  pages={6488--6497},
  year={2021}
}

@article{shang2024urbanworld,
  title={{UrbanWorld}: An urban world model for {3D} city generation},
  author={Shang, Yu and Lin, Yuming and Zheng, Yu and Fan, Hangyu and Ding, Jingtao and Feng, Jie and Chen, Jiansheng and Tian, Li and Li, Yong},
  journal={arXiv preprint arXiv:2407.11965},
  year={2024}
}

@inproceedings{yu2025wonderworld,
  title={{WonderWorld}: Interactive {3D} scene generation from a single image},
  author={Yu, Hong-Xing and Duan, Haoyi and Herrmann, Charles and Freeman, William T and Wu, Jiajun},
  booktitle={Proceedings of the Computer Vision and Pattern Recognition Conference},
  pages={5916--5926},
  year={2025}
}

@inproceedings{shriram2024realmdreamer,
    title={{RealmDreamer}: Text-Driven {3D} Scene Generation with Inpainting and Depth Diffusion},
    author={Jaidev Shriram and Alex Trevithick and Lingjie Liu and Ravi Ramamoorthi},
    booktitle={International Conference on 3D Vision (3DV)},
    year={2025}
}

@article{oquab2024dinov2,
    title={{DINOv2}: Learning Robust Visual Features without Supervision},
    author={Oquab, Maxime and Darcet, Timoth{\'e}e and Moutakanni, Th{\'e}o and Vo, Huy V. and Szafraniec, Marc and Khalidov, Vasil and Fernandez, Pierre and Haziza, Daniel and Massa, Francisco and El-Nouby, Alaaeldin and Howes, Russell and Huang, Po-Yao and Xu, Hu and Sharma, Vasu and Li, Shang-Wen and Galuba, Wojciech and Rabbat, Mike and Assran, Mido and Ballas, Nicolas and Synnaeve, Gabriel and Misra, Ishan and Jegou, Herve and Mairal, Julien and Labatut, Patrick and Joulin, Armand and Bojanowski, Piotr},
    journal={Transactions on Machine Learning Research (TMLR)},
    year={2024}
}

@inproceedings{huang2024vbench,
    title={{VBench}: Comprehensive Benchmark Suite for Video Generative Models},
    author={Huang, Ziqi and He, Yinan and Yu, Jiashuo and Zhang, Fan and Si, Chenyang and Jiang, Yuming and Zhang, Yuanhan and Wu, Tianxing and Jin, Qingyang and Chanpaisit, Nattapol and Wang, Yaohui and Chen, Xinyuan and Wang, Limin and Lin, Dahua and Qiao, Yu and Liu, Ziwei},
    booktitle={Proceedings of the IEEE/CVF Conference on Computer Vision and Pattern Recognition (CVPR)},
    year={2024}
}

\clearpage
\appendix
\section{Supplementary Material}
This supplementary material provides additional details complementing the main 
paper, organized into the following sections:

\begin{enumerate}
    \item \textbf{Implementation Details (Sec.~\ref{sup:imple})}: Expanded descriptions of method components including pseudo-camera depth supervision, 3DGS reconstruction parameters, and the FlowEdit-based refinement process. This section also covers system profiling (rendering efficiency, memory consumption) and mathematical validation of the RPC-to-perspective approximation.
    
    \item \textbf{Main Paper Experiments Detail \& Results (Sec.~\ref{sup:exp})}: Comprehensive dataset statistics, the user study methodology, and the evaluation protocol. Additionally, this section provides the complete per-scene quantitative benchmarks and extended qualitative comparisons across both the DFC2019 and GoogleEarth datasets.
    
    \item \textbf{Additional Experiments (Sec.~\ref{sup:additional-exp})}: Further evaluations demonstrating the robustness of our approach, including: (i) synthesis on complex irregular geometries and bridges; (ii) visualizing transient object handling; (iii) refinement prompt sensitivity analysis; (iv) multi-block scalability via combined imagery; (v) episode-vs-coverage analysis quantifying curriculum effectiveness; and (vi) structural consistency under different random seeds.
\end{enumerate}

We provide an interactive HTML visualization (\texttt{main.html}) 
for exploring video results across scenes and viewing conditions. To verify 
that Skyfall-GS learns a coherent, globally explorable 3D structure rather 
than overfitting to IDU-sampled viewpoints, we include free-flight trajectory 
renders along novel, smooth camera paths withheld during training. These 
renders confirm that our method produces a fully immersive and structurally 
consistent exploration experience, and enable direct perceptual comparison 
against baseline approaches and Google Earth Studio reference videos.

\section{Implementation Details} \label{sup:imple}

\subsection{Method Components}

\subsubsection{Codebase.}
Our method extends the Mip-Splatting~\cite{yu2024mip} codebase with custom modules for satellite imagery processing and our curriculum-based IDU refinement pipeline.

\subsubsection{Pseudo camera depth supervision.}
\begin{figure}[t]
\centering
\setlength{\tabcolsep}{4pt}
\resizebox{0.85\linewidth}{!}{%
\begin{tabular}{ccc}
\includegraphics[width=0.33\textwidth]{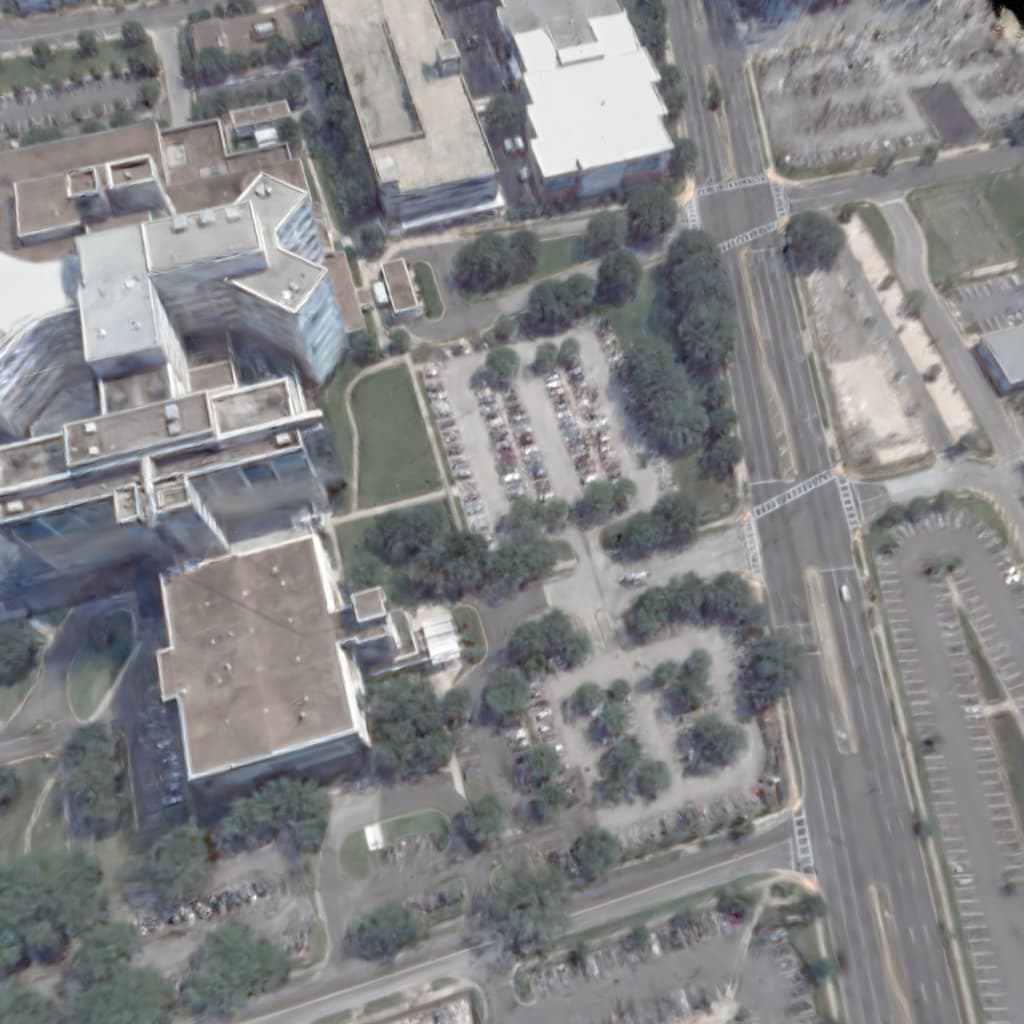} & \includegraphics[width=0.33\textwidth]{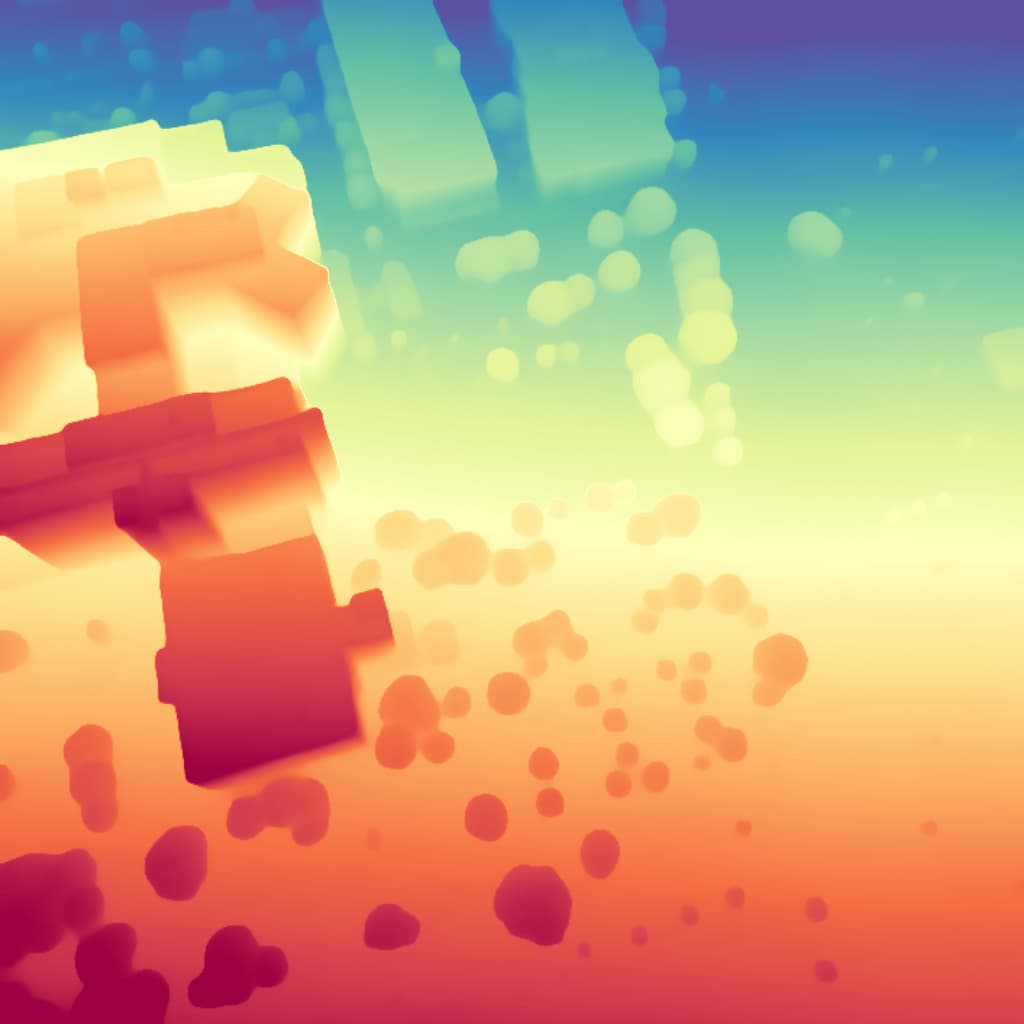} & \includegraphics[width=0.33\textwidth]{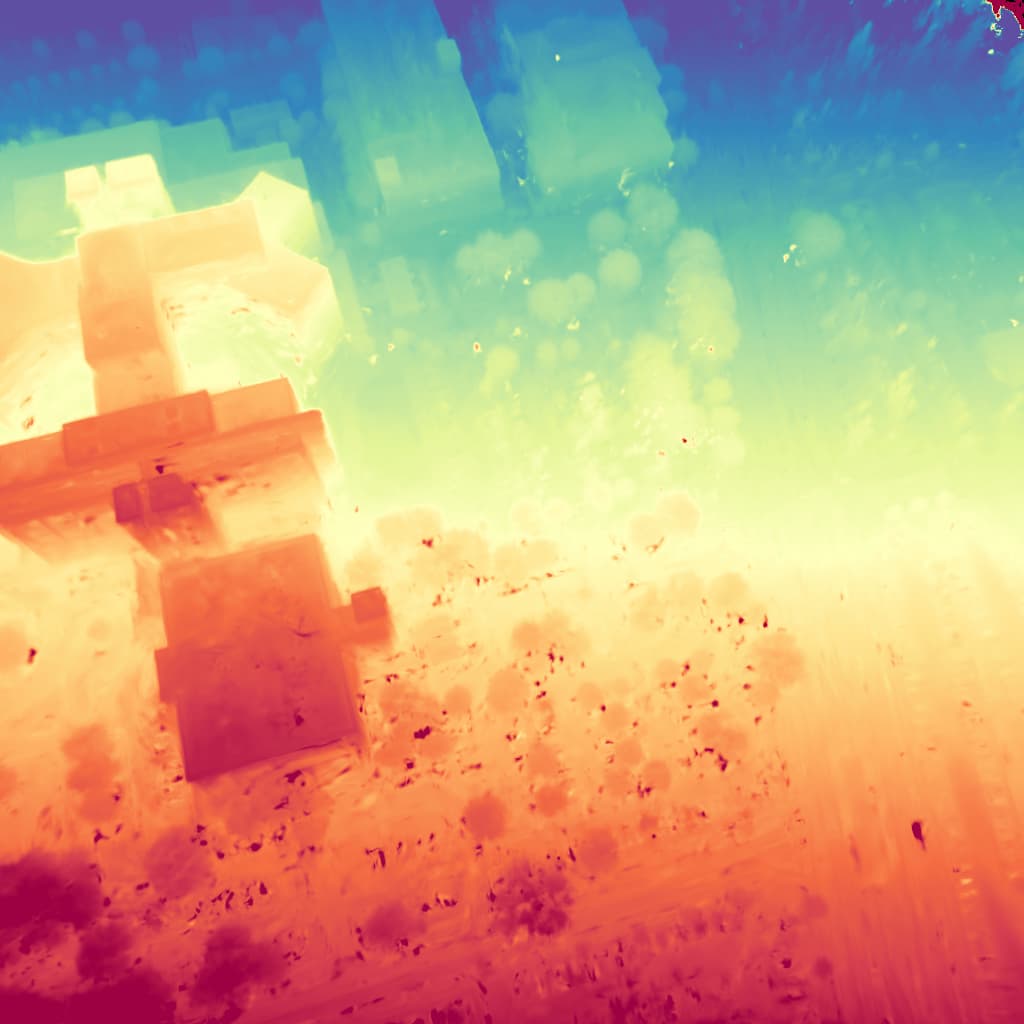} \\
\includegraphics[width=0.33\textwidth]{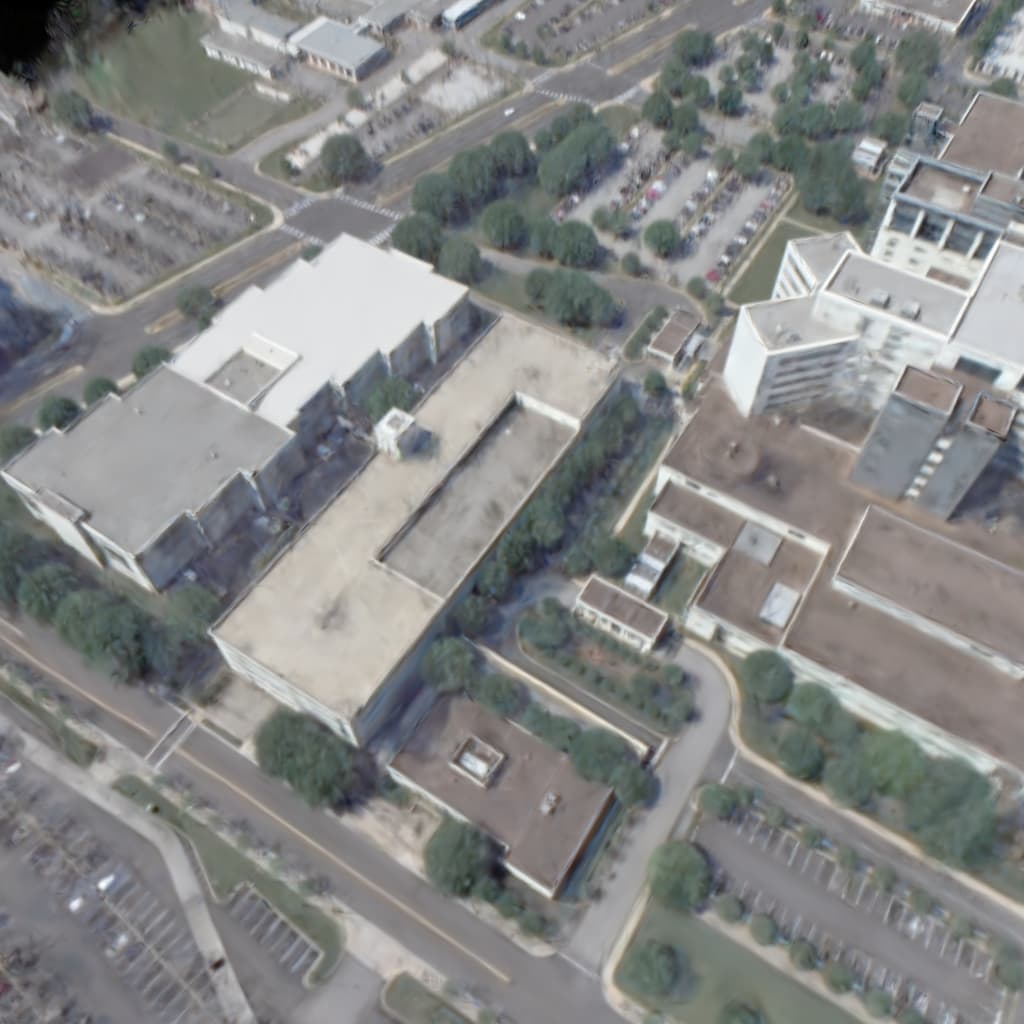} & \includegraphics[width=0.33\textwidth]{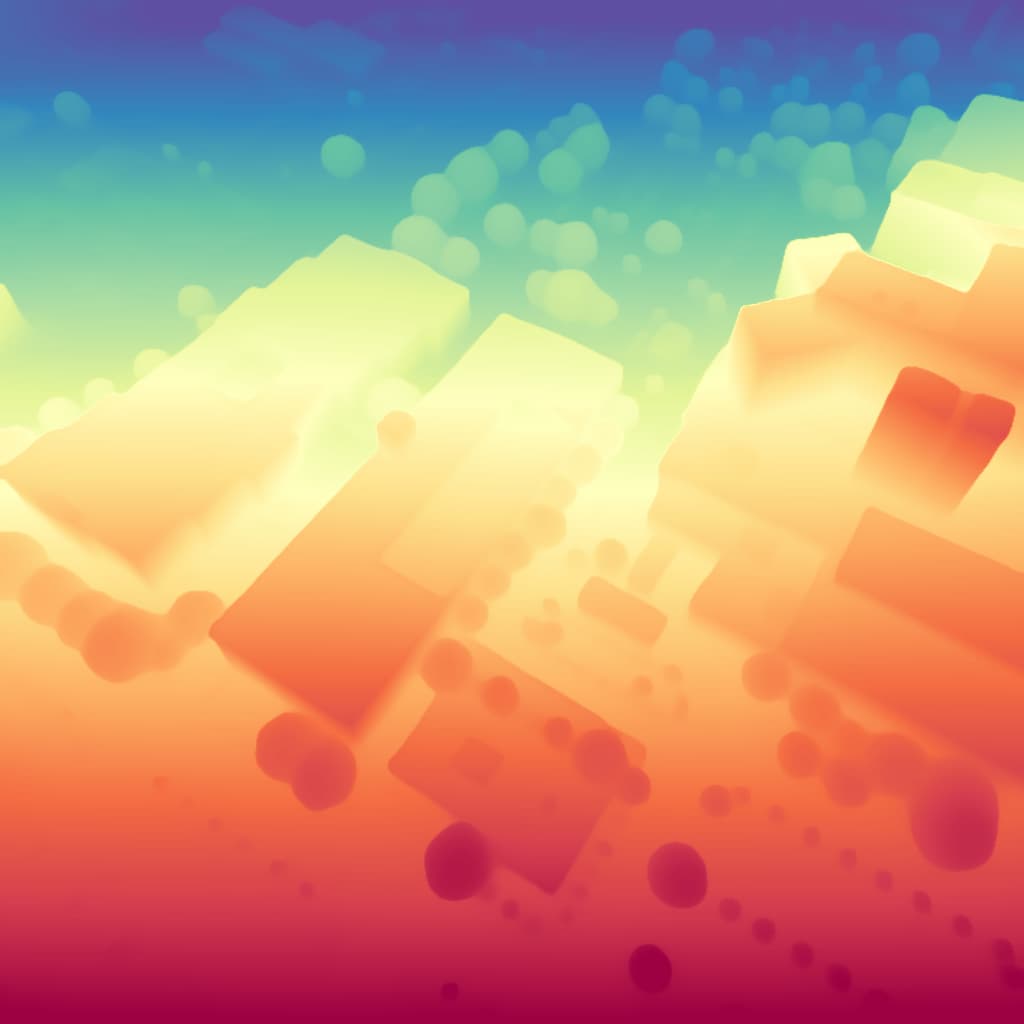} & \includegraphics[width=0.33\textwidth]{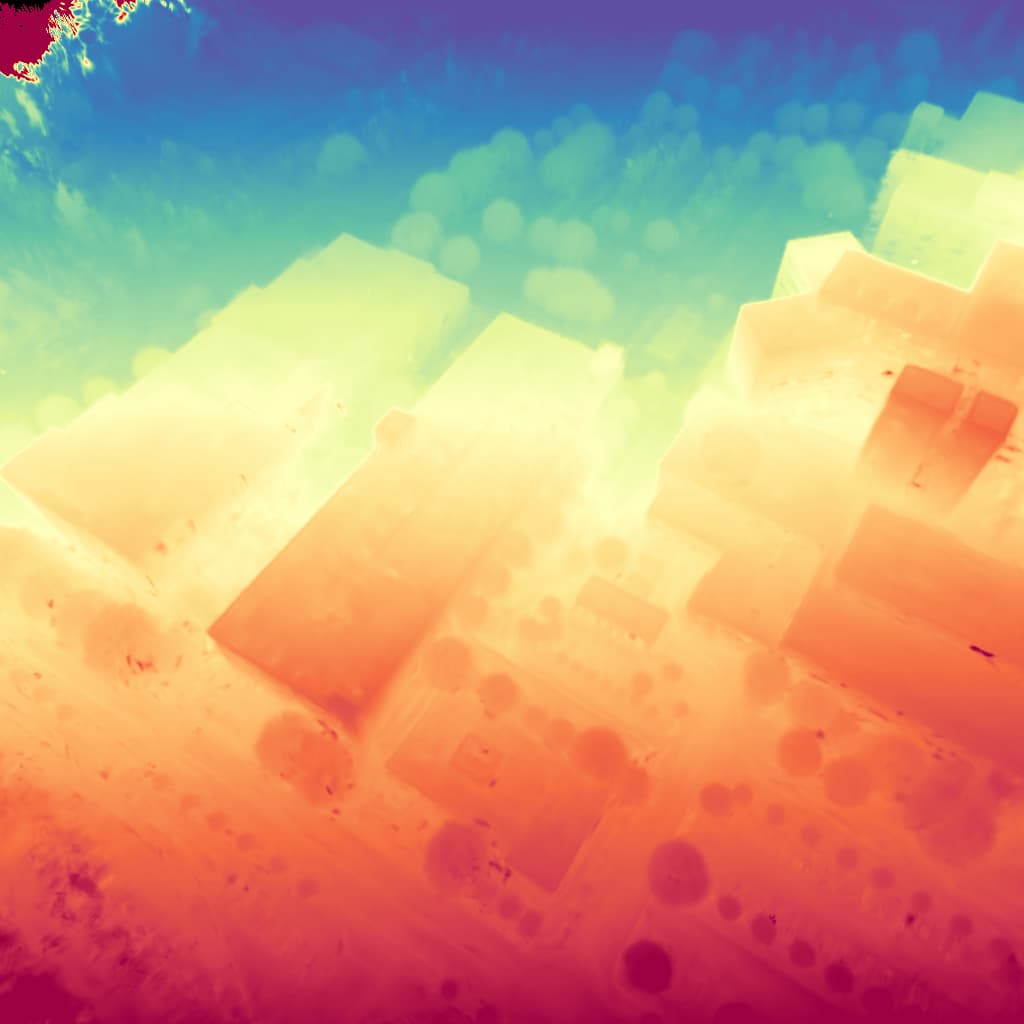} \\
Render RGB $I_\text{RGB}$ & Estimated Depth $D_\text{est}$ & 3DGS Depth $D_\text{GS}$ \\
\end{tabular}
}%
\caption{\textbf{Pseudo-cam Depth Supervision.} We use MoGe~\cite{wang2024moge} to estimate the scale-invariant depth $D_\text{est}$ from the rendered RGB image $I_\text{RGB}$. The rightmost figures show the rasterized depth $D_\text{GS}$ from 3DGS.}
\label{fig:suppl_depth}
\end{figure}

We sample cameras with varied azimuths and decreasing elevations, using random per-image embeddings. MoGe~\cite{wang2024moge} provides scale-invariant depth estimation. We sample 24 views every 10 iterations, with look-at points $(x, y, z)$, where $x,y \sim \mathcal{N}(0,128)$ and $z=0$. Camera azimuths are uniformly sampled between 0 and $2\pi$, while elevation angles and radii linearly decrease from $80\degree$ to $45\degree$ and 300 to 250 units, respectively. Rendered RGB images ($I_\text{RGB}$) are $1024\times1024$ pixels. We illustrate the 3DGS rendered RGB image $I_{\text{RGB}}$, scale-invariant depth $D_\text{est}$ estimated by MoGe~\cite{wang2024moge} and depth from 3DGS $D_{\text{GS}}$ in \Cref{fig:suppl_depth}.

\subsubsection{3DGS reconstruction from satellite imagery.}
Our satellite-view optimization process runs for 30,000 iterations, with densification enabled between iterations 1,000 and 21,000. We modify several key parameters in the standard 3DGS implementation to address satellite imagery's unique challenges. First, to prevent undesirable Gaussian elongation artifacts common with overhead views, we reduce the scaling learning rate from 0.005 to 0.001. Second, we implement pruning of Gaussians with maximum covariance exceeding 20 to eliminate floating artifacts. The loss function weights are set to $\lambda_{\text{D-SSIM}}=0.2$, $\lambda_{\text{op}}=10$, and $\lambda_{\text{depth}}=0.5$ for optimal reconstruction quality. For the GoogleEarth dataset, we disable the opacity regularization, since the dense training views are sufficient to eliminate floaters. For appearance modeling, we adopt the architecture from WildGaussians~\cite{kulhanek2024wildgaussians}, implementing an appearance MLP with 2 hidden layers (128 neurons each) and ReLU activation functions. The per-image and per-Gaussian embedding dimensions are set to 32 and 24, respectively, with learning rates of 0.001, 0.005, and 0.0005 for per-image embeddings $e_j$, per-Gaussian embeddings $g_i$, and the appearance MLP $f$, respectively. The complete satellite-view training requires approximately 1 hour on a single NVIDIA RTX A6000 GPU.

\subsubsection{FlowEdit-based refinement.}
We set FlowEdit noise parameters $n_{\text{min}}=4$ and $n_{\text{max}}=10$ to balance artifact removal with detail preservation. Our source prompt (``\textit{Satellite image of an urban area with modern and older buildings, roads, green spaces. Some areas appear distorted, with blurring and warping artifacts.}'') characterizes initial renders, while the target prompt (``\textit{Clear satellite image of an urban area with sharp buildings, smooth edges, natural lighting, and well-defined textures.}'') guides refinement. These parameters were determined through experimentation, with lower noise values preserving more original structure but removing fewer artifacts, and higher values creating more significant changes but potentially altering underlying geometry. All other FlowEdit parameters use default values.

\subsubsection{Curriculum-based refinement.}
Our IDU process comprises $N_e=5$ episodes of 10,000 iterations each, with densification through iteration 9,000. At the start of IDU, we randomly select and fix a single per-image appearance embedding $e_j$. Opacity regularization is disabled during IDU, as our curriculum naturally mitigates floating artifacts through multi-view consistency, enabling Gaussians to retain variable opacities beneficial for semi-transparent structures~\cite{kerbl20233d}. For DFC2019~\cite{c6tm-vw12-19} dataset, we utilize $N_p=9$ look-at points in a $3\times3$ grid (512 units wide, centered at origin), with $N_v=6$ cameras per point and $N_s=2$ samples per view. Camera elevations decrease from $85^{\circ}$ to $45^{\circ}$ and radii from 300 to 250 units across episodes. For GoogleEarth~\cite{xie2024citydreamer} dataset, we utilize $N_p=16$ look-at point at origin, with $N_v=6$ cameras per point and $N_s=2$ samples per view. Camera elevations decrease from $85^{\circ}$ to $45^{\circ}$ and radius is fixed 600-unit across episodes. All training images are rendered at $2048\times2048$ resolution. Our training strategy samples 75\% from refined images and 25\% from original satellite images, this sampling strategy makes sure that the final 3DGS scene faithfully follows the semantic and layout in the input satellite imagery. The complete synthesizing stage requires approximately 6 hours on a single NVIDIA RTX A6000 GPU.

\subsection{Resource and Approximation Analysis}

\vspace{\subsubsecmargin}
\subsubsection{Rendering efficiency. } Our method achieves 11~FPS on the modest NVIDIA T4, significantly outperforming CityDreamer's 0.18~FPS, which runs on the far more powerful NVIDIA A100 ($5\times$ the CUDA cores, $10\times$ the memory bandwidth). GaussianCity reaches comparable speeds (10.72 FPS) but requires the high-end A100. Furthermore, our fused representation enables real-time rendering at 40 FPS on consumer hardware (MacBook Air M2), demonstrating that our method
enables high-quality 3D urban navigation without specialized computing resources.

\subsubsection{Memory consumption.}
We distinguish between peak memory and final memory. The peak memory usage reaches 46 GB during the synthesis stage, driven by the overhead of loading the diffusion model (FLUX.1) and temporary densification of Gaussians. However, the final training memory footprint is significantly lower (28.04 GB) as our method actively prunes redundant and low-opacity points. In terms of scene complexity, the refinement process densifies the scene by approximately 27\%, increasing the Gaussian count from $\sim$1.65 million (reconstruction stage) to $\sim$2.1 million, specifically targeting the vertical facade geometry missing in the initial satellite reconstruction.

\subsubsection{Validity of RPC to perspective approximation.}
We adopt the methodology proposed in SatelliteSfM~\cite{VisSat-2019} to approximate the satellite linear pushbroom sensor as a perspective camera. This approximation relies on the ``weak perspective'' assumption, which holds valid when the satellite altitude ($Z$) is significantly larger than the depth variation within the scene ($\Delta Z$), i.e., $Z \gg \Delta Z$. Given that satellites orbit at distances of hundreds of kilometers while terrestrial depth variations are limited to a few hundred meters, the ratio $\Delta Z / Z$ remains negligible, allowing the geometry to converge to a perspective model. The approximation is achieved by generating a dense grid of 3D-2D correspondences using the rigorous RPC model and solving for a projection matrix $P$ via the Direct Linear Transformation (DLT) method, which is subsequently decomposed ($P = K[R|t]$) to recover camera parameters. Quantitative evaluations demonstrate that this process introduces negligible error: the average maximum forward projection error against the rigorous RPC model is only \textbf{0.126 pixels}, and the difference in triangulated 3D points is typically less than \textbf{5 cm}. Furthermore, this initialization allows Bundle Adjustment to achieve sub-pixel accuracy, with median reprojection errors recorded at \textbf{0.864 pixels}, confirming the suitability of this approximation for high-fidelity 3D reconstruction.

\section{Main Paper Experiments Detail \& Results} \label{sup:exp}

\subsection{Dataset Details} \label{sup:dataset-detail}

\subsubsection{DFC2019 dataset.}
The number of training images and geographical coordinates for each AOI is provided in \Cref{tab:num_train}. We also include four additional AOIs from Jacksonville to demonstrate our method's robustness across varying scene characteristics. The number of training images and geographical coordinates for these additional AOIs is provided in \Cref{tab:num_train_additional}. These additional AOIs feature distinct characteristics: one contains a city hall building (JAX\_164), another includes an American football stadium (JAX\_175), while the remaining two exhibit other notable urban features (JAX\_168 and JAX\_264).

\subsubsection{GoogleEarth dataset.}
The GoogleEarth dataset, introduced by CityDreamer~\cite{xie2024citydreamer}, contains semantic maps, height fields and renders from Google Earth Studio~\cite{google_earth_studio} of New York City. This dataset is used to train the generative model in CityDreamer~\cite{xie2024citydreamer} and GaussianCity~\cite{xie2025gaussiancity}. We pick four AOIs which contain diverse city elements, including complex architectures (\texttt{004}), squares (\texttt{010}), resident area (\texttt{219}) and riverside (\texttt{336}). However, original GES renders provided in GoogleEarth dataset are rendered from a lower elevation angle, which is not similar to satellite imagery. Therefore, for each AOI, we render 60 images from GES using an orbit trajectory with $80^\circ$ of elevation angle and 2219 meter of radius. These new renders serve as the input of our methods. The AOI ID, geographical coordinates, and the number of input images are detailed in \Cref{tab:nyc_detail}.

\begin{table*}[t]
\caption{\textbf{Number of training images and geographical coordinate per Area of Interest (AOI).} These AOIs correspond to standard evaluation scenarios established by previous works, ensuring consistent and fair comparisons with existing baselines (e.g., Sat-NeRF~\cite{mari2022sat}).}
\label{tab:num_train}
\resizebox{\linewidth}{!}{%
\begin{tabular}{rcccc}
\toprule
AOI              & JAX\_004 & JAX\_068 & JAX\_214 & JAX\_260 \\ 
\midrule
\# of training image & 9      & 17     & 21     & 15     \\ 
Geographical coordinate & 81.70643$\degree$W, 30.35782$\degree$N & 81.66375$\degree$W, 30.34880$\degree$N & 81.66353$\degree$W, 30.31646$\degree$N & 81.66350$\degree$W, 30.31184$\degree$N \\ 
\bottomrule
\end{tabular}
}%
\end{table*}

\begin{table*}[t]
\caption{\textbf{Number of training images and geographical coordinates for additional AOIs.} We selected 4 additional AOIs with distinct characteristics: JAX\_164 features a city hall building, JAX\_175 contains an American football stadium, while the remaining two AOIs present other notable urban structures.}\label{tab:num_train_additional}
\resizebox{\linewidth}{!}{%
\begin{tabular}{rcccc}
\toprule
AOI              & JAX\_164 & JAX\_168 & JAX\_175 & JAX\_264 \\ 
\midrule
\# of training image & 20      & 21     & 21     & 21     \\ 
Geographical coordinate & 81.66362$\degree$W, 30.33032$\degree$N & 81.65297$\degree$W, 30.33037$\degree$N & 81.63696$\degree$W, 30.32583$\degree$N & 81.65285$\degree$W, 30.31189$\degree$N \\ 
\bottomrule
\end{tabular}
}%
\end{table*}

\begin{table*}[t]
\caption{\textbf{Number of training images and geographical coordinate per Area of Interest (AOI).} We pick 4 AOIs from the GoogleEarth~\cite{xie2024citydreamer} dataset, ensuring fair comparisons with existing baselines (e.g., CityDreamer ~\cite{xie2024citydreamer} and GaussianCity~\cite{xie2025gaussiancity})}
\label{tab:nyc_detail}
\resizebox{\linewidth}{!}{%
\begin{tabular}{rcccc}
\toprule
AOI              & 4WorldFinancialCtr (004) & 10UnionSquareE\#5P (010) & 219E12thSt (219) & 336AlbanySt (336) \\ 
\midrule
\# of training image & 60      & 60     & 60     & 60 \\
Geographical coordinate & 74.01587$\degree$W, 40.71473$\degree$N & 73.98975$\degree$W, 40.73482$\degree$N & 73.98690$\degree$W, 40.73187$\degree$N & 74.01753$\degree$W, 40.71020$\degree$N \\
\bottomrule
\end{tabular}
}%
\end{table*}

\subsection{Evaluation Protocol} \label{sup:eval-proto}

\subsubsection{User study.}
We asked participants three specific questions and instructed them to select one video that best addressed each question:
\begin{enumerate}
    \item \textbf{Geometric Accuracy}: ``Which video's 3D structures (buildings, terrain, objects) more accurately represent the real-world geometry when compared to the ground truth video?''
    \item \textbf{Spatial Alignment}: ``Which video's layout and positioning of elements better matches the satellite imagery reference?''
    \item \textbf{Overall Perceptual Quality}: ``Considering all aspects (geometry, textures, lighting, consistency), which video presents a more convincing and high-quality 3D representation of the scene?''
\end{enumerate}
For the user study on DFC2019 dataset, each participant viewed videos from Sat-NeRF~\cite{mari2022sat}, EOGS~\cite{savantaira2024eogs}, CoR-GS~\cite{zhang2024cor}, Mip-Splatting~\cite{yu2024mip}, and our complete method, alongside Google Earth Studio reference footage and the original satellite imagery. For the user study on the GoogleEarth dataset, each participant viewed videos from CityDreamer~\cite{xie2024citydreamer}, GaussianCity~\cite{xie2025gaussiancity}, CoR-GS~\cite{zhang2024cor}, Mip-Splatting~\cite{yu2024mip} and our complete method, alongside Google Earth Studio reference footage and the reference satellite imagery.

\subsubsection{Comparison details.}
For quantitative comparisons with Sat-NeRF~\cite{mari2022sat}, Mip-Splatting~\cite{yu2024mip} and our method without IDU refinement, we used consistent camera parameters across all methods: $17^{\circ}$ elevation angle, 328-unit radius, and $20^{\circ}$ field of view, with cameras targeting the AOI's origin.
For comparisons with CityDreamer~\cite{xie2024citydreamer} and GaussianCity~\cite{xie2025gaussiancity}, we use $45^{\circ}$ elevation angle, 1067-unit radius, and $20^{\circ}$ field of view, with cameras also targeting the AOI's origin.
These parameters were selected to ensure equitable comparison with similar scene coverage across methods.

\subsection{Results} \label{sup:results}

\subsubsection{Per-scene quantitative comparison.}
We provide per-scene quantitative comparisons in \Cref{tab:cor_gs_comparison,tab:cor_gs_comparison_2}. On the DFC2019 dataset (\Cref{tab:cor_gs_comparison}), our method achieves the best $\text{FID}_{\text{CLIP}}$ and CMMD scores across all four AOIs by a substantial margin, reducing $\text{FID}_{\text{CLIP}}$ from the next-best $\sim$84 to $\sim$27 on average. Pixel-level metrics are more mixed: our method leads on PSNR in three out of four scenes, while CoR-GS occasionally achieves higher SSIM scores due to its tendency to produce overly smooth reconstructions, which artificially inflate SSIM as discussed in the main paper. LPIPS improvements are consistent, with our method achieving the best or second-best score on all four scenes. On the GoogleEarth dataset (\Cref{tab:cor_gs_comparison_2}), our method achieves the best $\text{FID}_{\text{CLIP}}$ and CMMD in three out of four. The one exception is scene 219, a low-rise residential area with limited vertical facades, where Mip-Splatting achieves lower $\text{FID}_{\text{CLIP}}$ (7.06 \vs 7.80) and CMMD (1.589 \vs 2.640). We attribute this to this scene's relatively flat geometry, which reduces the benefit of our curriculum-based facade synthesis and allows reconstruction-only baselines to perform competitively. This is consistent with our method's design intent: the gains from diffusion-based refinement are most pronounced in scenes with significant occluded vertical structure. Despite this single exception, our method achieves the best average performance across both datasets, confirming its robustness across diverse urban typologies.

\begin{table*}[t]
    \centering
    \caption{\textbf{Quantitative comparison on each AOI of DFC2019~\cite{c6tm-vw12-19}.} 
    Our method consistently outperforms baseline methods on distribution metrics and most pixel-level metrics, indicating superior image synthesis quality. Metrics are computed between renders from each method and reference frames from GES. \colorbox{red!25}{Red}, \colorbox{orange!25}{orange}, and \colorbox{yellow!25}{yellow} indicate the best, second best, and third best results, respectively.
    }
    \label{tab:cor_gs_comparison_2}
    \begin{tabular}{llccccc}
        \toprule
        & & \multicolumn{2}{c}{Distribution Metrics} & \multicolumn{3}{c}{Pixel-level Metrics*} \\
        \cmidrule(lr){3-4} \cmidrule(lr){5-7}
        Scene & Methods &  $\text{FID}_{\text{CLIP}}\, \downarrow$ & CMMD$\, \downarrow$ & PSNR$\, \uparrow$ & SSIM$\, \uparrow$ & LPIPS$\, \downarrow$ \\
        \midrule
        \multirow{5}{*}{JAX\_004} 
        & Sat-NeRF & \cellcolor{orange!25}77.71 & \cellcolor{orange!25}3.655 & \cellcolor{yellow!25}12.00 & 0.2282 & \cellcolor{yellow!25}0.8649 \\
        & EOGS & 106.69 & 5.889 & 8.21 & 0.1266 & 1.0165 \\
        & CoR-GS & \cellcolor{yellow!25}84.43 & 5.432 & 11.24 & \cellcolor{red!25}0.2550 & 0.9763 \\
        & Mip-Splatting & 85.28 & \cellcolor{yellow!25}5.010 & \cellcolor{red!25}13.08 & \cellcolor{yellow!25}0.2415 & \cellcolor{red!25}0.8134 \\
        & Ours & \cellcolor{red!25}24.07 & \cellcolor{red!25}1.481 & \cellcolor{orange!25}12.93 & \cellcolor{orange!25}0.2449 & \cellcolor{orange!25}0.8450 \\
        \midrule
        \multirow{5}{*}{JAX\_068} 
        & Sat-NeRF & 92.59 & \cellcolor{orange!25}5.371 & 9.86 & 0.2590 & \cellcolor{orange!25}0.8383 \\
        & EOGS & \cellcolor{orange!25}86.08 & \cellcolor{yellow!25}5.536 & 6.39 & 0.1593 & 0.9944 \\
        & CoR-GS & \cellcolor{yellow!25}88.05 & 6.402 & \cellcolor{orange!25}11.78 & \cellcolor{red!25}0.3231 & 1.0065 \\
        & Mip-Splatting & 93.29 & 6.201 & \cellcolor{yellow!25}11.66 & \cellcolor{yellow!25}0.2908 & \cellcolor{yellow!25}0.8435 \\
        & Ours & \cellcolor{red!25}28.38 & \cellcolor{red!25}2.893 & \cellcolor{red!25}11.82 & \cellcolor{orange!25}0.2939 & \cellcolor{red!25}0.8193 \\
        \midrule
        \multirow{5}{*}{JAX\_214} 
        & Sat-NeRF & 89.52 & 5.308 & 8.92 & 0.2659 & \cellcolor{yellow!25}0.8384 \\
        & EOGS & \cellcolor{orange!25}71.03 & \cellcolor{orange!25}4.362 & 7.39 & 0.2296 & 0.8890 \\
        & CoR-GS & 83.17 & 5.405 & \cellcolor{orange!25}11.67 & \cellcolor{red!25}0.4075 & 0.9075 \\
        & Mip-Splatting & \cellcolor{yellow!25}80.63 & \cellcolor{yellow!25}5.073 & \cellcolor{yellow!25}11.26 & \cellcolor{yellow!25}0.3845 & \cellcolor{orange!25}0.8049 \\
        & Ours & \cellcolor{red!25}26.10 & \cellcolor{red!25}2.000 & \cellcolor{red!25}12.31 & \cellcolor{orange!25}0.3886 & \cellcolor{red!25}0.7410 \\
        \midrule
        \multirow{5}{*}{JAX\_260} 
        & Sat-NeRF & \cellcolor{yellow!25}86.28 & \cellcolor{orange!25}4.819 & 9.52 & 0.3178 & 0.9050 \\
        & EOGS & 86.87 & 5.378 & 7.04 & 0.1568 & 0.9319 \\
        & CoR-GS & \cellcolor{orange!25}84.13 & 5.530 & \cellcolor{yellow!25}11.50 & \cellcolor{red!25}0.4164 & \cellcolor{yellow!25}0.8976 \\
        & Mip-Splatting & 87.68 & \cellcolor{yellow!25}5.333 & \cellcolor{orange!25}11.64 & \cellcolor{yellow!25}0.3584 & \cellcolor{orange!25}0.8136 \\
        & Ours & \cellcolor{red!25}29.58 & \cellcolor{red!25}2.067 & \cellcolor{red!25}12.59 & \cellcolor{orange!25}0.3589 & \cellcolor{red!25}0.7532 \\
        \bottomrule
    \end{tabular}
\end{table*}

\begin{table*}[t]
    \centering
    \caption{\textbf{Quantitative comparison on each scenes of the GoogleEarth dataset~\cite{xie2024citydreamer}.} 
    The results show that our approach consistently achieves highly competitive performance, indicating superior geometric and perceptual fidelity across most metrics compared to the baselines. Metrics are computed between renders from each method and reference frames from GES. \colorbox{red!25}{Red}, \colorbox{orange!25}{orange}, and \colorbox{yellow!25}{yellow} indicate the best, second best, and third best results, respectively.
    }
    \label{tab:cor_gs_comparison}
    \begin{tabular}{llccccc}
        \toprule
        & & \multicolumn{2}{c}{Distribution Metrics} & \multicolumn{3}{c}{Pixel-level Metrics} \\
        \cmidrule(lr){3-4} \cmidrule(lr){5-7}
        Scene & Methods & $\text{FID}_{\text{CLIP}}\, \downarrow$ & CMMD$\, \downarrow$& PSNR$\, \uparrow$ & SSIM$\, \uparrow$ & LPIPS$\, \downarrow$  \\
        \midrule
        \multirow{5}{*}{004} 
        & CityDreamer & 34.54 & 4.297 & 13.06 & 0.3519 & 0.5643 \\
        & GaussianCity & 29.76 & \cellcolor{yellow!25}2.833 & \cellcolor{yellow!25}14.00 & \cellcolor{yellow!25}0.3785 & 0.5654 \\
        & CoR-GS & \cellcolor{yellow!25}29.17 & 4.092 & 13.57 & 0.3760 & \cellcolor{yellow!25}0.4426 \\
        & Mip-Splatting & \cellcolor{orange!25}24.57 & \cellcolor{orange!25}2.611 & \cellcolor{orange!25}14.59 & \cellcolor{orange!25}0.3823 & \cellcolor{orange!25}0.4278 \\
        & Ours & \cellcolor{red!25}14.81 & \cellcolor{red!25}1.549 & \cellcolor{red!25}15.22 & \cellcolor{red!25}0.3859 & \cellcolor{red!25}0.4117 \\
        \midrule
        \multirow{5}{*}{010} 
        & CityDreamer & 39.95 & 3.948 & 12.24 & 0.1387 & 0.5541 \\
        & GaussianCity & \cellcolor{yellow!25}28.65 & \cellcolor{yellow!25}2.715 & 12.90 & 0.1661 & 0.5330 \\
        & CoR-GS & 30.10 & 3.741 & \cellcolor{yellow!25}12.90 & \cellcolor{orange!25}0.1807 & \cellcolor{yellow!25}0.4209 \\
        & Mip-Splatting & \cellcolor{orange!25}11.55 & \cellcolor{red!25}1.914 & \cellcolor{orange!25}13.63 & \cellcolor{red!25}0.1823 & \cellcolor{red!25}0.3514 \\
        & Ours & \cellcolor{red!25}9.43 & \cellcolor{orange!25}2.484 & \cellcolor{red!25}13.71 & \cellcolor{yellow!25}0.1800 & \cellcolor{orange!25}0.3959 \\
        \midrule
        \multirow{5}{*}{219} 
        & CityDreamer & 42.54 & 4.444 & 11.63 & 0.1344 & 0.5465 \\
        & GaussianCity & 32.78 & \cellcolor{yellow!25}2.898 & 12.37 & 0.1676 & 0.5248 \\
        & CoR-GS & \cellcolor{yellow!25}16.39 & 3.227 & \cellcolor{yellow!25}12.64 & \cellcolor{orange!25}0.1791 & \cellcolor{yellow!25}0.3971 \\
        & Mip-Splatting & \cellcolor{red!25}7.06 & \cellcolor{red!25}1.589 & \cellcolor{orange!25}13.26 & \cellcolor{red!25}0.1795 & \cellcolor{red!25}0.3321 \\
        & Ours & \cellcolor{orange!25}7.80 & \cellcolor{orange!25}2.640 & \cellcolor{red!25}13.32 & \cellcolor{yellow!25}0.1764 & \cellcolor{orange!25}0.3840 \\
        \midrule
        \multirow{5}{*}{336} 
        & CityDreamer & 29.59 & 4.110 & 13.39 & 0.4430 & 0.5654 \\
        & GaussianCity & \cellcolor{yellow!25}23.87 & \cellcolor{yellow!25}3.215 & \cellcolor{yellow!25}14.36 & 0.4532 & 0.5376 \\
        & CoR-GS & 29.74 & 3.970 & 14.29 & \cellcolor{yellow!25}0.4591 & \cellcolor{orange!25}0.3875 \\
        & Mip-Splatting & \cellcolor{orange!25}21.20 & \cellcolor{orange!25}2.232 & \cellcolor{orange!25}15.05 & \cellcolor{orange!25}0.4631 & \cellcolor{yellow!25}0.4039 \\
        & Ours & \cellcolor{red!25}9.11 & \cellcolor{red!25}1.164 & \cellcolor{red!25}15.45 & \cellcolor{red!25}0.4664 & \cellcolor{red!25}0.3805 \\
        \bottomrule
    \end{tabular}
\end{table*}

\subsubsection{Reprojection consistency to satellite views.}
A potential concern is whether the diffusion-based IDU stage drifts away from the geometry anchored by the input satellite imagery. To verify this, we evaluate \emph{reprojection} quality: we render each method at the held-out satellite camera poses (4 AOIs, 62 views in total) and compare against the corresponding satellite observations. As shown in \Cref{tab:reprojection}, our method matches or exceeds the strongest reconstruction baseline, Mip-Splatting~\cite{yu2024mip}, across PSNR, SSIM, and LPIPS. This confirms that IDU adds close-up facade detail \emph{without} sacrificing fidelity to the satellite-anchored geometry, which we attribute to our $75/25$ sampling of refined and original satellite views during optimization.

\begin{table}[t]
    \centering
    \caption{\textbf{Reprojection consistency to the input satellite views} (4 AOIs, 62 views). Rendering each method at the satellite camera poses and comparing against the satellite observations, our method matches or exceeds Mip-Splatting, indicating that IDU refinement does not drift from the satellite-anchored geometry. \colorbox{red!25}{Red}, \colorbox{orange!25}{orange}, and \colorbox{yellow!25}{yellow} indicate the best, second best, and third best results, respectively.}
    \label{tab:reprojection}
    \begin{tabular}{lccc}
        \toprule
        Method & PSNR$\,\uparrow$ & SSIM$\,\uparrow$ & LPIPS$\,\downarrow$ \\
        \midrule
        Sat-NeRF~\cite{mari2022sat} & \cellcolor{yellow!25}29.22 & \cellcolor{yellow!25}0.868 & \cellcolor{yellow!25}0.219 \\
        EOGS~\cite{savantaira2024eogs} & 25.26 & 0.823 & 0.260 \\
        CoR-GS~\cite{zhang2024cor} & 24.82 & 0.773 & 0.344 \\
        Mip-Splatting~\cite{yu2024mip} & \cellcolor{orange!25}29.24 & \cellcolor{red!25}0.936 & \cellcolor{red!25}0.121 \\
        \textbf{Ours} & \cellcolor{red!25}29.67 & \cellcolor{orange!25}0.934 & \cellcolor{orange!25}0.129 \\
        \bottomrule
    \end{tabular}
\end{table}

\subsubsection{Additional qualitative comparisons.}
\begin{figure*}[t]
    \centering
    \includegraphics[width=\linewidth]{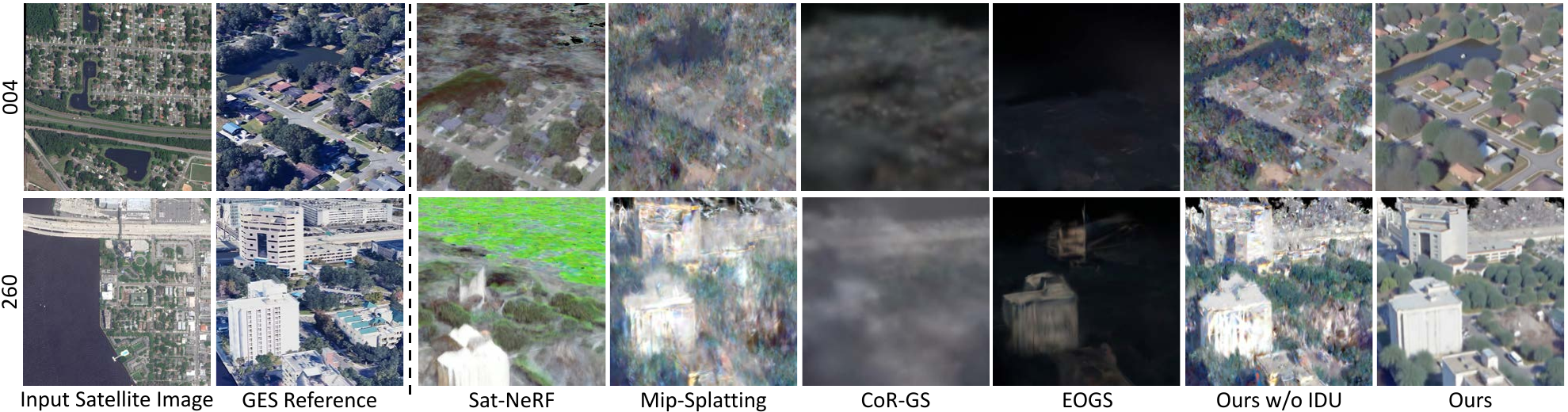}
    \caption{\textbf{Additional qualitative comparison on the DFC2019 dataset with Sat-NeRF~\cite{mari2022sat}, Mip-Splatting~\cite{yu2024mip}, CoR-GS~\cite{zhang2024cor}, and EOGS~\cite{savantaira2024eogs}.} Our method significantly outperforms baseline approaches in both geometric accuracy and texture quality when rendering low-altitude novel views. Note the superior building geometry, facade details, and reduced floating artifacts in our final result.
    }
    \label{fig:qualitative-3d-suppl}
\end{figure*}

\begin{figure*}[t]
    \centering
    \includegraphics[width=\linewidth]{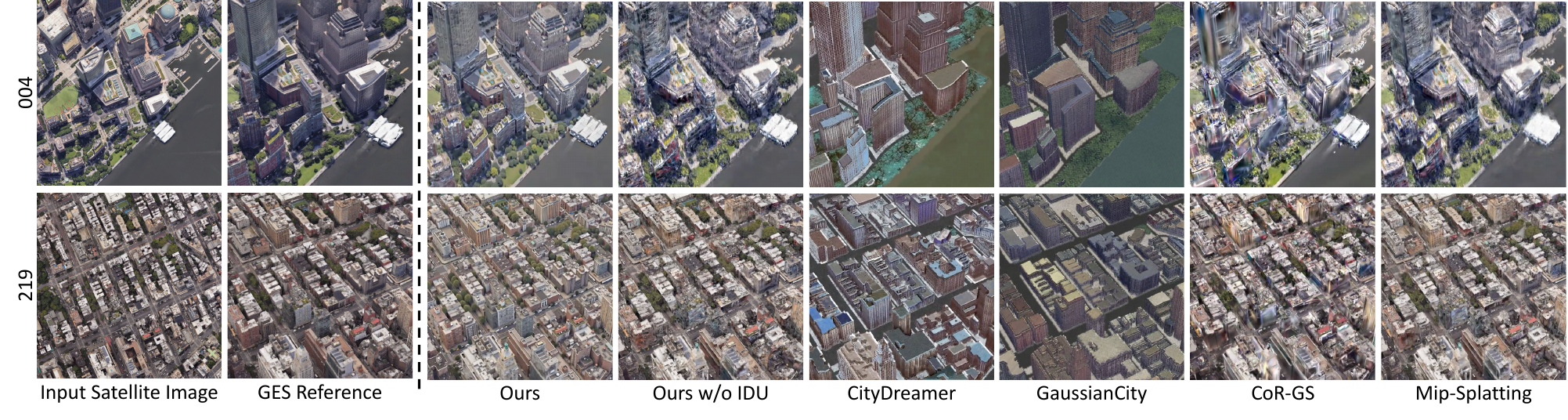}
    \caption{\textbf{Additional qualitative comparison on the GoogleEarth dataset with CityDreamer~\cite{xie2024citydreamer}, GaussianCity~\cite{xie2025gaussiancity}, CoR-GS~\cite{zhang2024cor}, and Mip-Splatting~\cite{yu2024mip}.} Our method is able to synthesize texture and geometry that is closer to the reference GES render.
    }
    \label{fig:qualitative-city-suppl}
\end{figure*}

Due to space constraints in the main paper, we present additional qualitative comparison results in this supplementary material for scenes JAX\_004 and JAX\_260 from the DFC2019 dataset, and scenes 004 and 219 from the GoogleEarth dataset. \Cref{fig:qualitative-3d-suppl} shows orbital view comparisons with Sat-NeRF~\cite{mari2022sat}, Mip-Splatting~\cite{yu2024mip}, CoR-GS~\cite{zhang2024cor}, and EOGS~\cite{savantaira2024eogs}, while \Cref{fig:qualitative-city-suppl} presents city-scale view comparisons with CityDreamer~\cite{xie2024citydreamer}, GaussianCity~\cite{xie2025gaussiancity}, and CoR-GS~\cite{zhang2024cor}. These additional results further demonstrate the consistent superiority of our method across diverse urban environments.

\begin{figure*}[t]
  \centering
  \setlength{\tabcolsep}{1pt}
  
  \newcommand{\imgw}{0.22\textwidth}

  \begin{tabular}{@{} m{.75em} m{\imgw} @{\hspace{4pt}} m{\imgw} m{\imgw} m{\imgw} @{}}
  & 
  \centering\textbf{\scriptsize Input Satellite Image} & 
  \multicolumn{3}{c}{\textbf{\scriptsize 3DGS Render}} \\
  \cmidrule(l){3-5}

  \rotatebox{90}{\scriptsize 004} &
  \includegraphics[width=\linewidth]{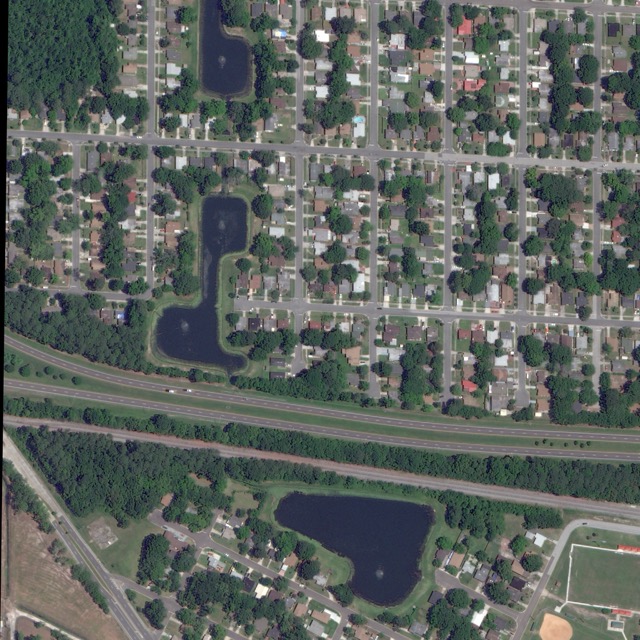} &
  \includegraphics[width=\linewidth]{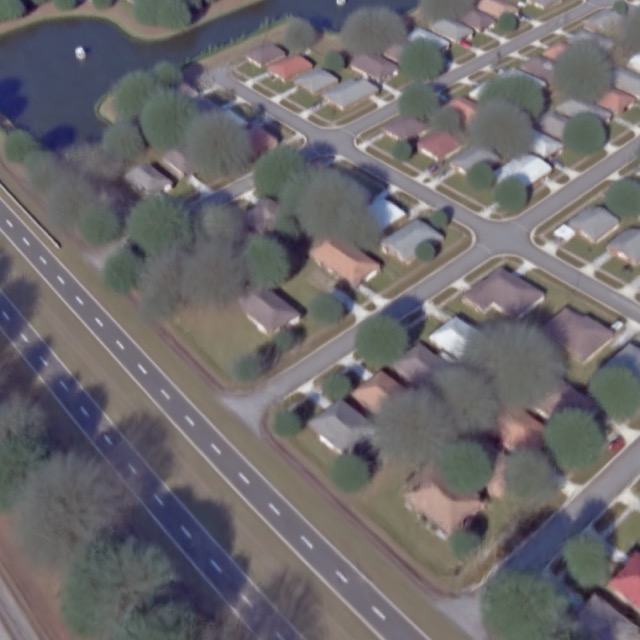} &
  \includegraphics[width=\linewidth]{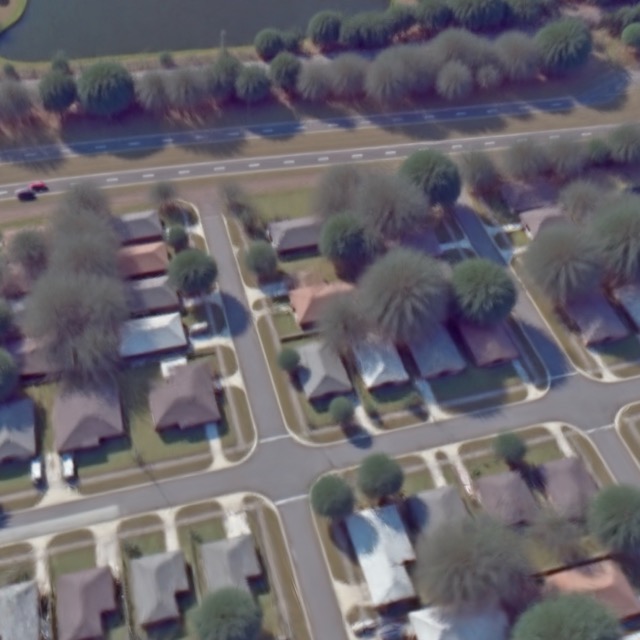} &
  \includegraphics[width=\linewidth]{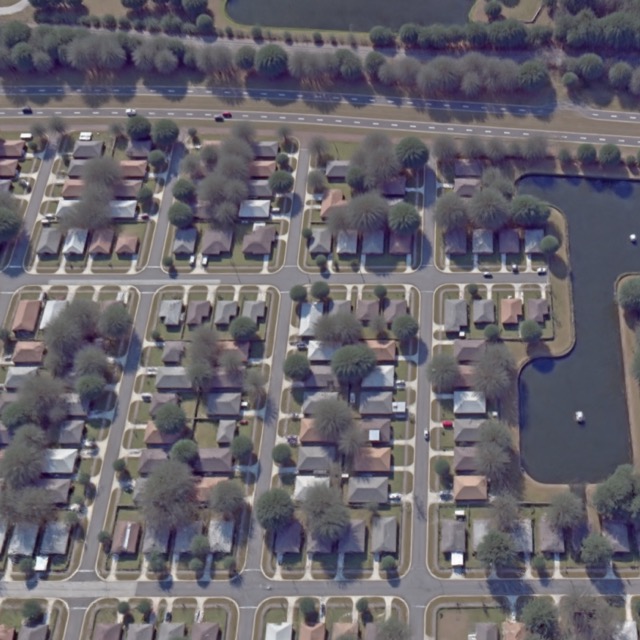} \\

  \rotatebox{90}{\scriptsize 068} &
  \includegraphics[width=\linewidth]{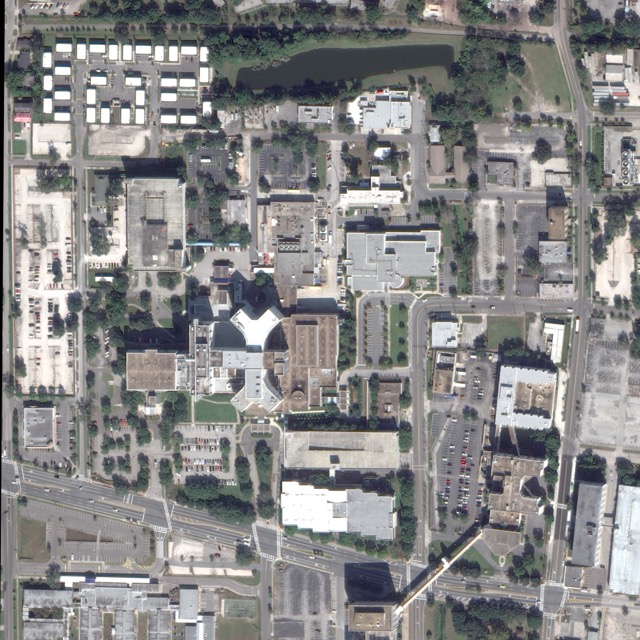} &
  \includegraphics[width=\linewidth]{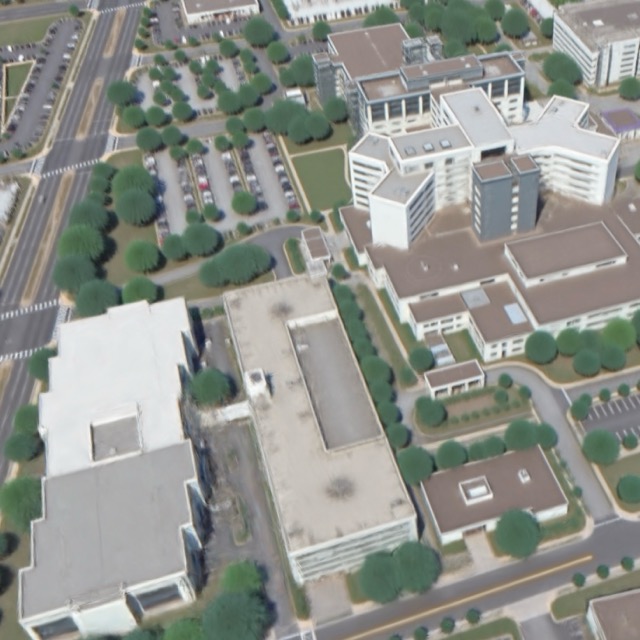} &
  \includegraphics[width=\linewidth]{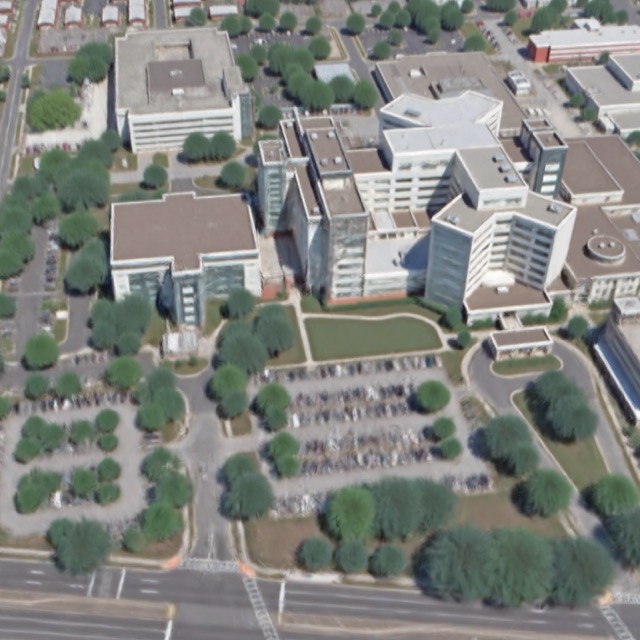} &
  \includegraphics[width=\linewidth]{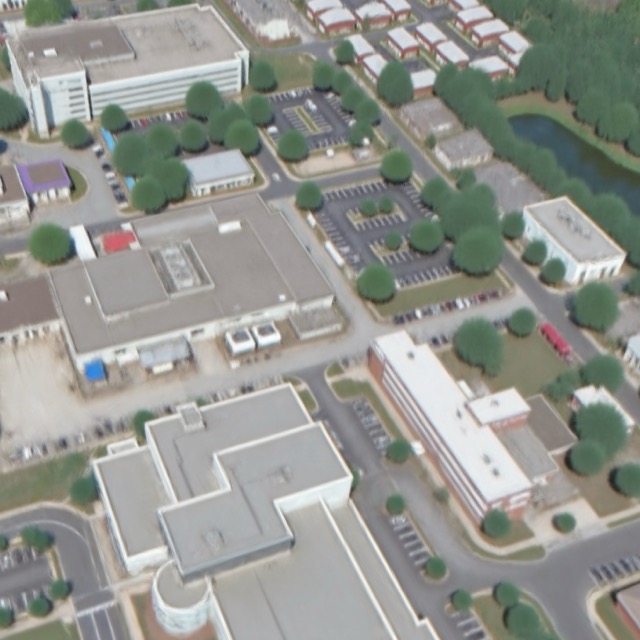} \\

  \rotatebox{90}{\scriptsize 214} &
  \includegraphics[width=\linewidth]{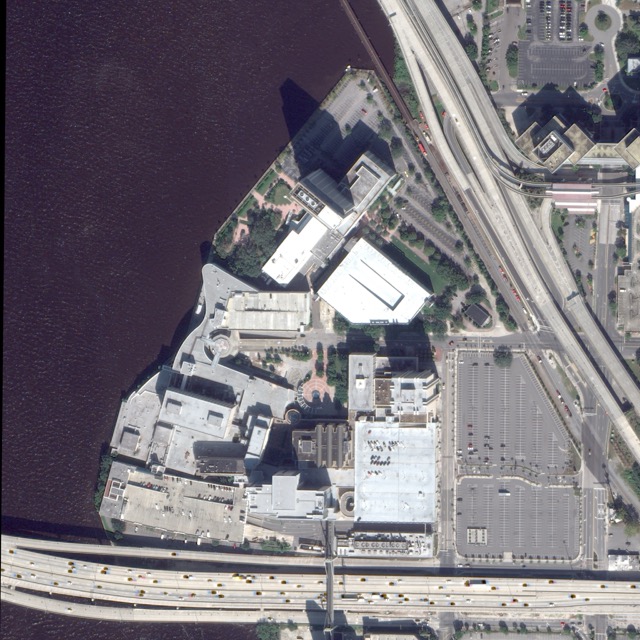} &
  \includegraphics[width=\linewidth]{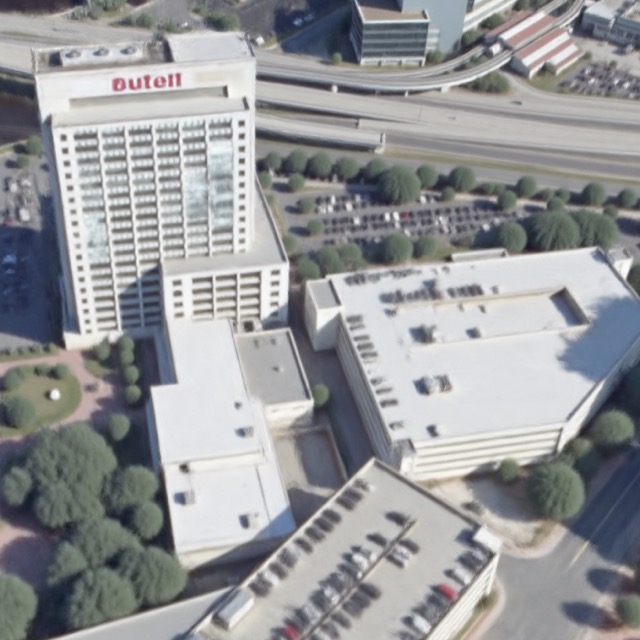} &
  \includegraphics[width=\linewidth]{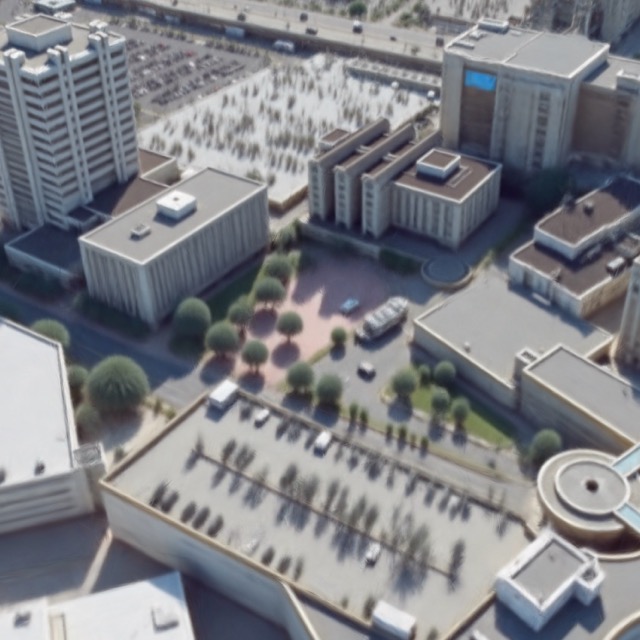} &
  \includegraphics[width=\linewidth]{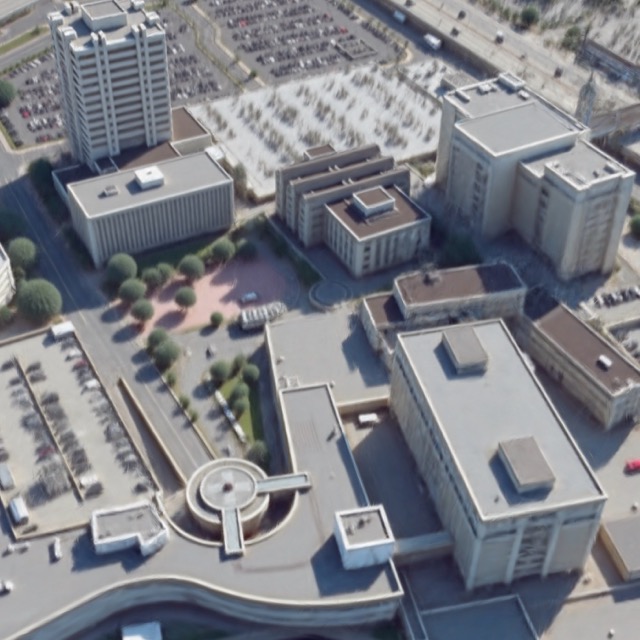} \\

  \rotatebox{90}{\scriptsize 260} &
  \includegraphics[width=\linewidth]{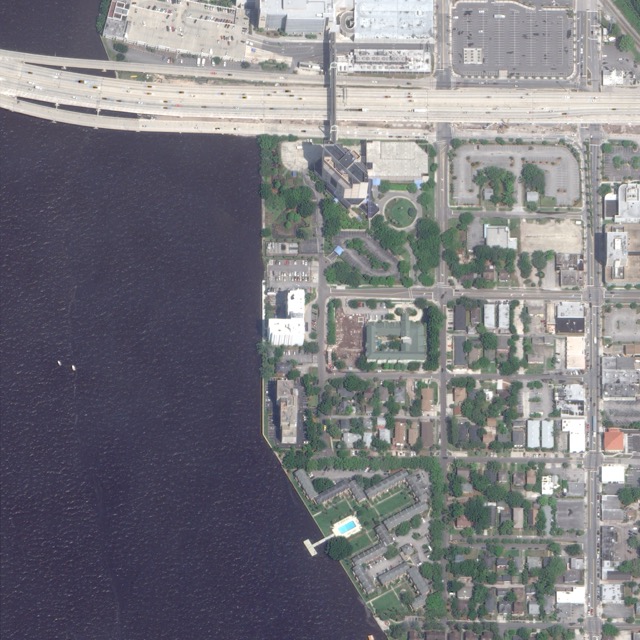} &
  \includegraphics[width=\linewidth]{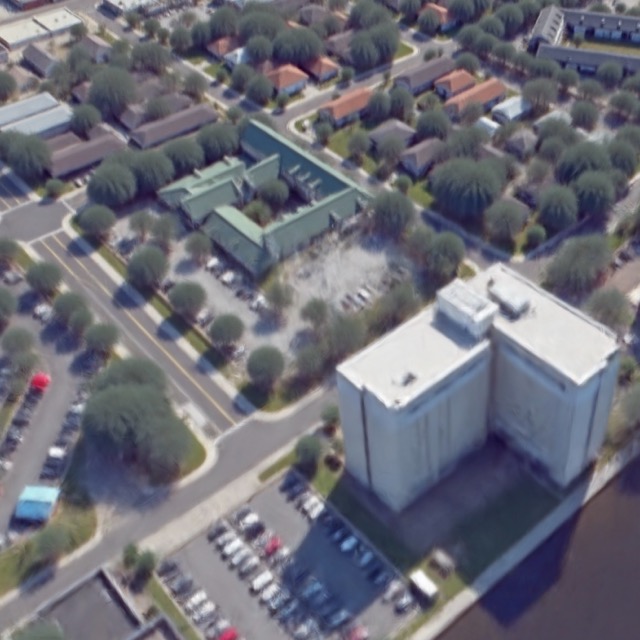} &
  \includegraphics[width=\linewidth]{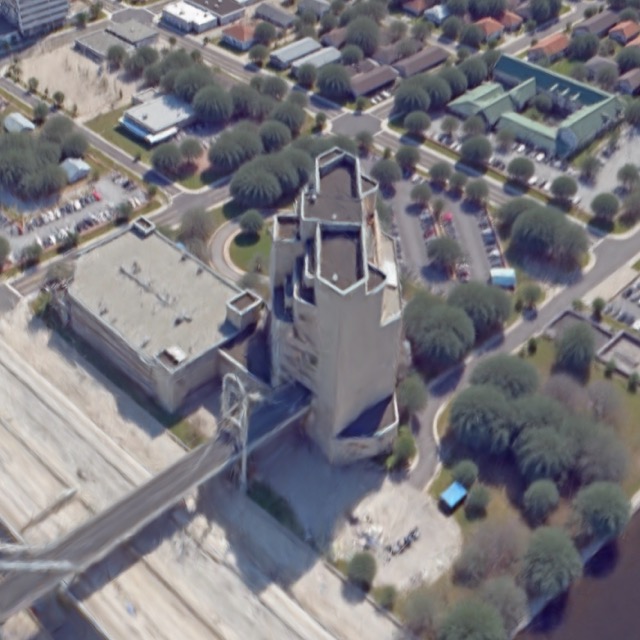} &
  \includegraphics[width=\linewidth]{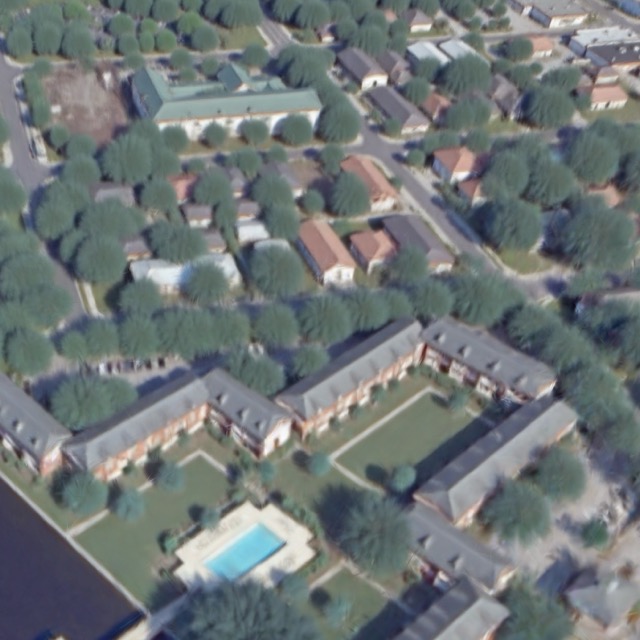} \\

  \end{tabular}
  \caption{\textbf{Qualitative results across primary scenes.} Visualization of satellite image inputs and corresponding rendered frames for our four main AOIs.}
  \label{tab:suppl_visual_1}
\end{figure*}

\begin{figure*}[t]
  \centering
  \setlength{\tabcolsep}{1pt}
  
  \newcommand{\imgw}{0.22\textwidth}

  \begin{tabular}{@{} m{.75em} m{\imgw} @{\hspace{4pt}} m{\imgw} m{\imgw} m{\imgw} @{}}
  & 
  \centering\textbf{\scriptsize Input Satellite Image} & 
  \multicolumn{3}{c}{\textbf{\scriptsize 3DGS Render}} \\
  \cmidrule(l){3-5}

  \rotatebox{90}{\scriptsize 164} &
  \includegraphics[width=\linewidth]{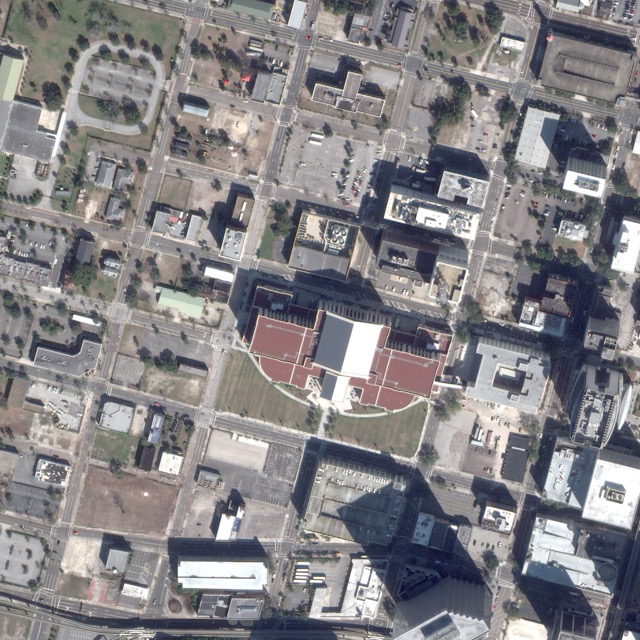} &
  \includegraphics[width=\linewidth]{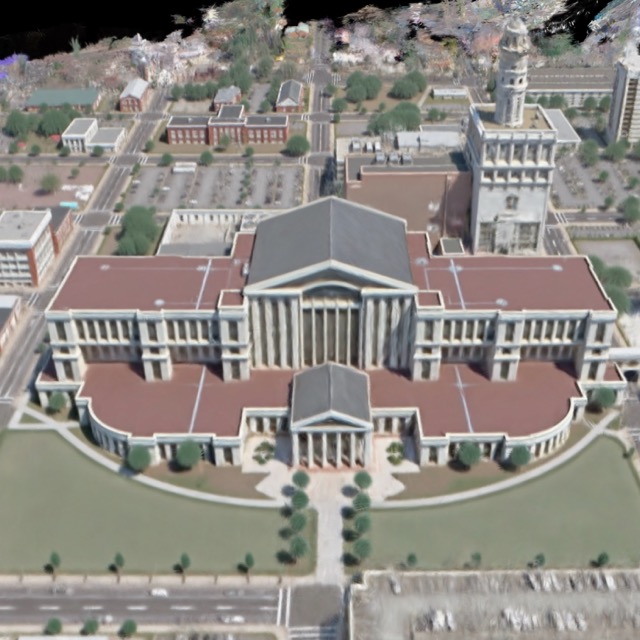} &
  \includegraphics[width=\linewidth]{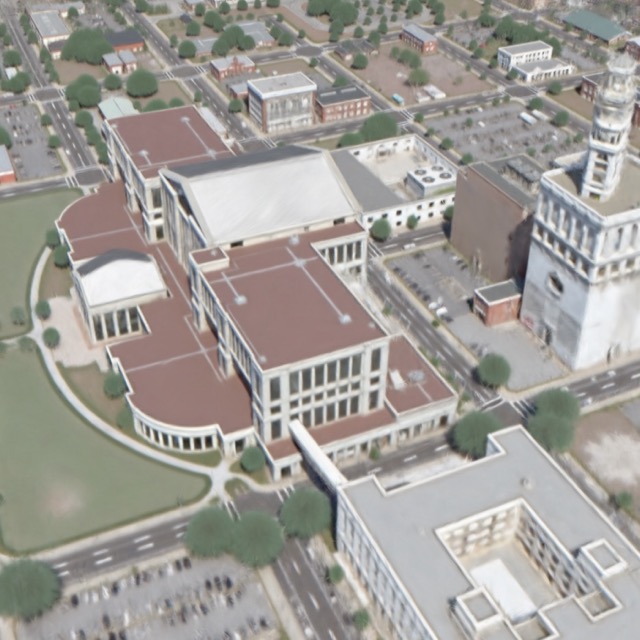} &
  \includegraphics[width=\linewidth]{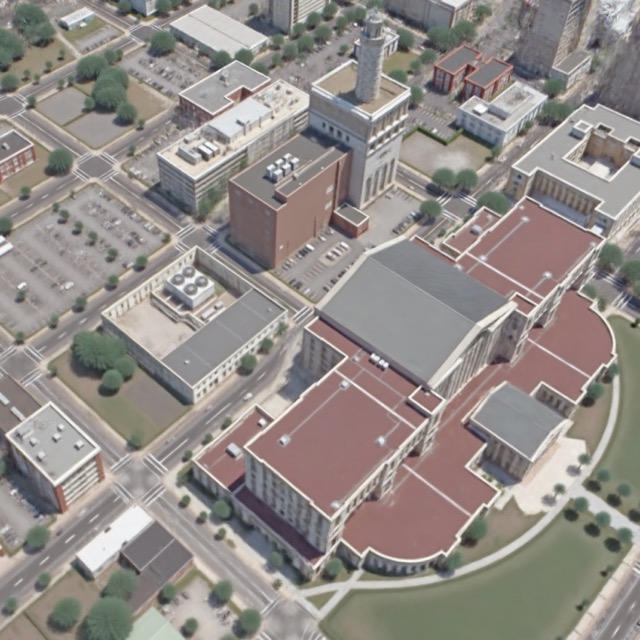} \\

  \rotatebox{90}{\scriptsize 168} &
  \includegraphics[width=\linewidth]{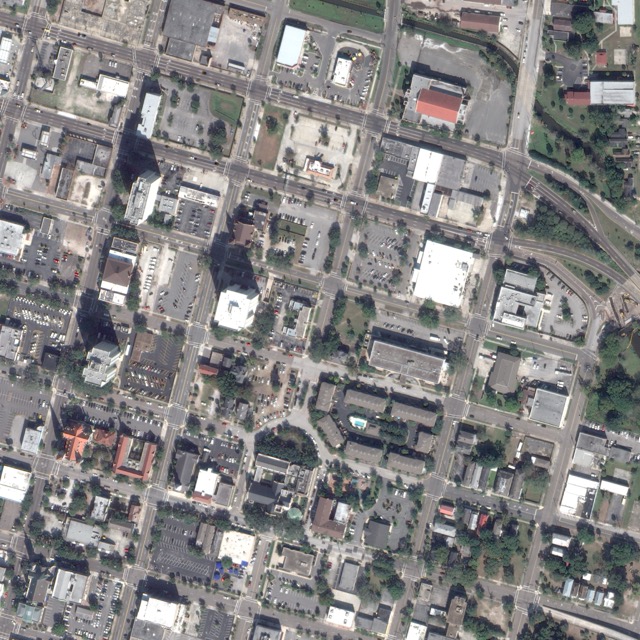} &
  \includegraphics[width=\linewidth]{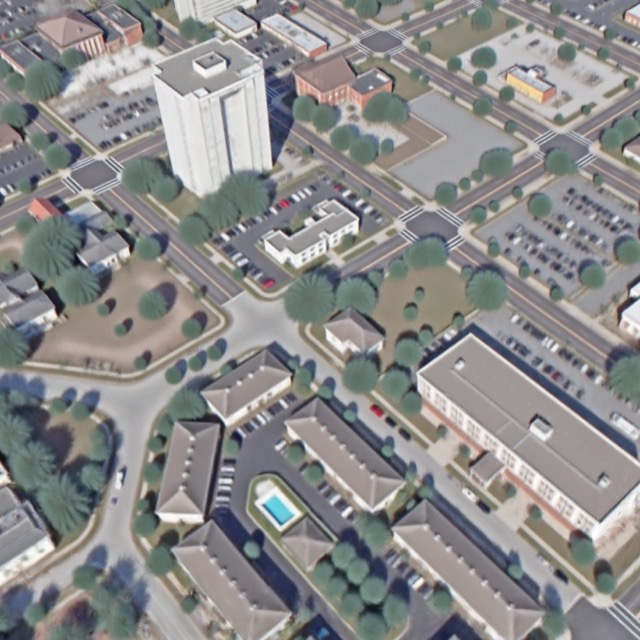} &
  \includegraphics[width=\linewidth]{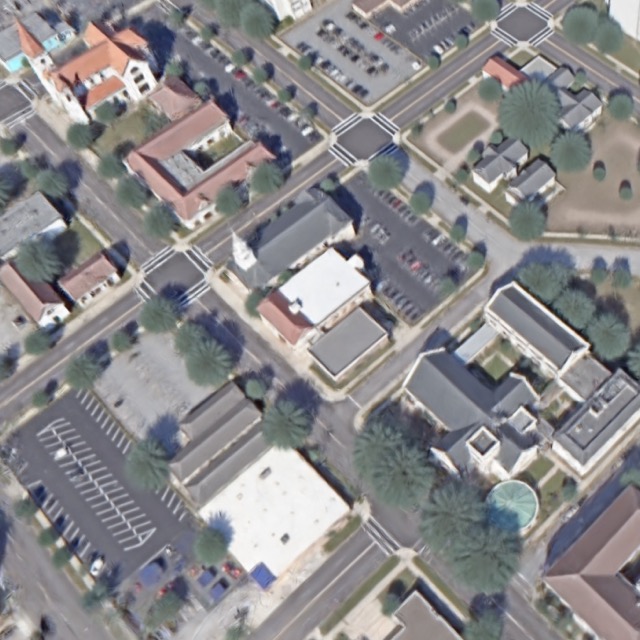} &
  \includegraphics[width=\linewidth]{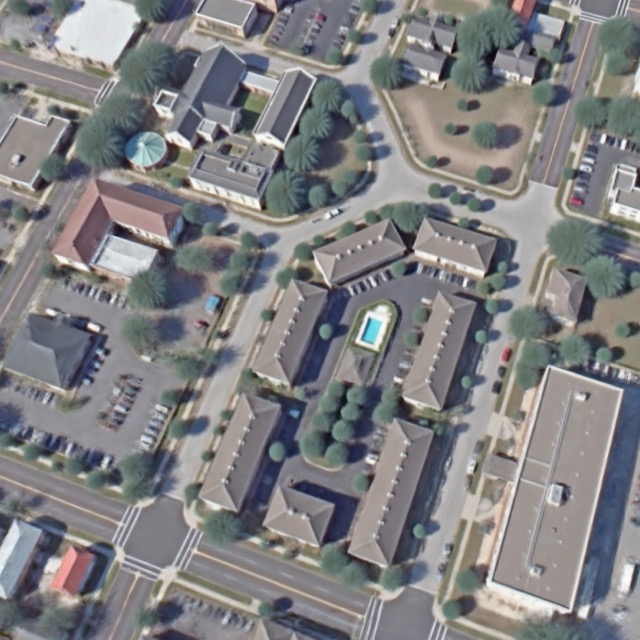} \\

  \rotatebox{90}{\scriptsize 175} &
  \includegraphics[width=\linewidth]{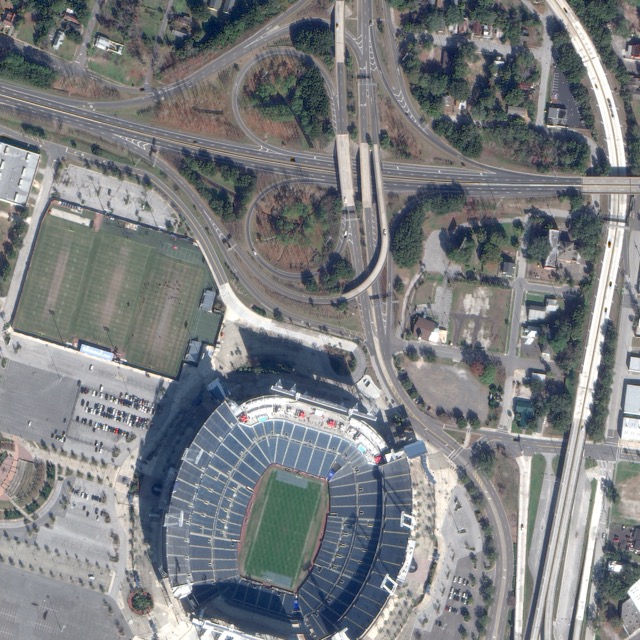} &
  \includegraphics[width=\linewidth]{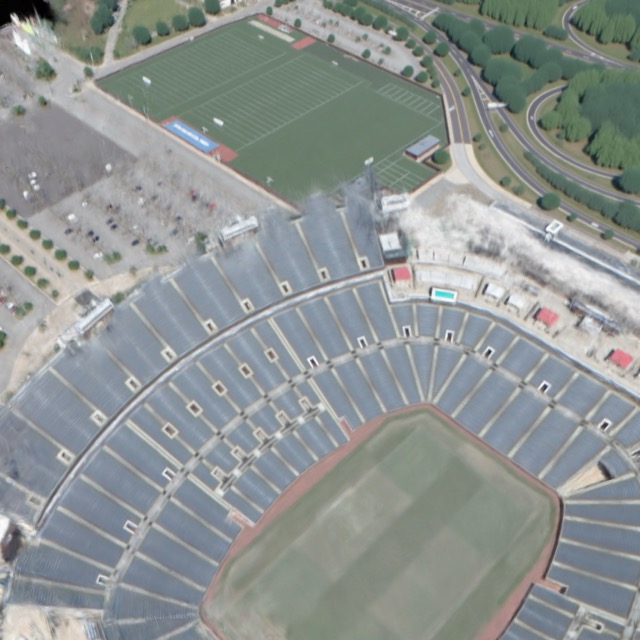} &
  \includegraphics[width=\linewidth]{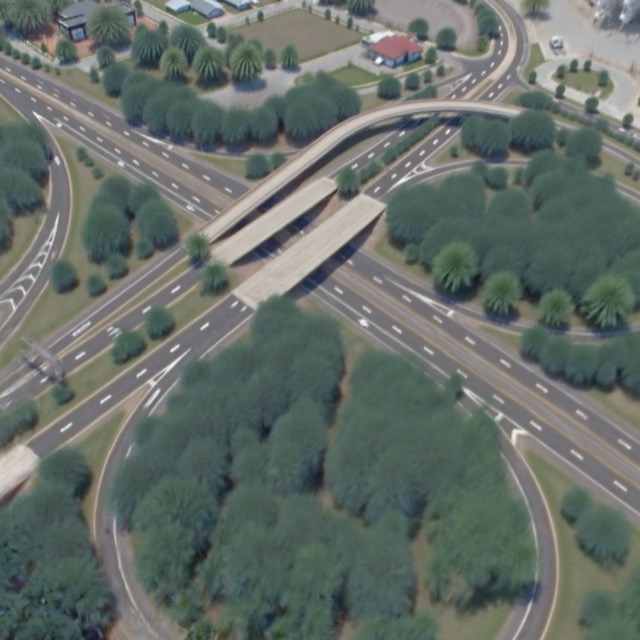} &
  \includegraphics[width=\linewidth]{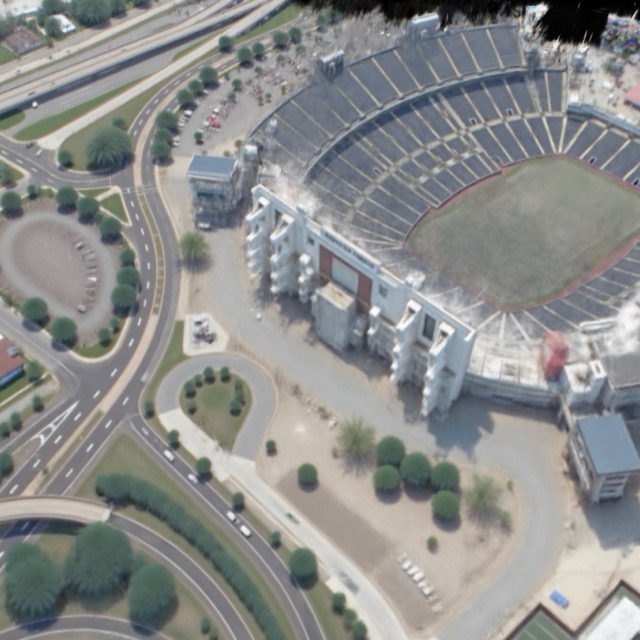} \\

  \rotatebox{90}{\scriptsize 264} &
  \includegraphics[width=\linewidth]{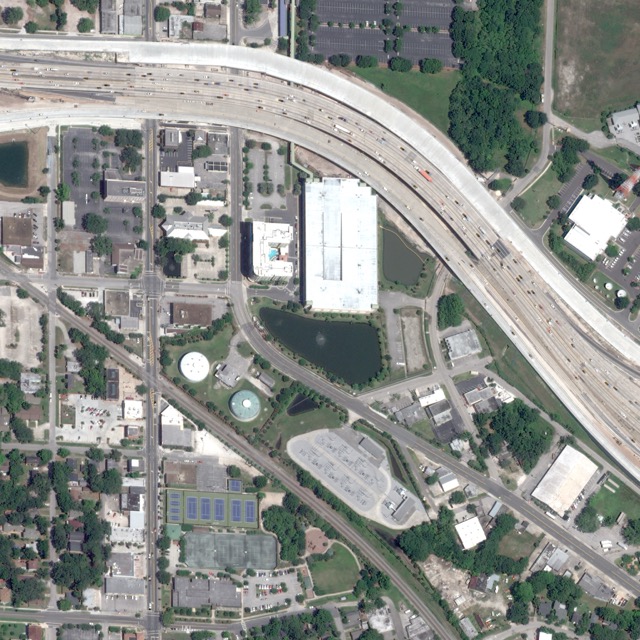} &
  \includegraphics[width=\linewidth]{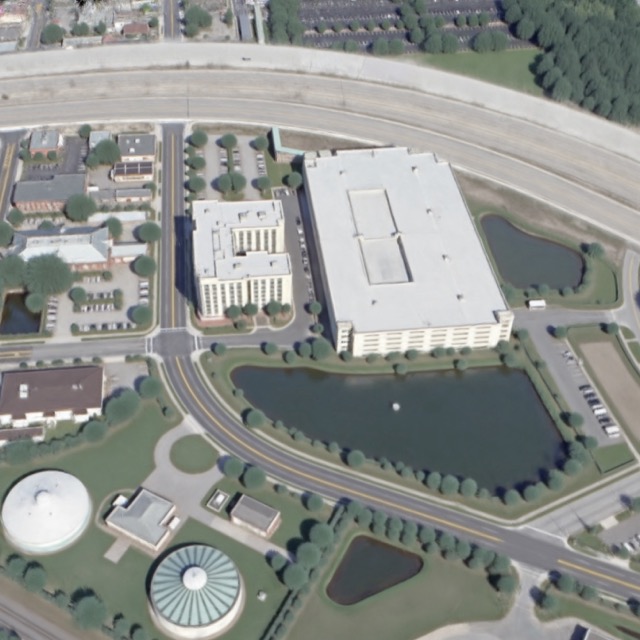} &
  \includegraphics[width=\linewidth]{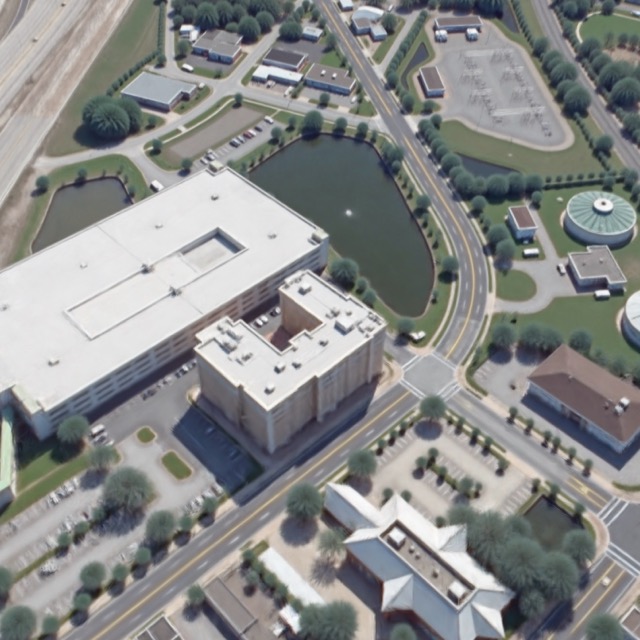} &
  \includegraphics[width=\linewidth]{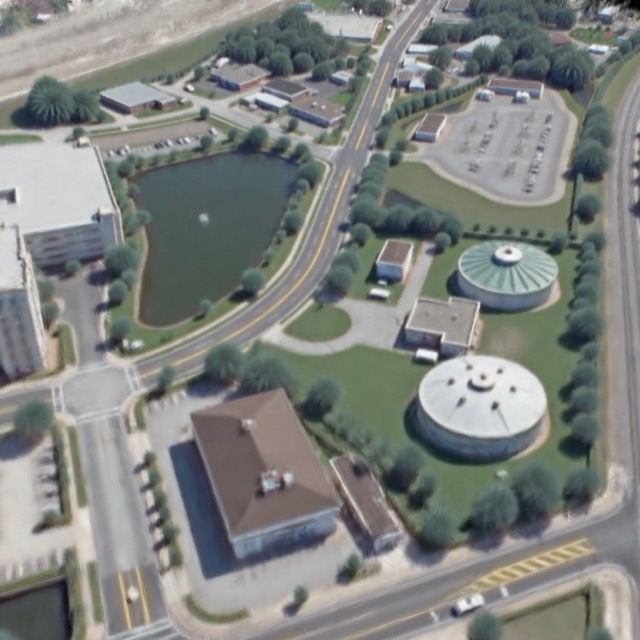} \\

  \end{tabular}
  \caption{\textbf{Qualitative results across additional scenes.} Visualization of satellite image inputs and corresponding rendered frames for four additional AOIs with distinctive characteristics: JAX\_164 features a city hall building, JAX\_175 contains an American football stadium, while JAX\_168 and JAX\_264 present other notable urban structures.}
  \label{tab:suppl_visual_2}
\end{figure*}

\begin{figure*}[t]
    \centering
    \includegraphics[width=0.95\linewidth]{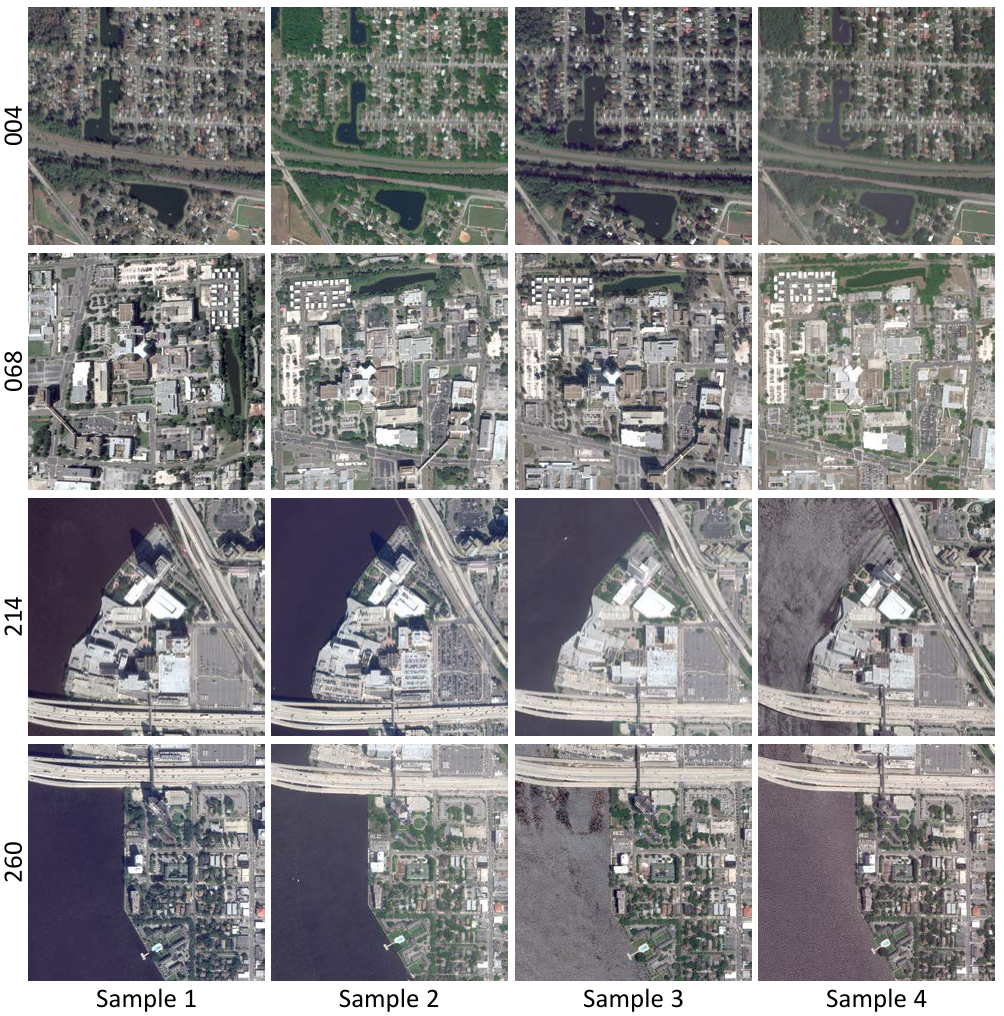}
    \caption{\textbf{Visualization of multi-date satellite imagery of the DFC2019 dataset.} 
    Note the substantial shifts in appearance, including changes in illumination, transient objects, and surface characteristics, which introduce challenges for consistent 3D reconstruction.
    } 
    \label{fig:multi-data-jax-suppl}
\end{figure*}

\subsubsection{Additional visual results.}
We also provide qualitative results on four AOIs presents on the main paper (\Cref{tab:num_train}), as well as four additional AOIs (\Cref{tab:num_train_additional}) from Jacksonville to demonstrate our method's robustness across diverse urban environments. As shown in \Cref{tab:suppl_visual_1} and \Cref{tab:suppl_visual_2}, these AOIs contain distinctive architectural features: JAX\_004 showcases a residential area with mixed housing types and green spaces; JAX\_164 features a prominent city hall building with its characteristic dome and symmetrical facade; JAX\_175 encompasses an American football stadium with its distinctive oval structure and surrounding parking facilities; JAX\_168 contains a commercial district with varied building heights and dense urban layout. Despite these varied urban typologies, our method successfully generates coherent three-dimensional renderings that preserve the spatial relationships and architectural features present in the satellite imagery. These additional results further validate the generalizability of our approach across diverse urban landscapes without requiring scene-specific parameter adjustments.

\subsubsection{Multi-date appearance variation.}
The use of multi-date satellite imagery introduces a significant challenge, as images of the same location, when captured on different days, exhibit drastic variations in appearance. As shown in Figure~\ref{fig:multi-data-jax-suppl}, these differences can fundamentally alter the scene's geometry and texture. Effectively synthesizing novel views requires a model capable of intelligently disentangling the static 3D scene structure from these challenging, temporally-varying appearance factors.

\section{Additional Experiments} \label{sup:additional-exp}

\subsubsection{Qualitative results on complex geometries.}
To demonstrate the robustness of our framework beyond standard city-block layouts, we evaluate our method on scenes featuring irregular and historically significant architectures. As shown in Figure \ref{tab:suppl_visual_2}, we present synthesis results for \textbf{Neuschwanstein Castle} and \textbf{Wells Cathedral}. These scenes pose significant challenges due to their intricate non-Manhattan geometries, including sharp spires, varying elevations, and gothic architectural details. Despite these complexities, our method successfully disentangles the underlying geometry from the satellite input and hallucinates plausible high-frequency details for facades that are heavily occluded in the nadir views. This confirms that our hybrid reconstruction-generation approach is not limited to simple urban prisms but extends effectively to complex, free-form structures.

\begin{figure*}[t]
  \centering
  \setlength{\tabcolsep}{4pt} %
  \renewcommand{\arraystretch}{1.2} %
  
  \newcommand{\imgw}{0.22\textwidth} 

  \begin{tabular}{@{} m{1.5em} m{\imgw} @{\hspace{10pt}} m{\imgw} m{\imgw} m{\imgw} @{}}
  & 
  \centering\textbf{\scriptsize Input Satellite Image} & 
  \multicolumn{3}{c}{\textbf{\scriptsize 3DGS Rendered Views}} \\
  \cmidrule(l){3-5}

  \rotatebox{90}{\scriptsize Neuschwanstein Castle} &
  \includegraphics[width=\linewidth]{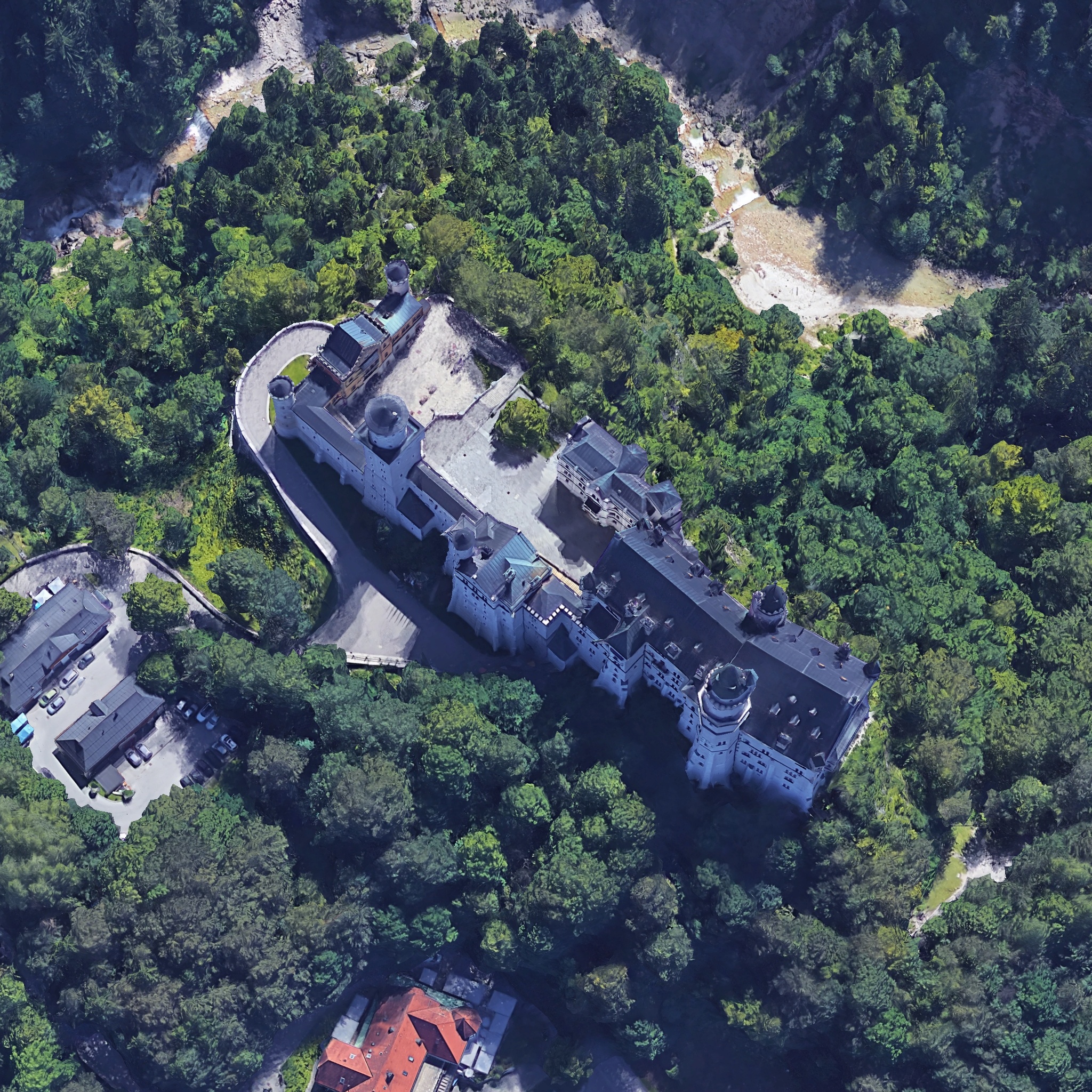} &
  \includegraphics[width=\linewidth]{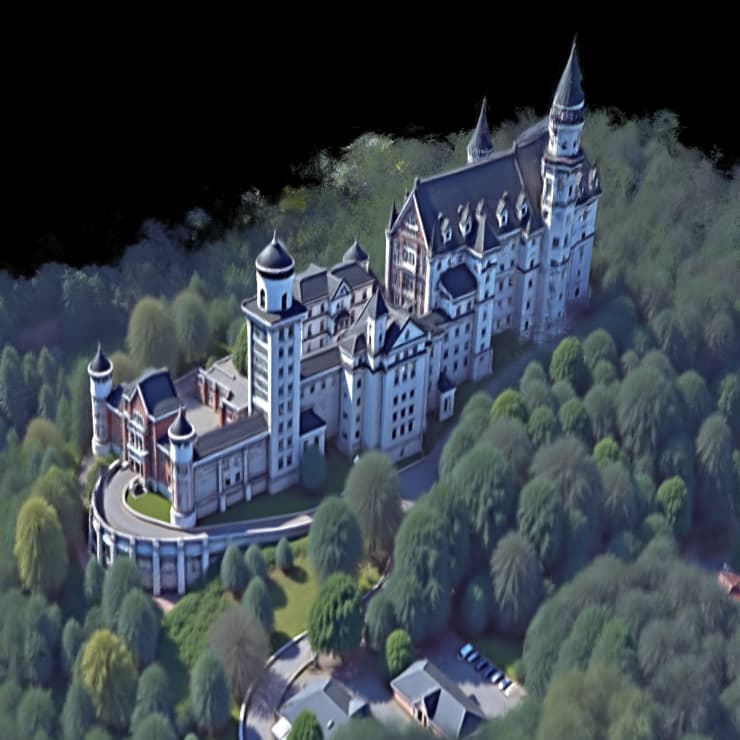} &
  \includegraphics[width=\linewidth]{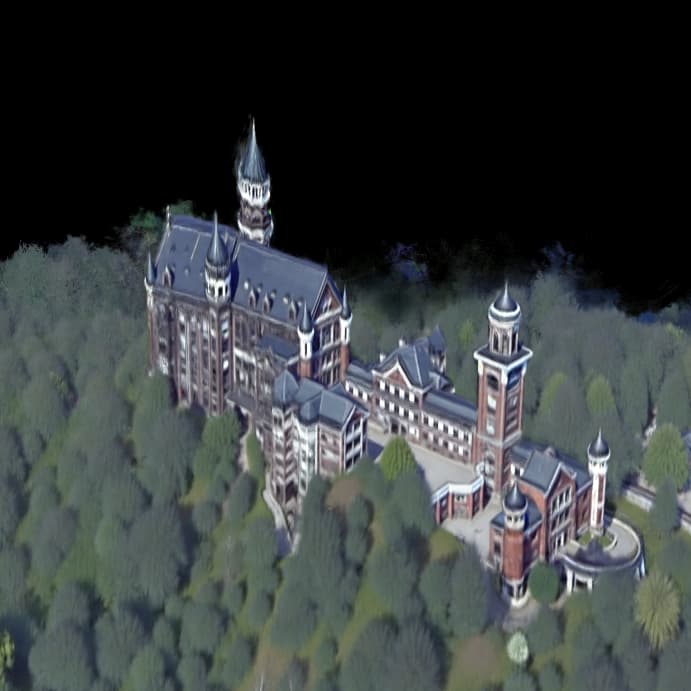} &
  \includegraphics[width=\linewidth]{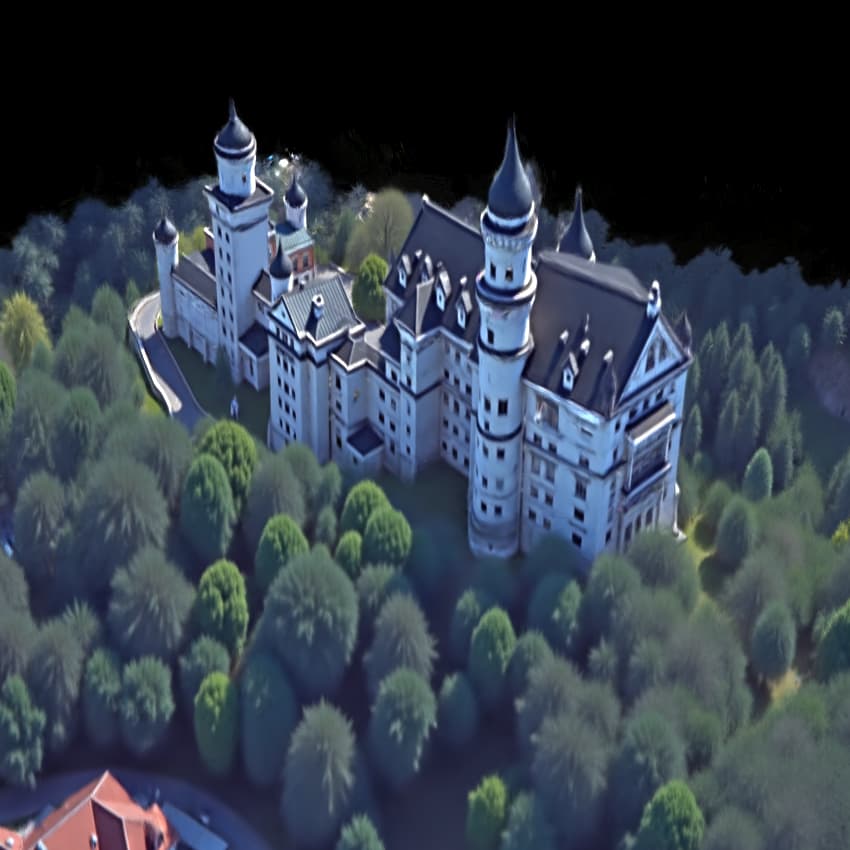} \\

  \addlinespace[6pt] %

  \rotatebox{90}{\scriptsize Wells Cathedral} &
  \includegraphics[width=\linewidth]{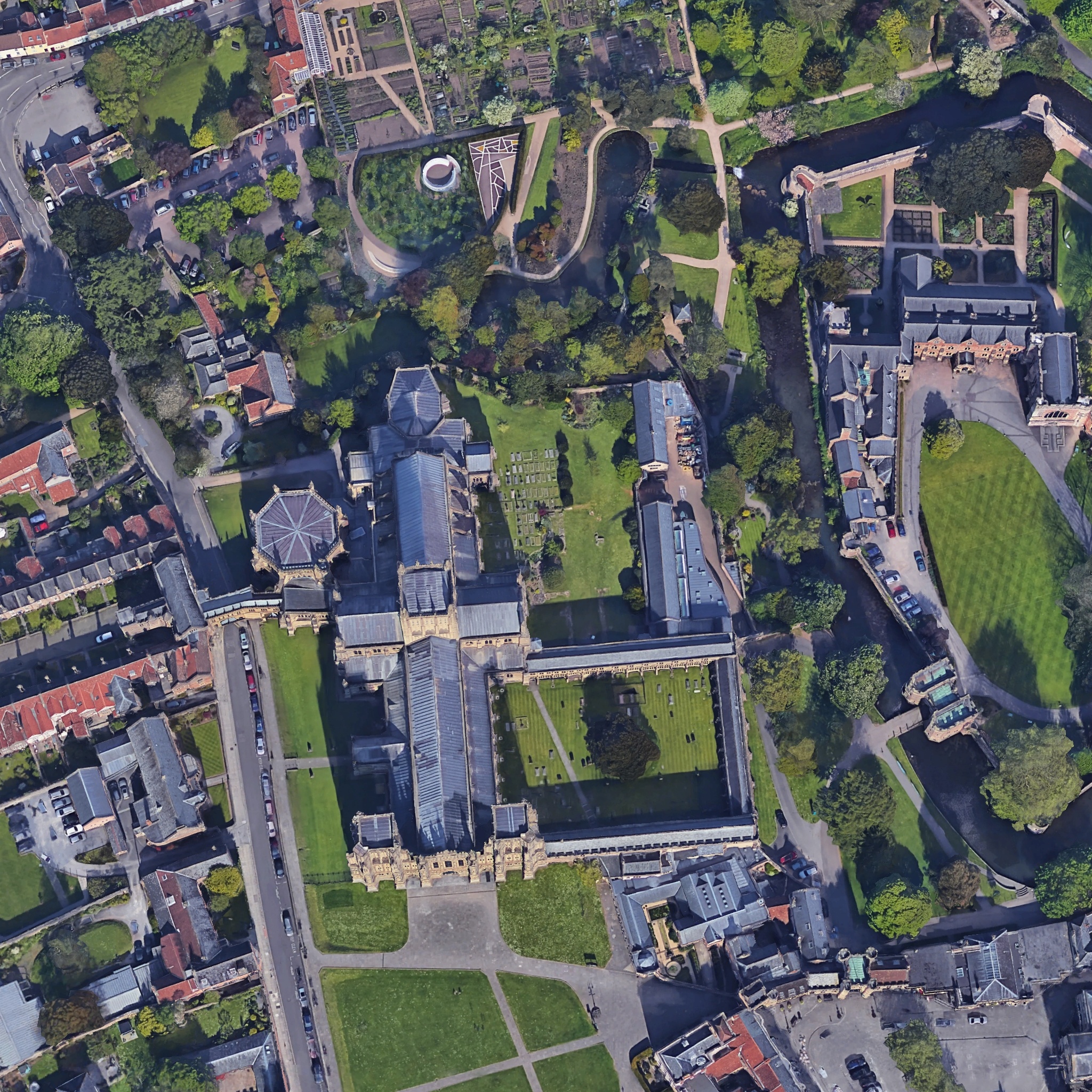} &
  \includegraphics[width=\linewidth]{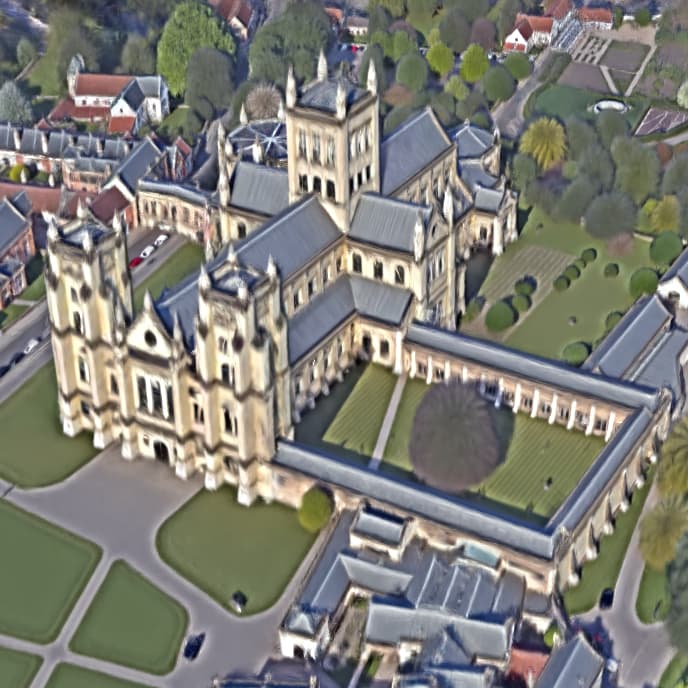} &
  \includegraphics[width=\linewidth]{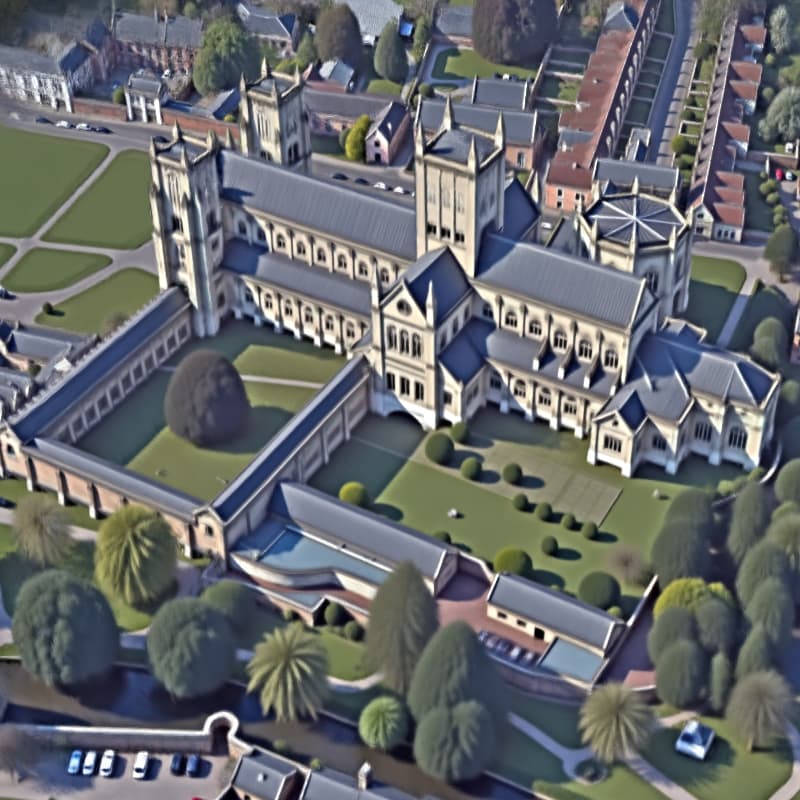} &
  \includegraphics[width=\linewidth]{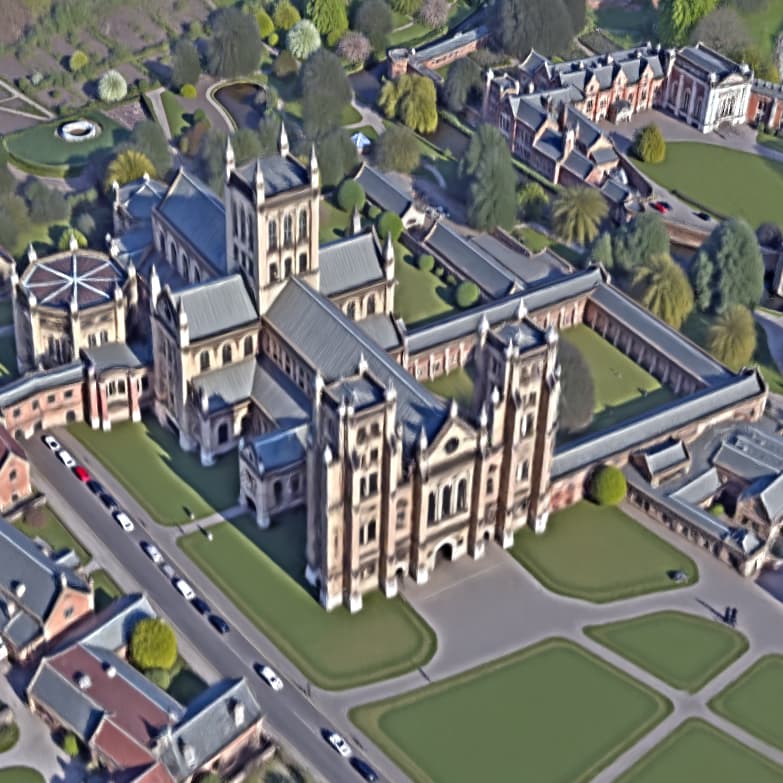} \\
  \end{tabular}

  \caption{\textbf{Qualitative results on complex geometries.}
  Given multi-view satellite imagery, our method synthesizes novel views of irregular historical structures, shown here for Neuschwanstein Castle and Wells Cathedral.}
  \label{fig:suppl_visual_2}
\end{figure*}

\subsubsection{Synthesis of bridges.}
\begin{figure}[t]
\centering
\setlength{\tabcolsep}{2pt}
\begin{tabular}{ccc}
\includegraphics[width=0.31\textwidth]{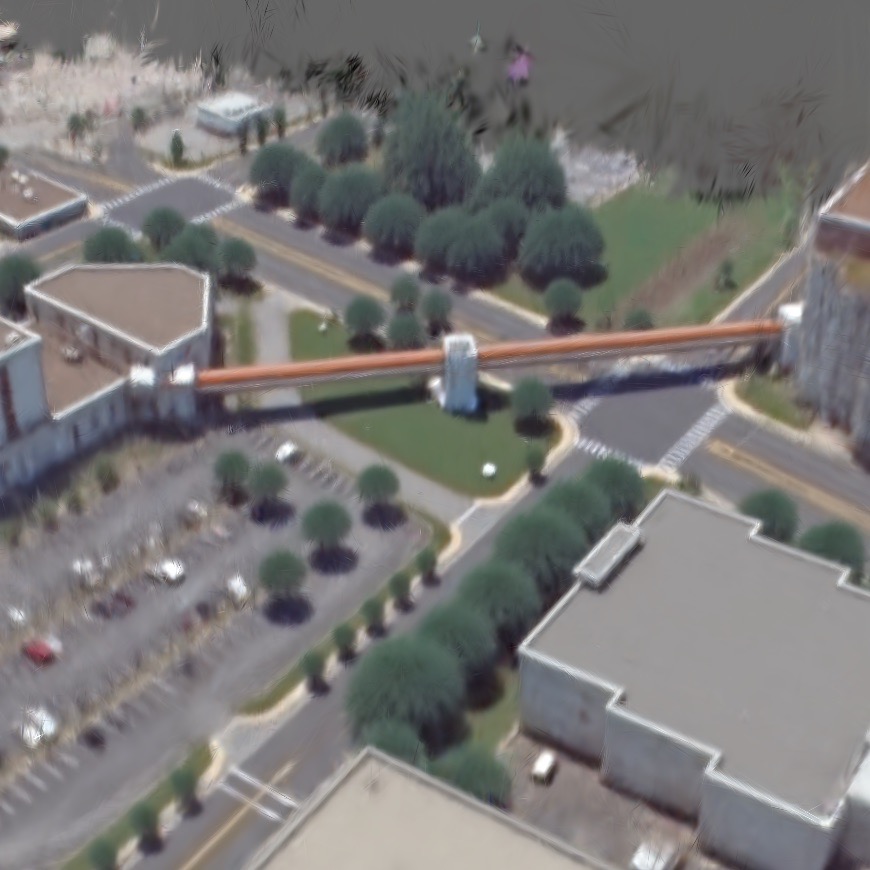} & \includegraphics[width=0.31\textwidth]{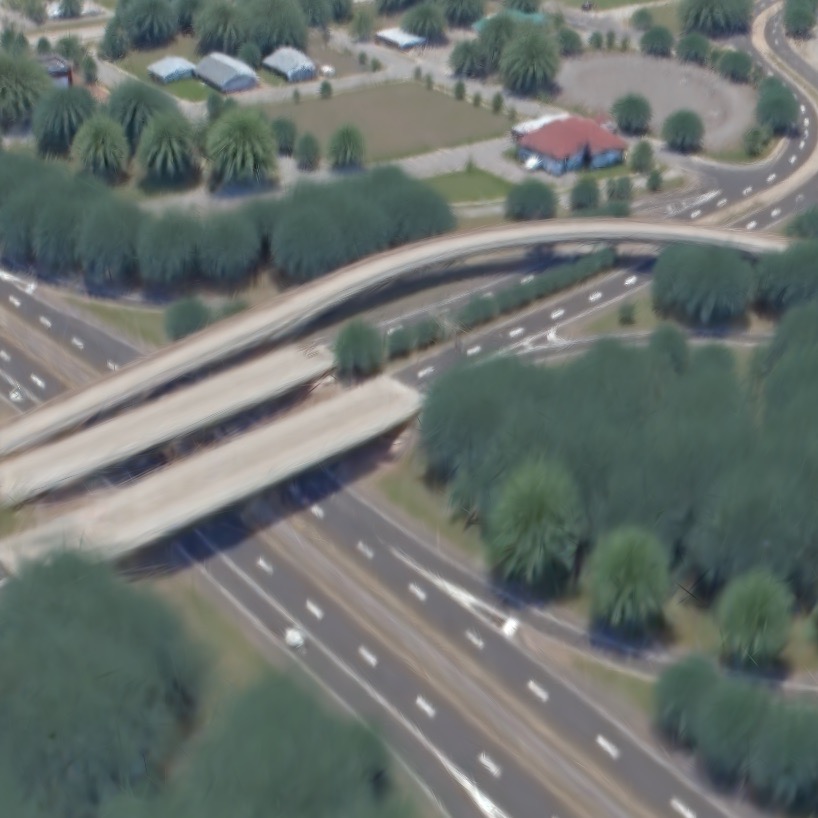} & \includegraphics[width=0.31\textwidth]{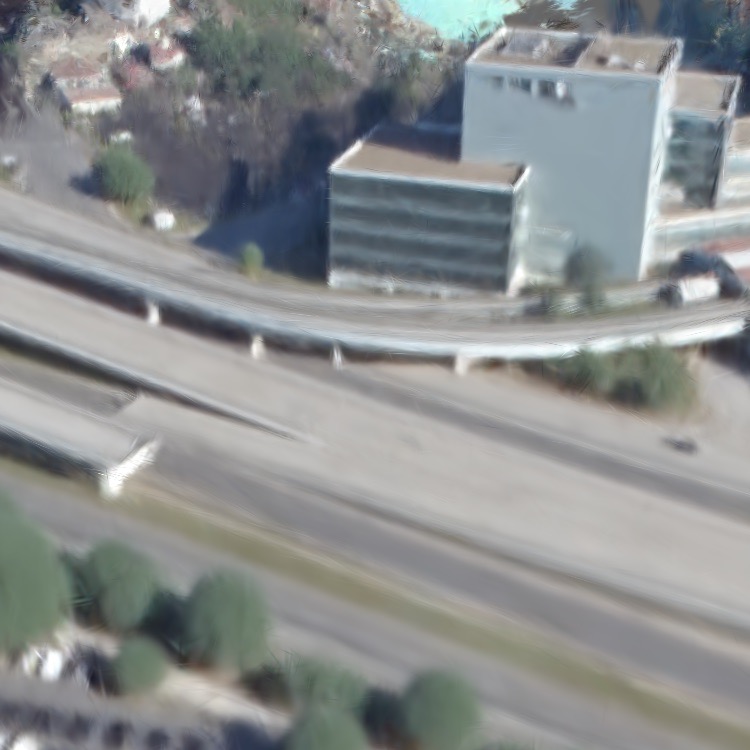} \\
(a) JAX\_068 & (b) JAX\_175 & (c) JAX\_214 \\
\end{tabular}
\caption{\textbf{Qualitative results for bridges.} We present the render results for bridges appears in JAX\_068, JAX\_214 and JAX\_175, demonstrating the method's ability to handle complex mulit-level structures that are typically challenging for standard reconstruction pipelines.}
\label{fig:bridge_qualitative}
\end{figure}

In addition to dense building clusters, we evaluate our method's performance on scenes with complex topological structures, such as bridges. Figure \ref{fig:bridge_qualitative} illustrates renders of bridges in JAX\_068, JAX\_214 and JAX\_175, a typically difficult case for standard photogrammetry due to the thin structural components. Our method successfully recovers the connectivity of the bridge span while synthesizing realistic water textures. The diffusion-based refinement effectively regularizes the geometry, preventing the characteristic "melting" artifacts often observed in thin structures when using satellite-only reconstruction.

\subsubsection{Visualizing transient object handling via per-image embeddings.}
\begin{figure*}[t]
\centering
\setlength{\tabcolsep}{1pt} %
\renewcommand{\arraystretch}{0.5} %
\resizebox{\textwidth}{!}{%
\begin{tabular}{ccccc}
  \includegraphics[width=0.19\textwidth]{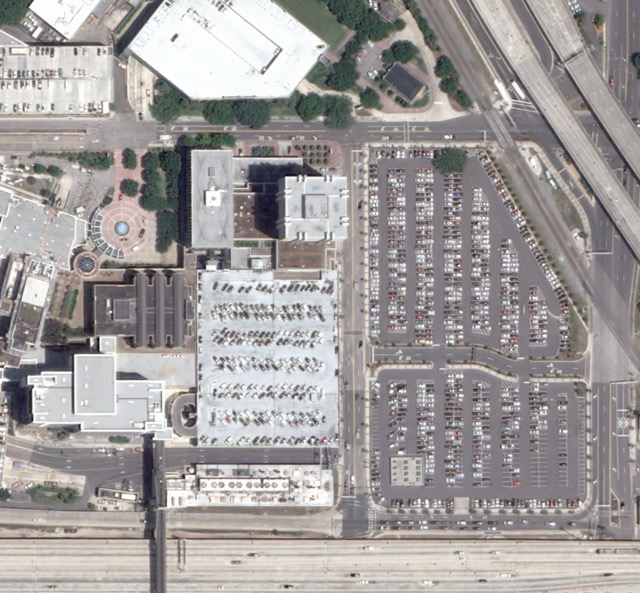} &
  \includegraphics[width=0.19\textwidth]{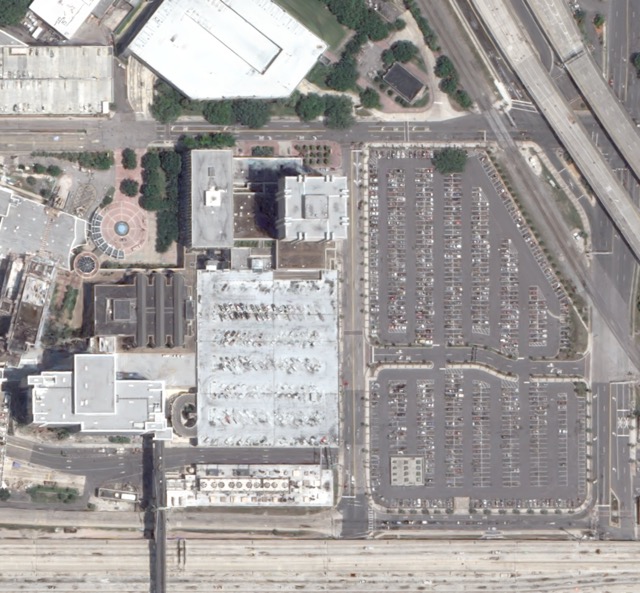} &
  \includegraphics[width=0.19\textwidth]{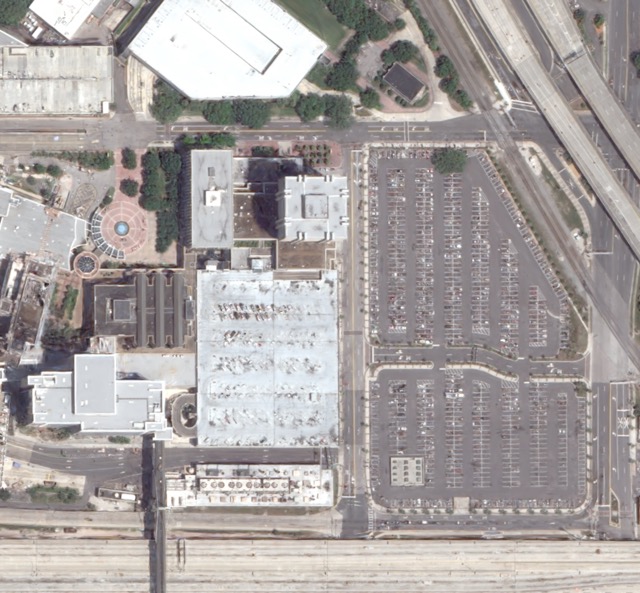} &
  \includegraphics[width=0.19\textwidth]{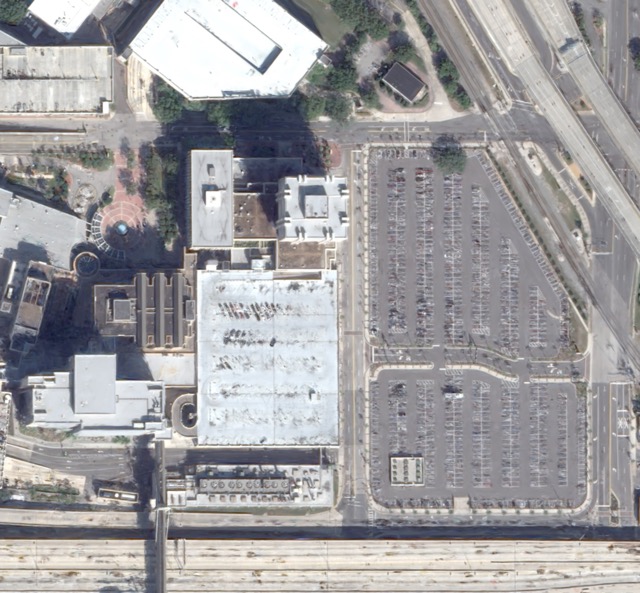} &
  \includegraphics[width=0.19\textwidth]{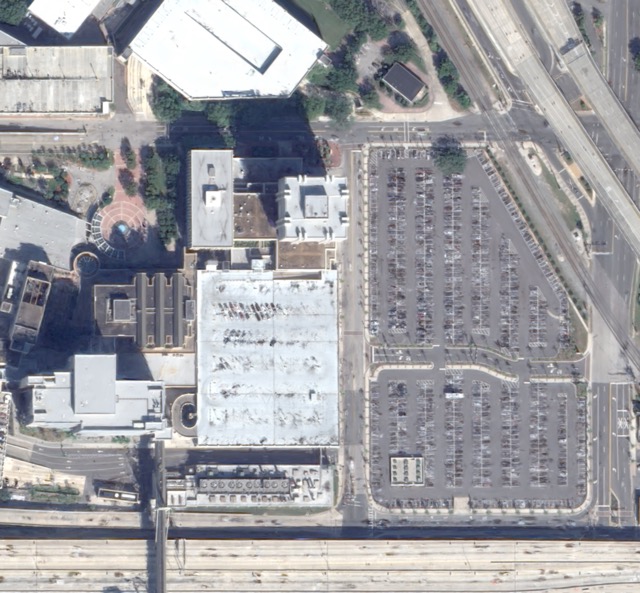} \\
  \scriptsize Emb. 00 & \scriptsize Emb. 01 & \scriptsize Emb. 02 & \scriptsize Emb. 03 & \scriptsize Emb. 04 \\[3pt]

  \includegraphics[width=0.19\textwidth]{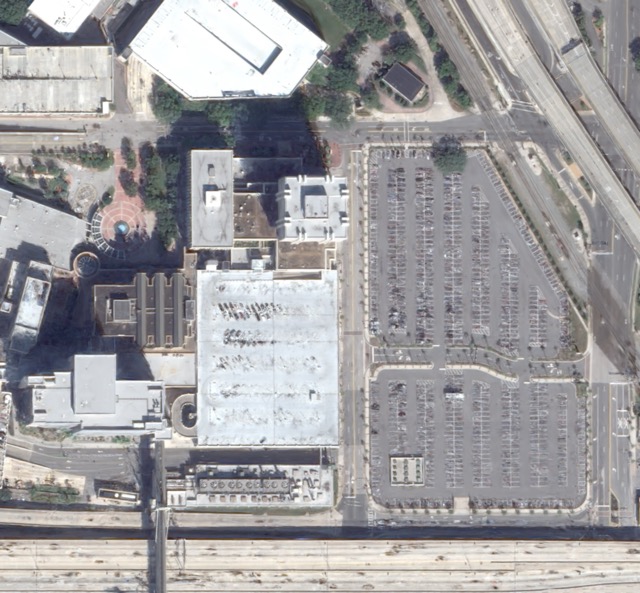} &
  \includegraphics[width=0.19\textwidth]{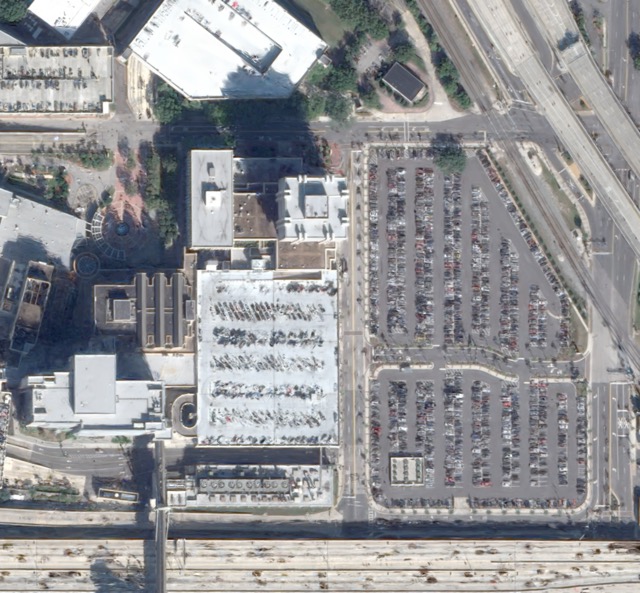} &
  \includegraphics[width=0.19\textwidth]{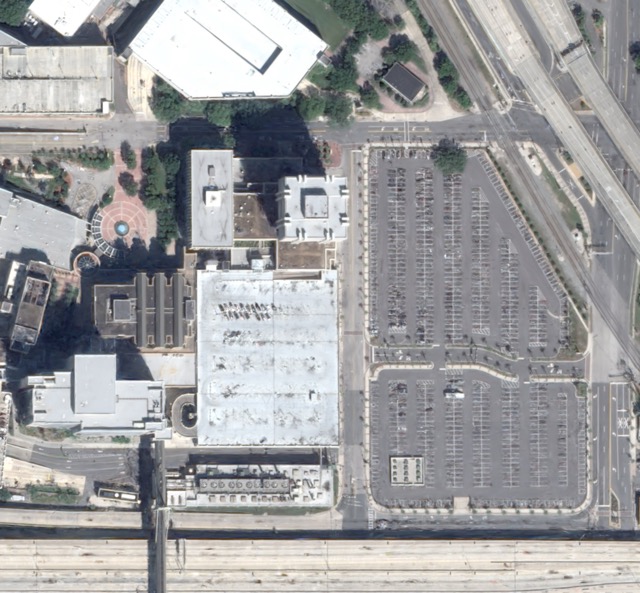} &
  \includegraphics[width=0.19\textwidth]{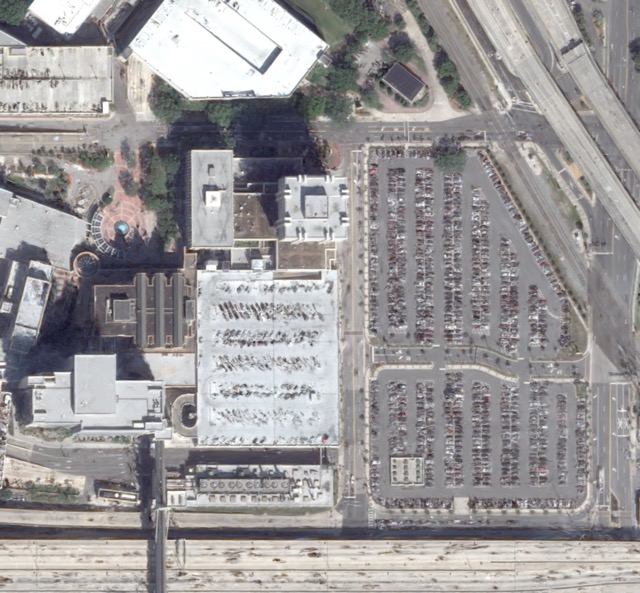} &
  \includegraphics[width=0.19\textwidth]{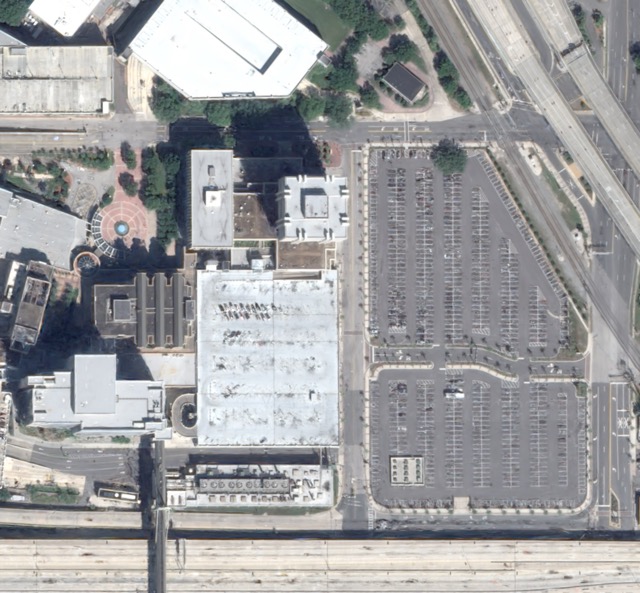} \\
  \scriptsize Emb. 05 & \scriptsize Emb. 06 & \scriptsize Emb. 07 & \scriptsize Emb. 08 & \scriptsize Emb. 09 \\[3pt]

  \includegraphics[width=0.19\textwidth]{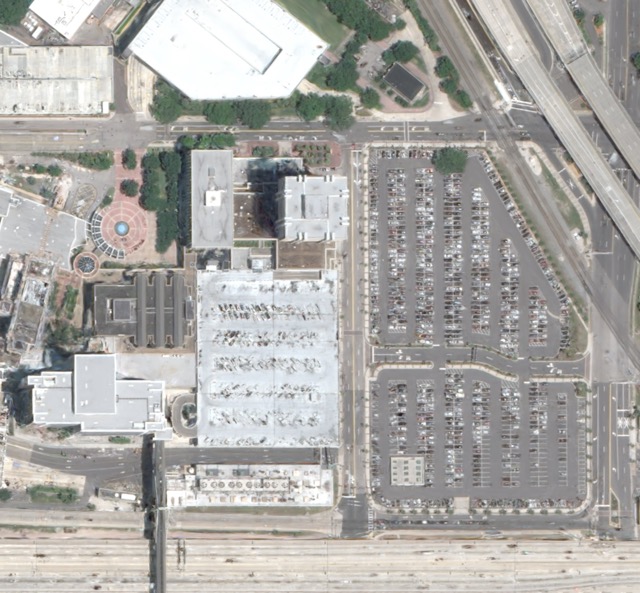} &
  \includegraphics[width=0.19\textwidth]{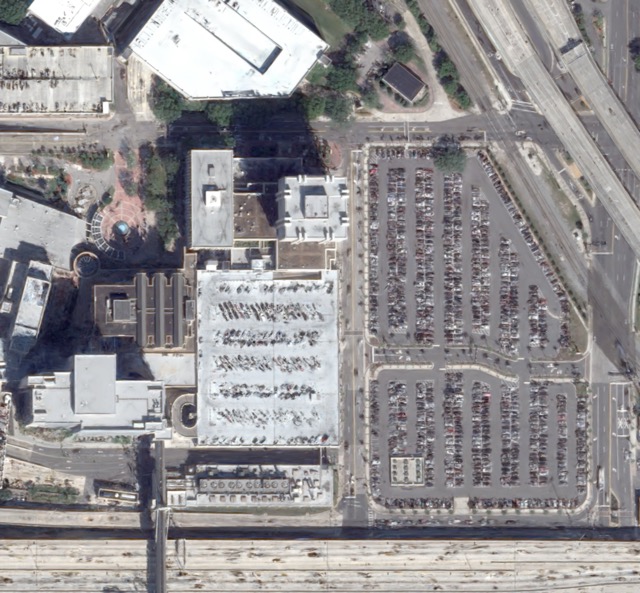} &
  \includegraphics[width=0.19\textwidth]{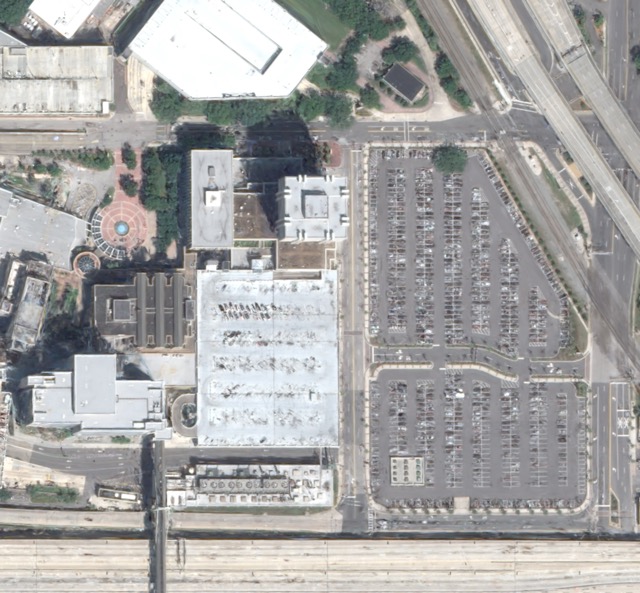} &
  \includegraphics[width=0.19\textwidth]{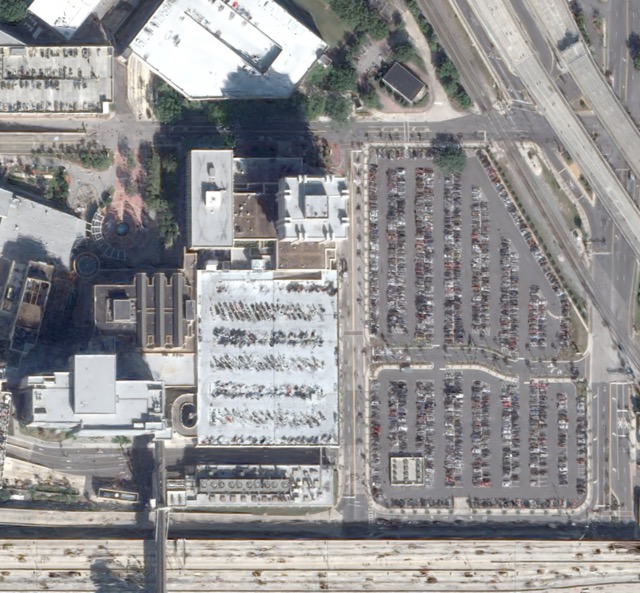} &
  \includegraphics[width=0.19\textwidth]{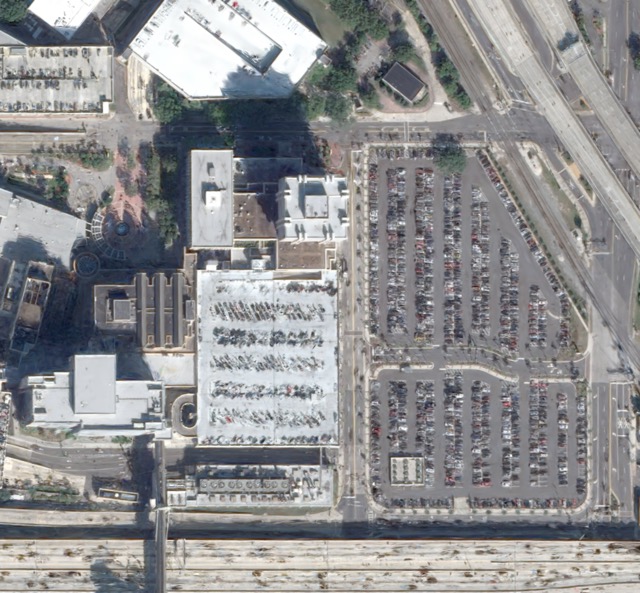} \\
  \scriptsize Emb. 10 & \scriptsize Emb. 11 & \scriptsize Emb. 12 & \scriptsize Emb. 13 & \scriptsize Emb. 14 \\[3pt]

  \includegraphics[width=0.19\textwidth]{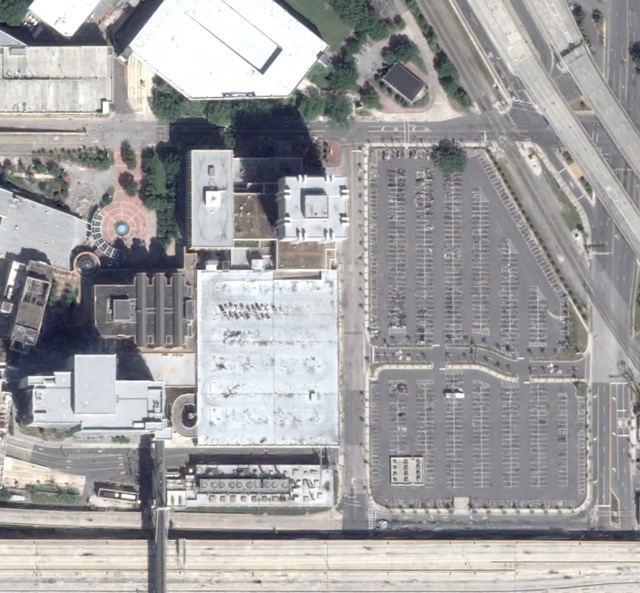} &
  \includegraphics[width=0.19\textwidth]{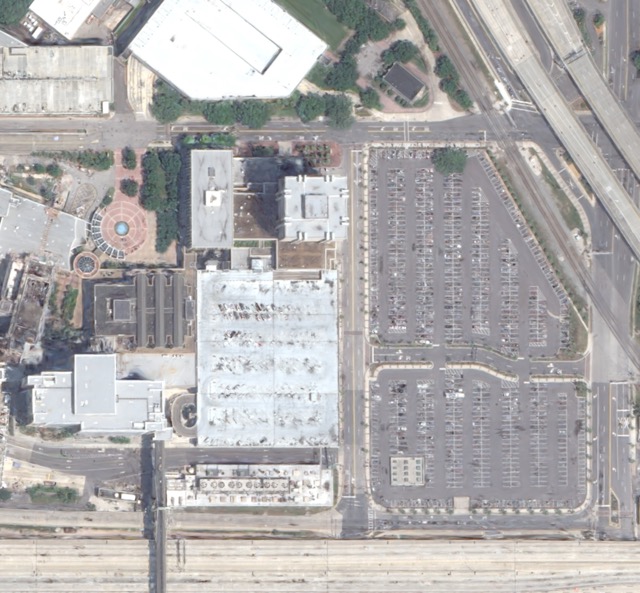} &
  \includegraphics[width=0.19\textwidth]{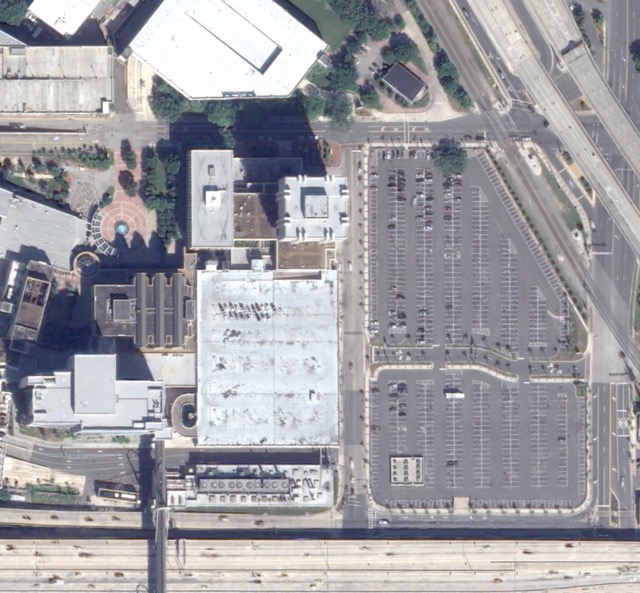} &
  \includegraphics[width=0.19\textwidth]{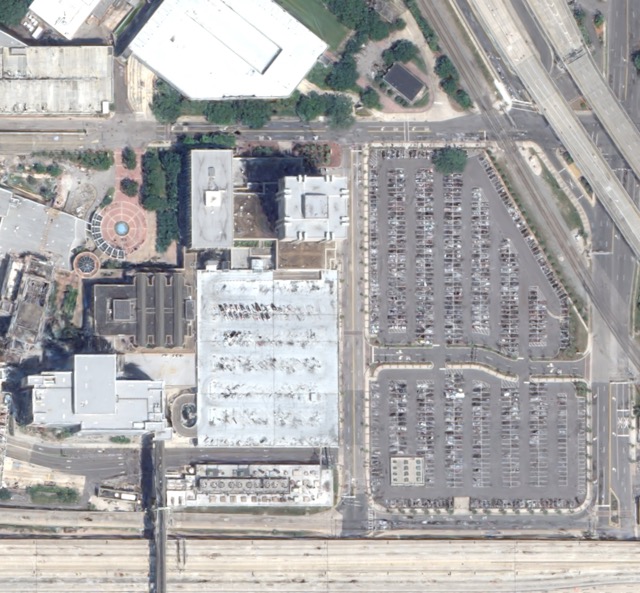} &
  \includegraphics[width=0.19\textwidth]{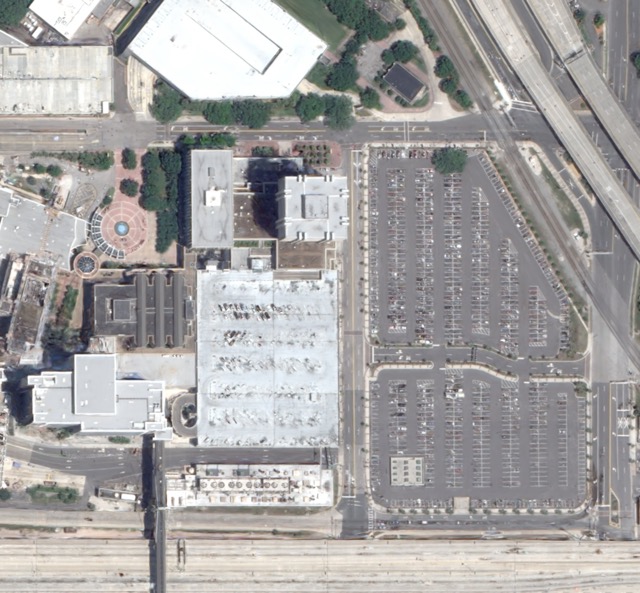} \\
  \scriptsize Emb. 15 & \scriptsize Emb. 16 & \scriptsize Emb. 17 & \scriptsize Emb. 18 & \scriptsize Emb. 19 \\
\end{tabular}%
}
\caption{\textbf{Visualizing transient object handling via per-image embeddings.}
We render the same viewpoint using 20 different learned appearance embeddings (Emb. 00--19).
Observe that transient objects, such as the vehicles on the road, exhibit varying degrees of visibility across different embeddings (e.g., clearly visible in some, faded or absent in others), while the static building geometry remains consistent.
This qualitatively demonstrates that our per-image appearance modeling effectively disentangles transient elements from the underlying static 3D structure, preventing dynamic artifacts from corrupting the geometric reconstruction.}
\label{fig:embedding_transients}
\end{figure*}

A key challenge in multi-date satellite reconstruction is the handling of dynamic elements, such as moving vehicles and pedestrians, which can introduce ghosting artifacts. Our approach addresses this by learning per-image appearance embeddings $e_j$ that capture photometric variations specific to each capture date. As visualized in \Cref{fig:embedding_transients}, rendering the same viewpoint across 20 distinct appearance embeddings reveals that transient objects exhibit significant variability, appearing clearly in some embeddings while fading or vanishing in others. This qualitative evidence suggests that our appearance modeling effectively acts as a ``sink'' for transient data that does not align with the static 3D geometry. By absorbing these inconsistencies into the appearance code rather than the geometric parameters, the optimization naturally disentangles transient elements from the underlying static structure, ensuring a clean and consistent geometric reconstruction.

\subsubsection{Sensitivity analysis of refinement text prompts.}
\begin{figure}[htbp]
    \centering
    \includegraphics[width=0.95\linewidth]{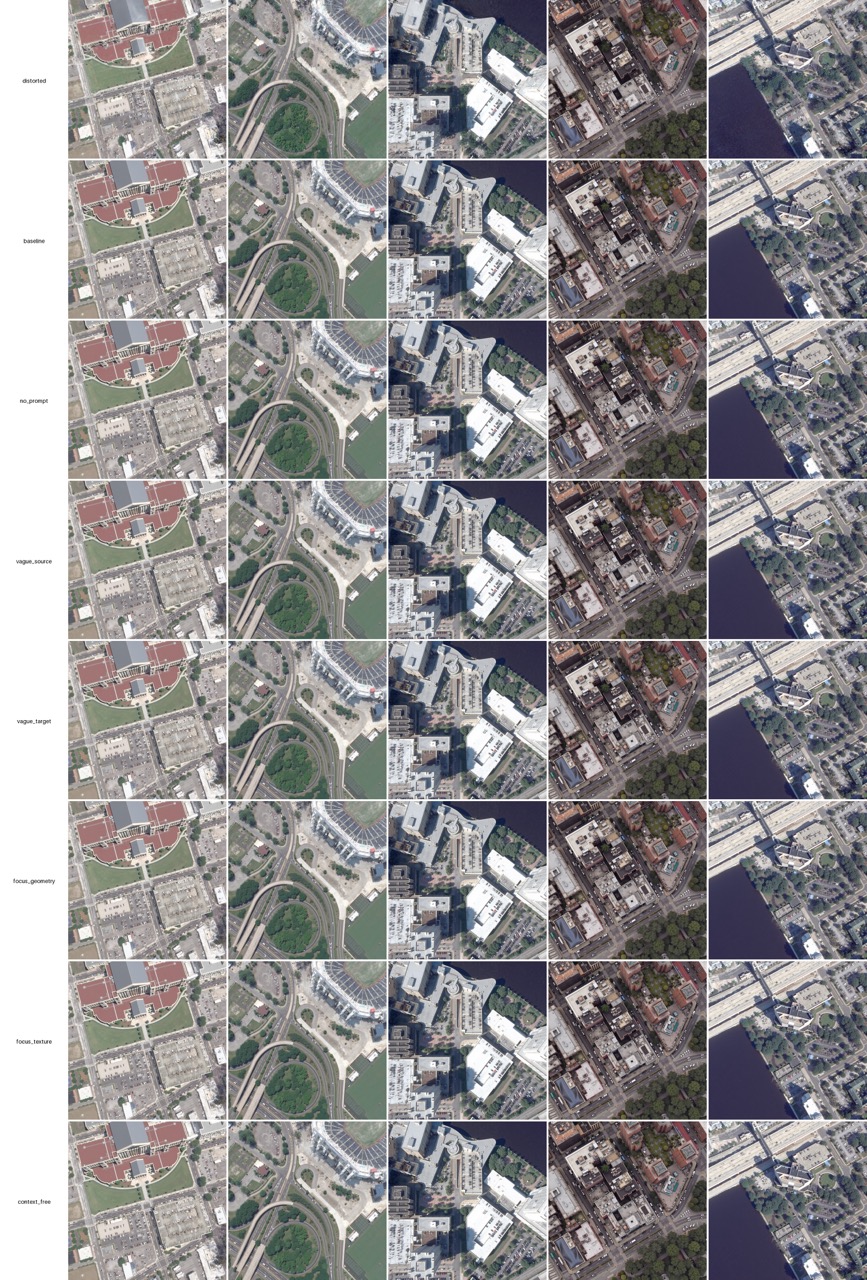}
    \caption{\textbf{Refine renders with different prompt strategies.} }
    \label{fig:prompt}
\end{figure}

A practical concern for deployment is whether our method is sensitive to the specific wording of the FlowEdit text prompts. To investigate this, we evaluate six prompting strategies listed in \Cref{tab:prompt_list}, ranging from highly descriptive source/target pairs (\textbf{Baseline}) to vague descriptions (\textbf{Vague Source}, \textbf{Vague Target}), domain-specific variants that emphasize geometry or texture (\textbf{Focus Geometry}, \textbf{Focus Texture}), and a minimal context-free pair (\textbf{Context Free}). As shown in \Cref{fig:prompt}, the visual quality of the refined renders remains largely consistent across all strategies, with only minor differences in fine-grained texture details. This robustness suggests that our method does not require careful prompt engineering and is tolerant to moderate variations in prompt specificity. We note that this limited prompt steerability is by design: because optimization anchors the scene to the input satellite imagery through our $75/25$ sampling of refined and original views, the text prompt acts as an appearance \emph{regularizer} that guides artifact removal and detail recovery, rather than as a free-form controller of scene content. Weak steerability is therefore a deliberate trade-off in favor of geometric fidelity to the captured scene.

\begin{table}[htbp]
\centering
\small
\caption{\textbf{List of text prompts used in sensitivity analysis.} We evaluate six different prompting strategies to test the robustness of our method.}
\label{tab:prompt_list}
\resizebox{\textwidth}{!}{%
\begin{tabular}{p{2.5cm}p{6cm}p{6cm}}
\toprule
\textbf{Strategy} & \textbf{Source Prompt ($P_\text{src}$)} & \textbf{Target Prompt ($P_\text{tar}$)} \\
\midrule
\textbf{Baseline} & Satellite image of an urban area with modern and older buildings, roads, green spaces. Some areas appear distorted, with blurring and warping artifacts. & Clear satellite image of an urban area with sharp buildings, smooth edges, natural lighting, and well-defined textures. \\
\midrule
\textbf{Vague Source} & A blurry satellite image of an urban area. & Clear satellite image of an urban area with sharp buildings, smooth edges, natural lighting, and well-defined textures. \\
\midrule
\textbf{Vague Target} & Satellite image of an urban area with modern and older buildings, roads, green spaces. Some areas appear distorted, with blurring and warping artifacts. & A clear satellite image of an urban area. \\
\midrule
\textbf{Focus Geometry} & Satellite image of an urban area with modern and older buildings, roads, green spaces. Some areas appear distorted, with blurring and warping artifacts. & Clear satellite image of an urban area with geometrically precise buildings, flat rooftops, straight edges, and well-defined roads. \\
\midrule
\textbf{Focus Texture} & Satellite image of an urban area with modern and older buildings, roads, green spaces. Some areas appear distorted, with blurring and warping artifacts. & Clear satellite image of an urban area with realistic, high-resolution textures, detailed facades, clear vegetation, and natural lighting. \\
\midrule
\textbf{No Prompt} & - & - \\
\midrule
\textbf{Context Free} & distorted, blurring, warping artifacts & clear, sharp, smooth edges, natural lighting, well-defined textures \\
\bottomrule
\end{tabular}%
}
\end{table}

\subsubsection{Multi-block scalability via combined imagery.} 
\begin{figure}
    \centering
    \includegraphics[width=\linewidth]{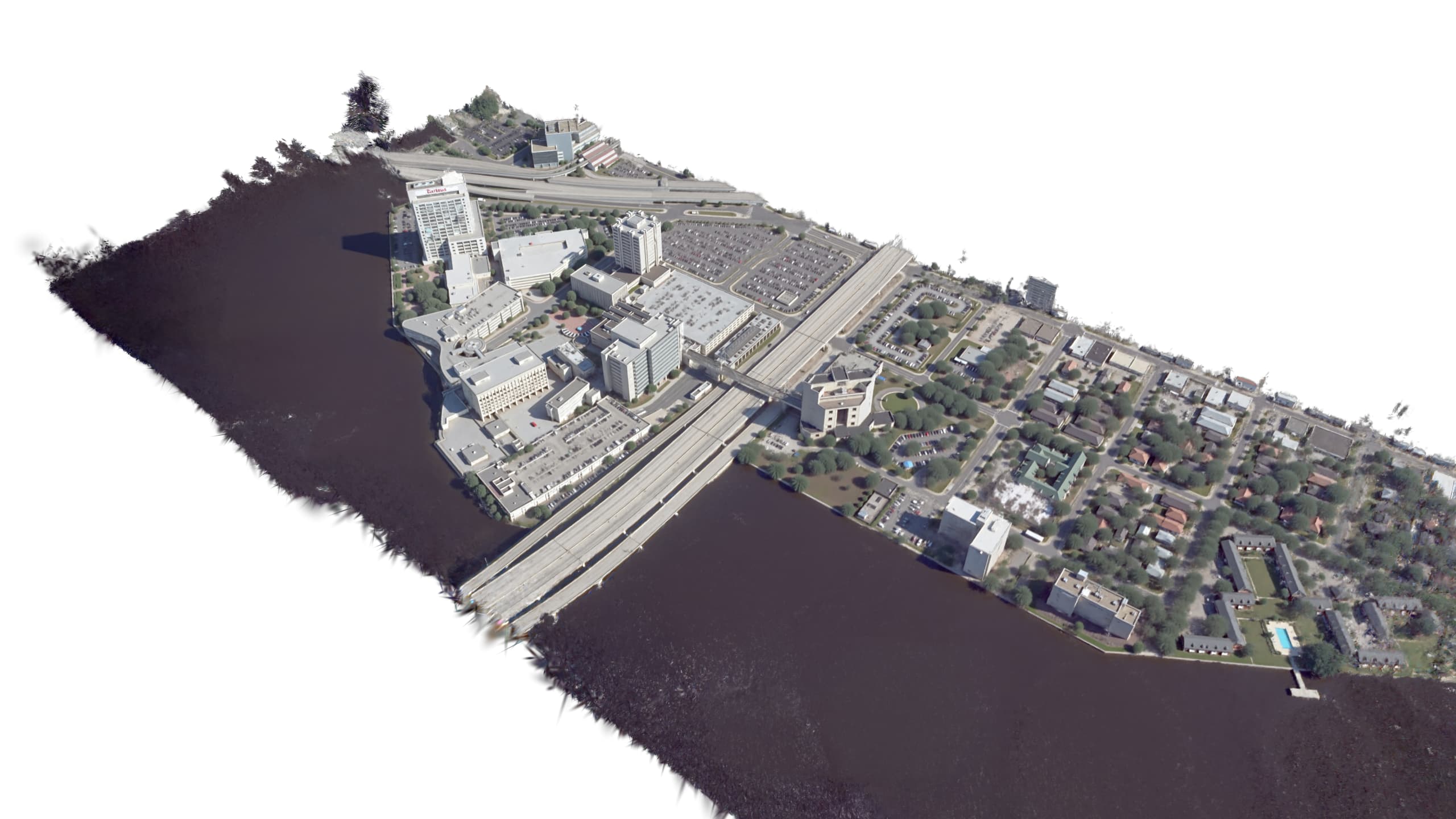}
    \caption{\textbf{Multi-block reconstruction from combined JAX\_214 and JAX\_260 AOIs.} Skyfall-GS produces a seamless, artifact-free 3D scene across a $\sim$1km $\times$ 512m urban region, demonstrating scalability to multi-block environments without boundary stitching artifacts.}
    \label{fig:combined_suppl}
\end{figure}

To validate the scalability of our framework across larger, continuous urban regions, we combined satellite imagery from two adjacent areas (JAX\_214 and JAX\_260) into a single, unified dataset covering approximately 1km $\times$ 512m. To accommodate the expanded spatial extent, we scaled the Iterative Dataset Update (IDU) look-at grid to $6 \times 3$, while maintaining our standard 5-episode refinement schedule (10,000 iterations per episode). The total training time for this expanded region took approximately 9 hours (1.5 hours for the initial reconstruction stage and 7.5 hours for the synthesis stage) on a single NVIDIA RTX A6000 GPU. The final converged scene yielded $\sim$3.5 million Gaussians with a peak final training memory footprint of $\sim$46GB. As shown in \Cref{fig:combined_suppl}, the expanded area demonstrate seamless boundary consistency, particularly evident along the connecting highway structures, confirming that our method scales robustly to multi-block environments without introducing stitching artifacts at the region boundaries. We further quantify this benefit on a boundary-spanning orbit that straddles the two tiles: the two single-AOI models, restricted to their own tile, score $50.13/3.54$ and $53.18/3.57$ ($\text{FID}_\text{CLIP}/$CMMD) on this trajectory, whereas the merged model achieves $32.15/2.42$ ($\sim$36--38\% lower $\text{FID}_\text{CLIP}$ and $\sim$32\% lower CMMD). The merged model can thus synthesize boundary regions that fall outside either single-AOI model's coverage.

\subsubsection{Episode-vs-coverage analysis of curriculum strategy.}
\begin{figure}[htbp]
\centering
\includegraphics[width=0.95\linewidth]{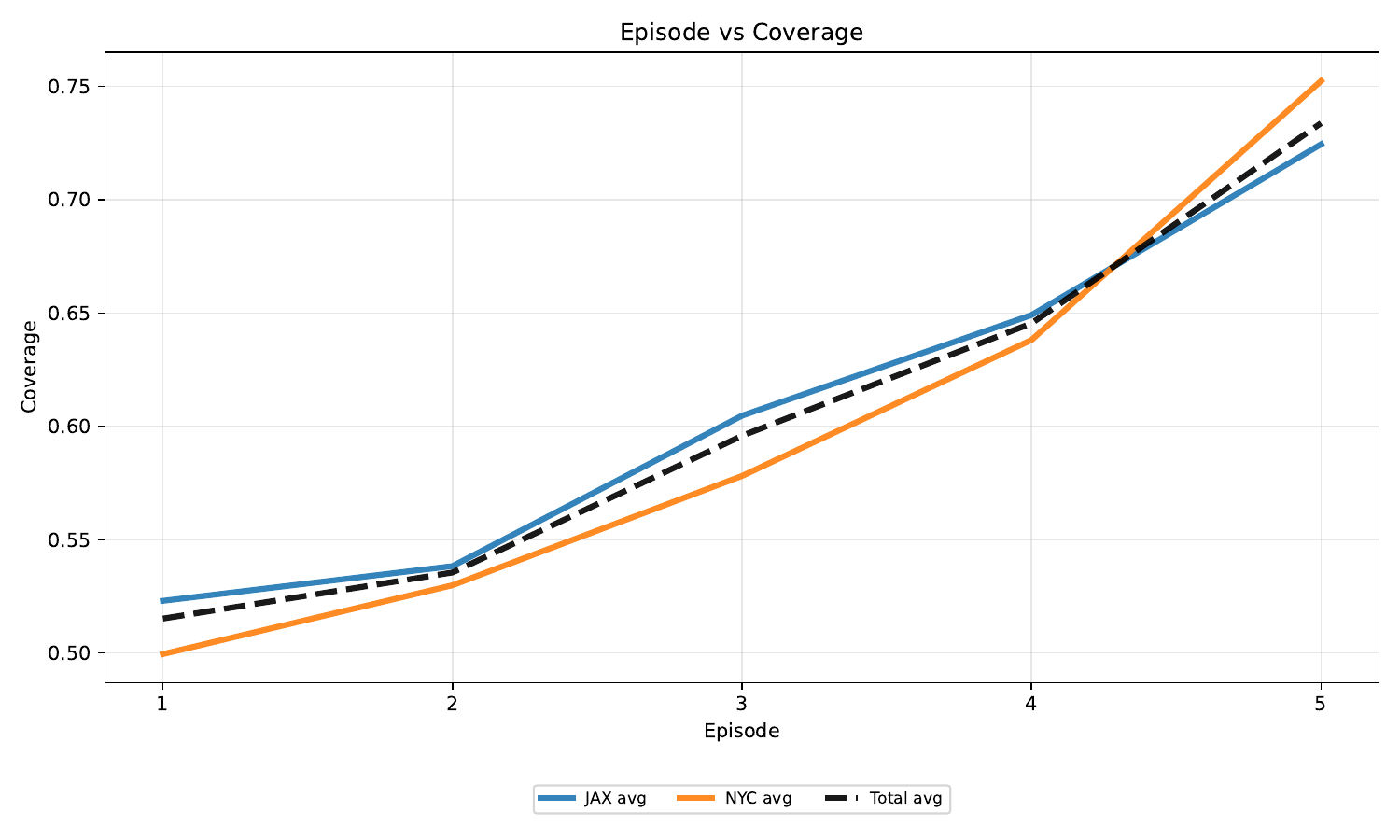}
\caption{\textbf{Episode-vs-Coverage analysis.} The plot illustrates the cumulative surface coverage ratio increasing across refinement episodes. The curriculum-based strategy effectively exposes occluded regions, particularly vertical facades, as the camera elevation descends.}
\label{fig:coverage_curve}
\end{figure}

To quantify the effectiveness of the IDU module in revealing occluded regions, we present an Episode-vs-Coverage analysis (Figure~\ref{fig:coverage_curve}). Since ground truth 3D geometry is unavailable for these satellite scenes, we use the final converged 3DGS model as a proxy for the total scene surface. We compute the cumulative coverage by optimizing a visibility attribute for every Gaussian point against the camera poses utilized in each episode. As shown in the figure, the coverage ratio steadily increases from $\sim$0.50 in Episode 1 to $\sim$0.75 in Episode 5. This consistent gain confirms that our curriculum strategy, which progressively lowers camera elevation from $85^\circ$ to $45^\circ$, successfully reveals and reconstructs vertical facade geometry that was initially occluded in the top-down satellite views. However, we acknowledge a limitation in this metric: because it calculates coverage based on reconstructed points, it cannot account for ``true holes'' (surface areas that were never generated at all because they were completely occluded from all sampled views). Future work could address this by dynamically sampling IDU cameras to target specific geometric uncertainties or detected holes.

\subsubsection{Iterative vs.\ single-pass refinement.}
We verify that the \emph{iterative} nature of IDU, rather than a single diffusion pass, is responsible for the quality gains. At matched compute on JAX\_068 (\Cref{tab:iter_vs_single}), the satellite-only reconstruction (Stage-1) scores $84.99/5.64$ ($\text{FID}_\text{CLIP}/$CMMD). A single-pass variant that refines five elevations once via FlowEdit and fits a single-pass 3DGS already reaches $51.46/4.16$, beating every reconstruction baseline (\eg, EOGS at $86.08/5.54$, a $-40\%/-25\%$ relative improvement). Our full iterative IDU further reduces these to $28.35/2.875$ (an additional $\sim$45\%/$\sim$31\%), because each episode re-renders from the \emph{updated} 3DGS so that subsequent edits operate on progressively cleaner inputs---a self-reinforcing loop the single-pass variant cannot exploit.

\begin{table}[t]
    \centering
    \caption{\textbf{Iterative vs.\ single-pass refinement} at matched compute on JAX\_068. Iterating the dataset update is essential: each episode refines renders from the updated 3DGS, compounding quality gains a single diffusion pass cannot achieve.}
    \label{tab:iter_vs_single}
    \begin{tabular}{lcc}
        \toprule
        Refinement strategy & $\text{FID}_{\text{CLIP}}\,\downarrow$ & CMMD$\,\downarrow$ \\
        \midrule
        Stage-1 (satellite reconstruction only) & 84.99 & 5.640 \\
        Single-pass diffusion & 51.46 & 4.160 \\
        \textbf{Iterative IDU (Ours)} & \textbf{28.35} & \textbf{2.875} \\
        \bottomrule
    \end{tabular}
\end{table}

\subsubsection{Cross-view consistency.}
Beyond per-frame quality, we quantify the temporal stability of our renders under continuous camera motion. Following the subject-consistency protocol of VBench~\cite{huang2024vbench} (with DINOv2~\cite{oquab2024dinov2} substituted for DINO), we compute $1-\cos$ between the DINOv2 ViT-L/14 features of consecutive frames along a $17^\circ$ orbit on JAX\_068, where a lower value indicates more consistent appearance across views. The GES reference (oracle) scores $0.0213$, while our renders improve from $0.0340$ at $N_s=1$ to $0.0264$ at $N_s=2$, closing roughly $60\%$ of the gap to the oracle. This confirms that multi-sample consensus suppresses high-frequency flickering across views and rules out severe inconsistency, consistent with the free-flight videos provided on our project website.

\subsubsection{Stochastic appearance diversity.}
\begin{figure}[htbp]
    \centering
    \includegraphics[width=0.95\linewidth]{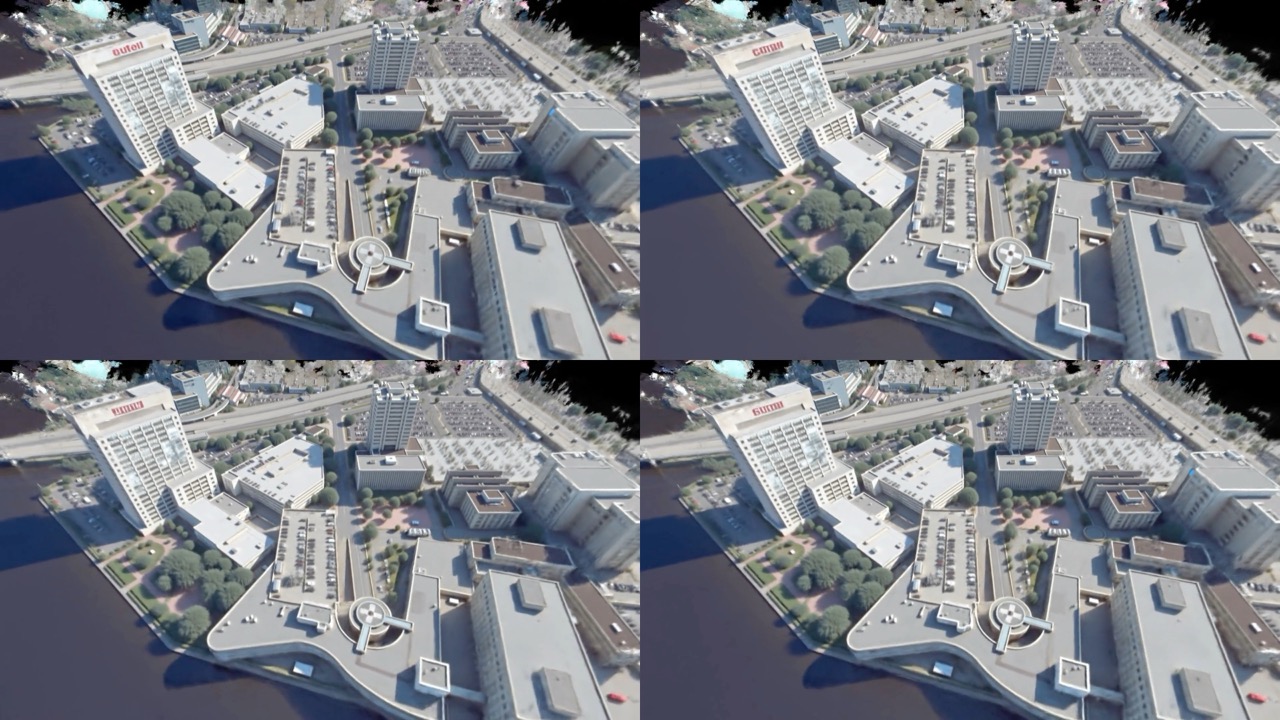}
    \caption{\textbf{Demonstration of stochastic appearance diversity while preserving geometric consistency.} Our method generates diverse plausible textures for identical underlying geometry across different random seeds. Notice how the red signage text on the building facade varies distinctively (e.g., ``Outeil'' \vs ``CUTAN'') while the building's structural footprint remains fixed, confirming that our framework successfully disentangles geometric reconstruction (grounded in satellite data) from generative appearance synthesis (variable via diffusion).}
    \label{fig:seed_diversity}
\end{figure}

To demonstrate the generative capacity of our hybrid framework, we evaluate the stochastic diversity of the synthesized textures in Figure~\ref{fig:seed_diversity}. By varying the random seed during the diffusion refinement stage while maintaining the same geometric initialization, our method produces diverse yet plausible surface details for identical underlying structures. As illustrated in the figure, detailed features such as the text on the red building signage vary distinctively (e.g., ``Outeil'' \vs ``CUTAN''). Crucially, the macroscopic building footprint remains geometrically fixed, confirming that our framework successfully disentangles the reconstruction of physical geometry (grounded in satellite constraints) from the generative synthesis of high frequency appearance.

\end{document}